\documentclass[preprint,11pt]{elsarticle}
\journal{}
\biboptions{sort&compress}
\usepackage{algorithm}
\usepackage{algpseudocode}
\usepackage{amsmath}
\usepackage{amsmath,bm}
\usepackage{amssymb}
\usepackage{bm}
\usepackage{booktabs}
\usepackage{color}
\usepackage{times}
\usepackage{epsfig}
\usepackage{epstopdf}
\usepackage{flafter}
\usepackage{float}
\usepackage{ifthen}
\usepackage{multirow}
\usepackage{makecell} 
\usepackage{pifont}
\usepackage{subfigure}
\usepackage{threeparttable}
\usepackage{url}  
\usepackage{subfigure,graphicx}
\usepackage{float}
\usepackage{color,multirow}
\usepackage{makecell}
\usepackage{url}  
\usepackage{ifthen}  
\usepackage[T1]{fontenc}
\usepackage{graphicx}
\usepackage{graphics}
\usepackage{multirow}
\usepackage{makecell}
\usepackage{booktabs}
\newtheorem{theorem}{\text{Theorem}}
\newtheorem{myDef}{Definition}

\newcommand{\argmin}{\mathop{\mathrm{arg\,min}}}

\renewcommand{\algorithmicensure}{\textbf{Output:}} 
\newcommand{\tabincell}[2]{\begin{tabular}{@{}#1@{}}#2\end{tabular}}

\usepackage[left=1in,right=1in,top=1in,bottom=1in,letterpaper]{geometry}
\begin{document}

\title{Tensor $N$-tubal rank and its convex relaxation for low-rank tensor recovery}
\author[a]{Yu-Bang Zheng}
\ead{zhengyubang@163.com}
\author[a]{Ting-Zhu Huang\corref{cor}}
\ead{tingzhuhuang@126.com}
\author[a]{Xi-Le Zhao}
\ead{xlzhao122003@163.com}
\author[a]{Tai-Xiang Jiang}
\ead{taixiangjiang@gmail.com}
\author[b]{Teng-Yu Ji}
\ead{tengyu\_j66@126.com}
\author[c]{Tian-Hui Ma}
\ead{nkmth0307@126.com}

\cortext[cor]{Corresponding author. Tel.: +86 28 61831016, Fax: 86 28 61831280.}
\address[a]{School of Mathematical Sciences/Research Center for Image and Vision Computing, University of Electronic Science and Technology of China, Chengdu, Sichuan 611731, P.R.China}
\address[b]{School of Science, Northwestern Polytechnical University, Xi'an, Shaanxi 710072, P.R.China.}
\address[c]{School of Mathematics and Statistics, Xi'an Jiaotong University, Xi'an, Shaanxi 710049, P.R.China.}

\begin{frontmatter}
\begin{abstract}
As low-rank modeling has achieved great success in tensor recovery, many research efforts devote to defining the tensor rank. Among them,
the recent popular tensor tubal rank, defined based on the tensor singular value decomposition (t-SVD), obtains promising results.
However, the framework of the t-SVD and the tensor tubal rank are applicable only to three-way tensors and lack of flexibility to handle different correlations along different modes.
To tackle these two issues, we define a new tensor unfolding operator, named mode-$k_1k_2$ tensor unfolding, as the process of lexicographically stacking the mode-$k_1k_2$ slices of an $N$-way tensor into a three-way tensor, which is a three-way extension of the well-known mode-$k$ tensor matricization. Based on it, we define a novel tensor rank, the tensor $N$-tubal rank, as a vector whose elements contain the tubal rank of all mode-$k_1k_2$ unfolding tensors, to depict the correlations along different modes. To efficiently minimize the proposed $N$-tubal rank, we establish its convex relaxation: the weighted sum of tensor nuclear norm (WSTNN). Then, we apply WSTNN to low-rank tensor completion (LRTC) and tensor robust principal component analysis (TRPCA).
The corresponding WSTNN-based LRTC and TRPCA models are proposed, and two efficient alternating direction method of multipliers (ADMM)-based algorithms are developed to solve the proposed models. Numerical experiments demonstrate that the proposed models significantly outperform the compared ones.
\end{abstract}
\begin{keyword}
  Low-rank tensor recovery (LRTR),
  mode-$k_1k_2$ tensor unfolding,
  the tensor $N$-tubal rank,
  the weighted sum of tensor nuclear norm (WSTNN),
  alternating direction method of multipliers (ADMM).
\end{keyword}
\end{frontmatter}

\section{Introduction}  \label{sec:Int}
As a multidimensional array, the tensor \cite{kolda2009tensor} plays an increasingly significant role in many applications, such as color image/video processing \cite{MA2016510,HUANG2018147,TIPjiang2018fastderain,NonlocalTXie,MADATHIL2018376,WangSparse2018,XiongjunZhang}, hyperspectral/multispectral image (HSI/MSI) processing \cite{Wang2017Hyperspectral,zhao2014SIAM,UXxiaofu,chang2017weighted,CSTF,JiTGRS,FastHyDe}, background subtraction \cite{kajo2018svd,XuTensor}, video rain streak removal \cite{jiang2017novel,Wei2017video,li2018video}, and magnetic resonance imaging (MRI) data recovery \cite{JiTV, jiang2018matrix}.
Many of them can be formulated as the tensor recovery problem, i.e., recovering the underlying tensor from a corrupted observation tensor. Particularly, as two typical tensor recovery problems, tensor completion aims to complete missing elements, and tensor robust principal component analysis (TRPCA) aims to remove sparse outliers. The key of tensor recovery is to explore the redundancy prior of the underlying tensor, which is usually formulated as low-rankness. Thus, low-rank modeling has been widely studied and has achieved great success in tensor recovery task.

The traditional matrix recovery problem is actually the two-way tensor recovery problem. Since the matrix rank, measured by the number of non-zero singular values, is powerful to capture the global information of matrix, most matrix recovery methods aim to minimize the matrix rank \cite{Cands2009,JianfengCai2010,LRMR2014,LRMR2016}. However, directly minimizing the matrix rank is NP-hard \cite{Hillar2009Most}. To tackle this issue, the nuclear norm ($\|\cdot\|_{\ast}$), the sum of non-zero singular values, has been proposed to approximate the matrix rank, leading to great successes \cite{Cands2009,JianfengCai2010}.

The tensor recovery problem can be viewed as an extension of the matrix recovery problem. Inspired by the success of matrix rank minimization, it seems natural to recovery the underlying tensor by minimizing the tensor rank. Mathematically, the low-rank tensor recovery (LRTR) model can be generally written as

\begin{equation*}
\begin{aligned}
    \min_{\mathcal{X}}&~~\text{rank}(\mathcal{X})+\lambda L(\mathcal{X},\mathcal{F}),\\
\end{aligned} \label{LRTR}
\end{equation*}
where $\mathcal{X}$ is the underlying tensor, $\mathcal{F}$ is the observed tensor, and $L(\mathcal{X},\mathcal{F})$ is the loss function between $\mathcal{X}$ and $\mathcal{F}$, e.g., $\mathcal{X}_{\Omega}=\mathcal{F}_{\Omega}$ for low-rank tensor completion (LRTC) and $\|\mathcal{F}-\mathcal{X}\|_{1}$ for TRPCA. The conclusive issue of LRTR is the definition of the tensor rank. However, unlike matrix rank, the definition of the tensor rank is not unique.
Many research efforts have been devoted to defining the tensor rank, and most of them are defined based on the corresponding tensor decomposition, such as the CANDECOMP/PARAFAC (CP) rank based on the CP decomposition \cite{carroll1980candelinc,zhou2018tensorCP,ZhangCP}, the Tucker rank based on the Tucker decomposition \cite{tucker1966some,gandy2011tensor,kressner2014low,Li2016,Liu2013PAMItensor,Li2017LowRankTC}, and the tensor tubal rank based on the tensor singular value decomposition (t-SVD) \cite{2013Kilmer,JilogDet,zhang2017exact}.

The CP rank and the Tucker rank are two most typical definitions of the tensor rank. The CP rank is defined as the minimum number of rank-one tensors required to express a tensor \cite{kolda2009tensor}, i.e.,
\begin{equation*}
\text{rank}_{\text{cp}}(\mathcal{X}):=\min\bigg\{r\big|\mathcal{X}=\sum_{i=1}^r \mathbf{a}_i^{1} \circ \mathbf{a}_i^{2} \circ \cdots \circ \mathbf{a}_i^{N},\mathbf{a}_i^{k}\in \mathbb{R}^{n_k}\bigg\},
\end{equation*}
where $\mathcal{X}$ is an $N$-way tensor and $\circ$ denotes the vector outer product. Although the measure of the CP rank is consistent with that of the matrix rank, it is hard to establish a solvable relaxation form for it. The Tucker rank is defined as a vector, the $k$-th element of which is the rank of the mode-$k$ unfolding matrix \cite{kolda2009tensor}, i.e.,
\begin{equation*}
\text{rank}_{\text{tc}}(\mathcal{X}):=\big(\text{rank}(X_{(1)}),\text{rank}(X_{(2)}),\cdots,\text{rank}(X_{(N)})\big),
\end{equation*}
where $\mathcal{X}$ is an $N$-way tensor and $X_{(k)}~(k=1,2,\cdots,N)$ is the mode-$k$ unfolding of $\mathcal{X}$. To efficiently minimize it, Liu et al. \cite{Liu2013PAMItensor} adopted the sum of the nuclear norm (SNN) of the mode-$k$ unfolding matrix as its convex relaxation, i.e.,
\begin{equation*}
\begin{aligned}
    \|\mathcal{X}\|_{\text{SNN}}:=\sum_{k=1}^{N}\alpha_k\|X_{(k)}\|_{*},\\
\end{aligned} \label{SNN}
\end{equation*}
where $\alpha_k \geq 0~(k=1,2,\cdots,N)$ and $\sum_{k=1}^{N}\alpha_k=1$. Based on SNN, Liu et al. \cite{Liu2013PAMItensor} established a LRTC model with three solving algorithms (SiLRTC, FaLRTC, and HaLRTC), and Goldfarb and Qin \cite{Goldfarb2014} proposed a TRPCA model. Although SNN can flexibly exploit the correlation along each mode by adjusting the weights $\alpha_k$ \cite{NgTGRS}, as pointed out in \cite{2013Kilmer,Meng2018}, when a tensor is unfolded to a matrix along one mode, the structures information along other modes will be destroyed. Thus, SNN is difficult to preserve the intrinsic structure of the tensor.

Recently, the t-SVD and the corresponding tensor tubal rank (and multi rank) have received considerable attentions \cite{2013Kilmer,zhang2017exact,JilogDet,ZhangCVPR,jiang2017novel2, zhou2017outlier, Lu2016, HutTNN,zhou2018tensor,lu2018tensor, xie2016unifying, bibi2017high,xue2017low,2014Semerci}.
As a generalization of the matrix singular value decomposition (SVD), the t-SVD regards a three-way tensor $\mathcal{X}$ as a matrix, each element of which is a tube (mode-3 fiber), and then decomposes $\mathcal{X}$ as \begin{equation*}
\begin{aligned}
    \mathcal{X}=\mathcal{U}\ast \mathcal{S}\ast \mathcal{V}^{\mathrm{T}},\\
\end{aligned}
\end{equation*}
where $\mathcal{U}$ and $\mathcal{V}$ are orthogonal tensors, $\mathcal{S}$ is a f-diagonal tensor, $\mathcal{V}^{\mathrm{T}}$ denotes the conjugate transpose of $\mathcal{V}$, and $\ast$ denote the t-product (see details in Section \ref{Not}).
Mathematically, this decomposition is equivalent to a series of matrix SVDs in the Fourier domain \cite{zhang2017exact},
i.e.,
\begin{equation*}
\begin{aligned}
    \bar{X}^{(i)}=\bar{U}^{(i)}\bar{S}^{(i)} (\bar{V}^{(i)})^{\mathrm{T}},~~i=1,2,\cdots,n_3,\\
\end{aligned}
\end{equation*}
where $\bar{X}^{(i)}$ is the $i$-th frontal slice of $\mathcal{\bar{X}}$, and $\mathcal{\bar{X}}$ is generated by performing the Discrete Fourier Transformation (DFT) along each tube of $\mathcal{X}$.
The multi rank of $\mathcal{X}$ is defined as a vector whose $i$-th element is the rank of $\bar{X}^{(i)}$, i.e.,
\begin{equation*}
\begin{aligned}
    \text{rank}_{\text{m}}(\mathcal{X}):=\big(\text{rank}(\bar{X}^{(1)}), \text{rank}(\bar{X}^{(2)}),\cdots,\text{rank}(\bar{X}^{(n_3)}\big).
\end{aligned}
\end{equation*}
The tubal rank of $\mathcal{X}$ is defined as the number of non-zero tubes of $\mathcal{S}$, i.e.,
\begin{equation*}
\begin{aligned}
    \text{rank}_{\text{t}}(\mathcal{X}):=\#\{i:\mathcal{S}(i,i,:)\neq0\}.
\end{aligned}
\end{equation*}
Especially, the tensor tubal rank is equal to the maximum value of tensor multi rank.
Since directly minimizing the tensor tubal/multi rank is NP-hard \cite{Hillar2009Most}, Semerci et al. \cite{2014Semerci} developed a tensor nuclear norm (TNN) as their convex surrogate, i.e.,
\begin{equation*}
\|\mathcal{X}\|_{\text{TNN}}:=\sum_{i=1}^{n_3}\|\bar{X}^{(i)}\|_{*}.
\end{equation*}
Then, Zhang et al. \cite{ZhangCVPR,zhang2017exact} proposed the TNN-based LRTC and TRPCA models, Lu et al. \cite{Lu2016} further proved the exactly-recover-property for TNN-based TRPCA model, and Hu et al. \cite{HutTNN} proposed a twist tensor nuclear norm (t-TNN) for the video completion.

\begin{table}[t]
\scriptsize
\setlength{\tabcolsep}{8pt}
\renewcommand\arraystretch{1.2}
\caption{The rank estimation of two HSIs.}
\begin{center}
\begin{tabular}{c|c|c|c}
 \Xhline{1pt}
Data  &Size & Tucker rank  &  $N$-tubal rank\\
 \hline
 \emph{Washington DC Mall}    & $256\times 256\times150$      &(107,110,6)    & (182,8,8) \\
  \hline
  \emph{Pavia University}      & $256\times 256\times87$      &(115,119,7)    & (137,8,8) \\
 \Xhline{1pt}
\end{tabular}\vspace{-0.5cm}
\end{center}
\label{rank}
\end{table}

TNN has shown its effectiveness to preserve the intrinsic structure of the tensor \cite{ZhangCVPR,zhang2017exact,HutTNN}. However, TNN has two obvious shortcomings, one is its inapplicability for $N$-way tensors ($N>3$),
and another is its inflexibility for handling different correlations along different modes, especially the third mode. Specifically, under the framework of t-SVD, for a three-way tensor, the correlations along the first and the second mode are characterized by the SVD while that along the third mode is encoded by the embedded circular convolution \cite{zhang2017exact,lu2018tensor}. And most of real-world data are always with different correlations along each of its modes, e.g., the correlation of a HSI along its spectral mode should be much stronger than those along its spatial modes. Thus, treating each mode flexibly like SNN is expected to compensate for this defect.

To overcome the above two issues, in this paper, we define a three-way extension of the tensor matricization operator, named mode-$k_1k_2$ tensor unfolding ($k_1<k_2$), as the process of lexicographically stacking the mode-$k_1k_2$ slices of an $N$-way tensor $\mathcal{X} \in \mathbb{R}^{n_1\times n_2 \times \cdots \times n_N}$ into the frontal slices of a three-way tensor $\mathcal{X}_{(k_1k_2)}\in \mathbb{R}^{n_{k_1}\times n_{k_2}\times \prod_{s\neq k_1,k_2}n_s}$ (see details in Section \ref{secmodel}). Mathematically, the $(i_1, i_2,\cdots,i_N)$-th element of $\mathcal{X}$ maps to the $(i_{k_1},i_{k_2},j)$-th element of $\mathcal{X}_{(k_1k_2)}$, where
$$j=1+\sum_{\substack{s=1\\s\neq k_1,s\neq k_2}}^N (i_s-1)J_s~~\text{with}~~J_s=\prod_{\substack{m=1\\m\neq k_1, m\neq k_2}}^{s-1}n_m.$$
To character the correlations along different modes in a more flexible manner, we propose a new tensor rank, termed as tensor $N$-tubal rank, which is defined as a vector, whose elements contain the tubal rank of all mode-$k_1k_2$ unfolding tensors, i.e.,
\begin{equation*}
\begin{aligned}
N\text{-rank}_\text{t}(\mathcal{X})=\big(\text{rank}_\text{t}(\mathcal{X}_{(12)}),&\text{rank}_\text{t}(\mathcal{X}_{(13)}),\cdots,\text{rank}_\text{t}(\mathcal{X}_{(1N)}),\\&
\text{rank}_\text{t}(\mathcal{X}_{(23)}),\cdots,\text{rank}_\text{t}(\mathcal{X}_{(2N)}),\\&~~~~~~~~~~~~~~~~~~~~~\cdots,\text{rank}_\text{t}(\mathcal{X}_{(N-1N)})\big)\in \mathbb{R}^{N(N-1)/2}.
\end{aligned}
\end{equation*}
Table \ref{rank} gives the rank estimation\footnote{The rank is approximated by the numbers of the singular values which are larger than 0.01 of the largest one.} of two HSIs. As observed, the rank of the mode-$3$ unfolding matrix is much lower than the size of the third mode, which implies a strong correlation along the third mode. According to the tensor $N$-tubal rank, such a strong correlation is inadequately depicted by the first element (the tubal rank), while can be exactly depicted by the other two elements. This observation from Table \ref{rank} demonstrates that compared with the tensor tubal rank, the novel tensor $N$-tubal rank has the advantage of flexible and simultaneous depiction for the correlations along different modes.

To efficiently minimize the proposed tensor $N$-tubal rank, we design its convex relaxation: the weighted sum of tensor nuclear norm (WSTNN), which can be expressed as the weighted sum of the TNN of each mode-$k_1k_2$ unfolding tensor, i.e.,
$$\|\mathcal{X}\|_{\text{WSTNN}}:=\sum_{1\leq k_1<k_2\leq N}\alpha_{k_1k_2}\|\mathcal{X}_{(k_1k_2)}\|_{\text{TNN}},$$
where $\alpha_{k_1k_2} \geq 0~(1\leq k_1<k_2\leq N,k_1,k_2\in\mathbb{Z})$ and $\sum_{1\leq k_1<k_2\leq N}\alpha_{k_1k_2}=1$.
Then, we apply the WSTNN to two typical LRTR problems, LRTC and TRPCA problems, and propose the WSTNN-based LRTC and TRPCA models. Meanwhile, two efficient alternating direction method of multipliers (ADMM)-based algorithms \cite{Zhao2013TGRS,ma2017truncated} are developed to solve the proposed models. Numerous numerical experiments on synthetic and real-world data are conducted to illustrate the effectiveness and the efficiency of the proposed methods.

The rest of this paper is organized as follows. Section \ref{Not} presents some preliminary knowledge. Section \ref{secmodel} gives the definitions of tensor $N$-tubal rank and its convex surrogate WSTNN. Section \ref{LRTCandRCA} proposes the WSTNN-based LRTC and TRPCA models and develops two efficient ADMM-based solvers. Section \ref{sec:NE} evaluates the performance of the proposed models and compares the results with state-of-the-art competing methods. Section \ref{sec:Con} concludes this paper.

\section{Notations and preliminaries}\label{Not}

In this section, we give the basic notations and briefly introduce some definitions that will be used in this work \cite{kolda2009tensor,zhang2017exact}.

We denote vectors as bold lowercase letters (e.g., $\mathbf{x}$), matrices as uppercase letters (e.g., $X$), and tensors as calligraphic letters (e.g., $\mathcal{ X}$). For a three-way tensor $\mathcal{X}\in \mathbb{R}^{n_1\times n_2\times n_3}$, with the MATLAB notation, we denote its $(i,j,s)$-th element as $\mathcal{X}(i,j,s)$ or $\mathcal{X}_{i,j,s}$, its $(i,j)$-th mode-1, mode-2, and mode-3 fibers as $\mathcal{X}(:,i,j)$, $\mathcal{X}(i,:,j)$, and $\mathcal{X}(i,j,:)$, respectively. We use $ \mathcal{X}(i,:,:)$, $\mathcal{X}(:,i,:)$, and $\mathcal{X}(:,:,i)$ to denote the $i$-th horizontal, lateral, and frontal slices of $\mathcal{X}$, respectively.
More compactly, $X^{(i)}$ is used to represent $\mathcal{X}(:,:,i)$.
The Frobenius norm of $\mathcal{ X}$ is defined as $\|\mathcal{X}\|_F:=(\sum_{i,j,s}|\mathcal{X}(i,j,s)|^2)^{1/2}$.
The $\ell_1$ norm of $\mathcal{ X}$ is defined as $\|\mathcal{X}\|_1:=\sum_{i,j,s}|\mathcal{X}(i,j,s)|$.
We use $\bar{\mathcal{X}}$ to denote the tensor generated by performing DFT along each tube of $\mathcal{X}$, i.e., $\mathcal{\bar{X}}={\tt{fft}}(\mathcal{X},[],3)$. Naturally, we can compute $\mathcal{X}$ via $\mathcal{X}={\tt{ifft}}(\mathcal{\bar{X}},[],3)$.

The vectorization of an $N$-way tensor $\mathcal{X} \in \mathbb{R}^{n_1\times n_2 \times \cdots \times n_N}$, denote as $\mathbf{x}={\tt vec}(\mathcal{X})\in\mathbb{R}^{n_1n_2\cdots n_N}$, is defined as
$$
\mathbf{x}(j)=\mathcal{X}(i_1,i_2,\cdots,i_N)~\text{with}~j=i_1 +\sum_{s=2}^N\bigg((i_s-1)\prod_{m=1}^{s-1}n_m\bigg).
$$

The mode-$k$ tensor matricization of an $N$-way tensor $\mathcal{X}\in \mathbb{R}^{n_1\times n_2 \times\cdots \times n_N}$ is denoted as $X_{(k)}\in \mathbb{R}^{n_k\times \prod_{s\neq k}n_s}$, whose $(i_k, j)$-th element maps to the $(i_1,i_2,\cdots,i_N)$-th element of $\mathcal{X}$, where
$$j = 1 +\sum_{s=1,s\neq k}^N(i_s-1)J_s~\text{with}~J_s=\prod_{m=1,m\neq k}^{s-1}n_m.$$
The corresponding operator and inverse operator is denoted as ``$\tt unfold$'' and ``$\tt fold$'', i.e., $ X_{(k)}= {\tt unfold}(\mathcal{X},k)$ and $\mathcal{X} = {\tt fold}(X_{(k)},k)$

For a three-way tensor $\mathcal{X}\in \mathbb{R}^{n_1\times n_2\times n_3}$,
the block circulation operation is defined as
\begin{equation*}
\setlength{\arraycolsep}{0.6pt}
\begin{aligned}
&\mathbf{\tt bcirc}(\mathcal{X}):=
\begin{pmatrix}
X^{(1)}   &   X^{(n_3)}  &\ldots~&   X^{(2)} \\
X^{(2)}   &   X^{(1)}    &\ldots~&   X^{(3)} \\
\vdots        &   \vdots         &\ddots~&   \vdots      \\
X^{(n_3)} &   X^{(n_3-1)}  &\ldots~&   X^{(1)} \\
\end{pmatrix}\in \mathbb{R}^{n_1n_3\times n_2n_3}.\\
\end{aligned}
\end{equation*}
The block diagonalization operation and its inverse operation are defined as
\begin{equation*}
\setlength{\arraycolsep}{0.6pt}
\begin{aligned}
\mathbf{\tt bdiag}(\mathcal{X}):=&
\begin{pmatrix}
X^{(1)}   &                    &      &               \\
              &   X^{(2)}      &      &               \\
              &                    &\ddots~&          \\
              &                    &      & X^{(n_3)} \\
\end{pmatrix}\in \mathbb{R}^{n_1n_3\times n_2n_3},~~\mathbf{\tt bdfold}\big(\mathbf{\tt bdiag}(\mathcal{X})\big):=\mathcal{X}.\\
\end{aligned}
\end{equation*}
The block vectorization operation and its inverse operation are defined as
\begin{equation*}
\setlength{\arraycolsep}{0.3pt}
\begin{aligned}
\mathbf{\tt bvec}&(\mathcal{X}):=
\begin{pmatrix}
(X^{(1)})^{\text{T}},&
(X^{(2)})^{\text{T}},&
\cdots,&
(X^{(n_3)})^{\text{T}}
\end{pmatrix}^{\text{T}}\in \mathbb{R}^{n_1n_3 \times n_2},~~\mathbf{\tt bvfold}\big(\mathbf{\tt bvec}(\mathcal{X})\big):=\mathcal{X}.
\end{aligned}
\end{equation*}

\begin{figure}[t]
\begin{center}
\includegraphics[width=0.7\textwidth]{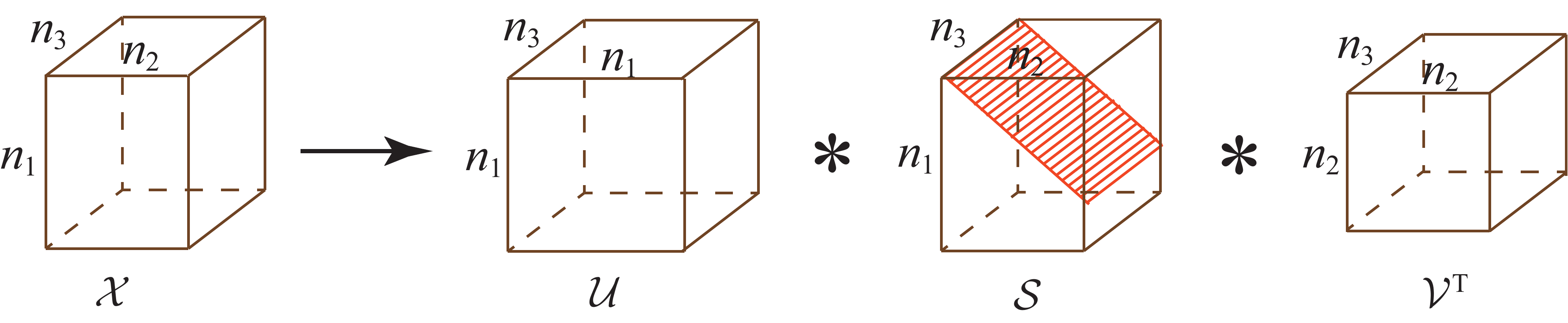}
\vspace{-0.2cm}
\caption{Illustration of the t-SVD of an $n_1\times n_2\times n_3$ tensor.}\label{modetSVD}
\end{center}\vspace{-0.2cm}
\end{figure}

\begin{algorithm}[t]
\renewcommand\arraystretch{1.2}
\caption{The t-SVD for three-way tensors}
\begin{algorithmic}[1]
\renewcommand{\algorithmicrequire}{\textbf{Input:}} 
\renewcommand{\algorithmicensure}{\textbf{Output:}}
\Require
$\mathcal{X}\in\mathbb{R}^{n_1\times n_2\times n_3}$.
\State $\bar{\mathcal{X}}\leftarrow$ $\tt{fft}$$(\mathcal{X},[],3)$.
\For {$i=1$ to $n_3$}
\State $[ U, S, V]=$ $\tt{svd}$$(\bar{X}^{(i)})$.
\State $\bar{U}^{(i)}\leftarrow U$; $\bar{S}^{(i)}\leftarrow S$;
$\bar{V}^{(i)}\leftarrow V$.
\EndFor
\State
$\mathcal{U}\leftarrow$ $\tt{ifft}$$(\bar{\mathcal{U}},[],3)$.\\
$\mathcal{S}\leftarrow$ $\tt{ifft}$$(\bar{\mathcal{S}},[],3)$.\\
$\mathcal{V}\leftarrow$ $\tt{ifft}$$(\bar{\mathcal{V}},[],3)$.
\Ensure
$\mathcal{U}$,
$\mathcal{S}$,
$\mathcal{V}$.
\end{algorithmic}
\label{t-svd}
\end{algorithm}

\begin{myDef}[t-product]
\emph{The t-product between two three-way tensors $\mathcal{X}\in \mathbb{R}^{n_1\times n_2\times n_3}$ and $\mathcal{Y}\in \mathbb{R}^{n_2\times n_4\times n_3}$ is defined as
$$\mathcal{X}\ast \mathcal{Y}:=\mathrm{\tt bvfold}\big(\mathrm{\tt bcirc}(\mathcal{X}) \mathrm{\tt bvec}(\mathcal{Y})\big)\in \mathbb{R}^{n_1\times n_4\times n_3}.$$}
\end{myDef}

Indeed, the t-product can be regarded as a matrix-matrix multiplication, except that the multiplication operation between scalars is replaced by circular convolution between the tubes, i.e.,
\begin{equation*}
\begin{aligned}
&\mathcal{F}=\mathcal{X}\ast \mathcal{Y} \Leftrightarrow \mathcal{F}(i,j,:)=\sum_{t=1}^{n_2}\mathcal{X}(i,t,:)\star\mathcal{Y}(t,j,:),
\end{aligned}
\end{equation*}
where $\star$ denotes the circular convolution between two tubes. Note that the circular convolution in the spatial domain is equivalent to the multiplication in the Fourier domain, the t-product between two tensors $\mathcal{F}=\mathcal{X}\ast \mathcal{Y}$ is equivalent to
\begin{equation*}
\begin{aligned}
\bar{\mathcal{F}}=\mathcal{\tt bdfold}\big(\mathcal{\tt bdiag}(\bar{\mathcal{X}}) \mathcal{\tt bdiag}(\bar{\mathcal{Y}})\big).
\end{aligned}
\end{equation*}

\begin{myDef}[special tensors]\emph{The conjugate transpose of a three-way tensor $\mathcal{X}\in \mathbb{R}^{n_1\times n_2\times n_3}$, denote as $\mathcal{X}^{\mathrm{T}}$, is the tensor obtained by conjugate transposing each of the frontal slices and then reversing the order of transposed frontal slices 2 through $n_3$.} \emph{The identity tensor $\mathcal{I} \in \mathbb{R}^{n_1\times n_2\times n_3}$ is the tensor whose first frontal slice is the identity matrix, and other frontal slices are all zeros.} \emph{A three-way tensor $\mathcal{Q}$ is orthogonal if $\mathcal{Q}\ast \mathcal{Q}^{\mathrm{T}}=\mathcal{Q}^{\mathrm{T}}\ast \mathcal{Q}=\mathcal{I}.$}
\emph{A three-way tensor $\mathcal{S}$ is f-diagonal if each of its frontal slices is a diagonal matrix.}
\end{myDef}

\begin{theorem}[t-SVD]
\emph{Let $\mathcal{X}\in \mathbb{R}^{n_1\times n_2\times n_3}$ is a three-way tensor, then it can be factored as
$$\mathcal{X}=\mathcal{U}\ast \mathcal{S}\ast \mathcal{V}^{\mathrm{T}},$$
where $\mathcal{U}\in \mathbb{R}^{n_1\times n_1\times n_3}$ and $\mathcal{V}\in \mathbb{R}^{n_2\times n_2\times n_3}$ are the orthogonal tensors, and $\mathcal{S}\in \mathbb{R}^{n_1\times n_2\times n_3}$ is a f-diagonal tensor.}
\end{theorem}

Fig. \ref{modetSVD} illustrates the t-SVD scheme. The t-SVD can be efficiently obtained by computing a
series of matrix SVDs in the Fourier domain; see Algorithm \ref{t-svd}. Now, we give the definition of the corresponding tensor tubal rank and multi rank.

\begin{myDef}[tensor tubal rank and multi rank]\label{Def:tubal}
\emph{Let $\mathcal{X}\in\mathbb{R}^{n_1\times n_2\times n_3}$ be a three-way tensor, the tensor multi rank of $\mathcal{X}$ is a vector $\text{rank}_{\text{m}}(\mathcal{X}) \in\mathbb{R}^{n_3}$, whose $i$-th element is the rank of $i$-th frontal slice of $\bar{\mathcal{X}}$, where $\bar{\mathcal{X}}=\mathrm{\tt fft}(\mathcal{X},[],3)$.
The tubal rank of $\mathcal{X}$, denote as $\text{rank}_{\text{t}}(\mathcal{X})$, is defined as the number of non-zero tubes of $\mathcal{S}$, where $\mathcal{S}$ comes from the t-SVD of $\mathcal{X}$: $\mathcal{X}=\mathcal{U}\ast \mathcal{S}\ast \mathcal{V}^{\mathrm{T}}$.
That is, $\text{rank}_{\text{t}}(\mathcal{X})=\max\big(\text{rank}_{\text{m}}(\mathcal{X})\big)$.}
\end{myDef}

\begin{myDef}[tensor nuclear norm (TNN)]\label{kTNN}
\emph{The tensor nuclear norm of a tensor $\mathcal{X}\in \mathbb{R}^{n_{1}\times n_2\times n_{3}}$, denoted as $\|\mathcal{X}\|_{\text{TNN}}$, is defined as the sum of singular values of all the frontal slices of $\bar{\mathcal{X}}$, i.e.,
\begin{equation*}
\|\mathcal{X}\|_{\text{TNN}}:=\sum_{i=1}^{n_3}\big\|\bar{X}^{(i)}\big\|_{*}.
\end{equation*}
where $\bar{X}^{(i)}$ is the $i$-th frontal slice of $\mathcal{\bar{X}}$, and $\bar{\mathcal{X}}=\mathrm{\tt fft}(\mathcal{X},[],3)$.}
\end{myDef}

\section{Tensor \emph{N}-tubal rank and convex relaxation}\label{secmodel}

In this section, we first propose the mode-$k_1k_2$ tensor unfolding operation, then give the definitions of tensor $N$-tubal rank and its convex relaxation WSTNN.

Slices of a tensor are its two-dimensional sections, as the three-way tensors can be directly visualized in three-dimensional space, scholars commonly and visually denote their three kinds of slices as horizontal, lateral, and frontal slices, respectively. However, this naming method is not feasible for higher-way tensors, such as four-way tensors. Therefore, we give the following definition for the slices of $N$-way tensors.

\begin{myDef}[mode-$\bm{k_1k_2}$ slices]\label{tslice}
\emph{The mode-$k_1k_2$ slices ($X^{k_1k_2}$, $1\leq k_1<k_2\leq N,k_1,k_2\in\mathbb{Z}$) of an $N$-way tensor $\mathcal{X} \in \mathbb{R}^{n_1\times n_2 \times \cdots \times n_N}$ are its two-dimensional sections, defined by fixing all but the mode-$k_1$ and the mode-$k_2$ indexes.}
\end{myDef}

For example, for a four-way tensor $\mathcal{X}\in \mathbb{R}^{2\times 3 \times 3\times 2}$, its $(i_2,i_4)$-th mode-$13$ slice and $(i_1,i_3)$-th mode-24 slice are
$$
\setlength{\arraycolsep}{0.6pt}
X^{13}=
\begin{pmatrix}
\mathcal{X}(\mathbf{1},i_2,\mathbf{1},i_4)~&~\mathcal{X}(\mathbf{1},i_2,\mathbf{2},i_4)~&~\mathcal{X}(\mathbf{1},i_2,\mathbf{3},i_4)  \\
\mathcal{X}(\mathbf{2},i_2,\mathbf{1},i_4)~&~\mathcal{X}(\mathbf{2},i_2,\mathbf{2},i_4)~&~\mathcal{X}(\mathbf{2},i_2,\mathbf{3},i_4)  \\
\end{pmatrix}~\text{and}~
X^{24}=
\begin{pmatrix}
\mathcal{X}(i_1,\mathbf{1},i_3,\mathbf{1})~&~\mathcal{X}(i_1,\mathbf{1},i_3,\mathbf{2})\\
\mathcal{X}(i_1,\mathbf{2},i_3,\mathbf{1})~&~\mathcal{X}(i_1,\mathbf{2},i_3,\mathbf{2})\\
\mathcal{X}(i_1,\mathbf{3},i_3,\mathbf{1})~&~\mathcal{X}(i_1,\mathbf{3},i_3,\mathbf{2})\\
\end{pmatrix},$$
respectively. With the definition of mode-$k_1k_2$ slices, we define the following mode-$k_1k_2$ tensor unfolding.

\begin{myDef}[mode-$\bm{k_1k_2}$ tensor unfolding]\label{tunfold}
\emph{For an $N$-way tensor $\mathcal{X} \in \mathbb{R}^{n_1\times n_2 \times \cdots \times n_N}$, its mode-$k_1k_2$ unfolding is a three-way tensor denoted by $\mathcal{X}_{(k_1k_2)}\in \mathbb{R}^{n_{k_1}\times n_{k_2}\times \prod_{s\neq k_1,k_2}n_s}$, whose frontal slices are the lexicographic ordering of the mode-$k_1k_2$ slices of $\mathcal{X}$. Mathematically, the $(i_1, i_2,\cdots,i_N)$-th element of $\mathcal{X}$ maps to the $(i_{k_1},i_{k_2},j)$-th element of $\mathcal{X}_{(k_1k_2)}$, where
$$j=1+\sum_{\substack{s=1\\s\neq k_1,s\neq k_2}}^N (i_s-1)J_s~~\text{with}~~J_s=\prod_{\substack{m=1\\m\neq k_1, m\neq k_2}}^{s-1}n_m.$$
We define the corresponding operation as $\mathcal{X}_{(k_1k_2)}:={\tt t\text{-}unfold}(\mathcal{X},k_1,k_2)$, and its inverse operation as $\mathcal{X}:={\tt t\text{-}fold}(\mathcal{X}_{(k_1k_2)},k_1,k_2)$ \footnote{By using the Matlab commands, we can compute $\mathcal{X}_{(k_1k_2)}$ via $$\mathcal{X}_{(k_1k_2)}={\tt{reshape}}\big({\tt{permute}}\big(\mathcal{X},[k_1,k_2,{\tt setdiff}([1:N],[k_1,k_2])]\big),dim(k_1),dim(k_2),[]\big),$$ and compute $\mathcal{X}$ via $$\mathcal{X}={\tt{ipermute}}\big({\tt{reshape}}\big(\mathcal{X}_{(k_1k_2)},dim([k_1,k_2,{\tt setdiff}([1:N],[k_1,k_2])])\big),[k_1,k_2,{\tt setdiff}([1:N],[k_1,k_2])]\big),$$ where  $dim=\tt{size}(\mathcal{X})$ and $N=\tt{ndims}(\mathcal{X})$.}.
}
\end{myDef}

\begin{figure}[t]
\begin{center}
\includegraphics[width=0.85\textwidth]{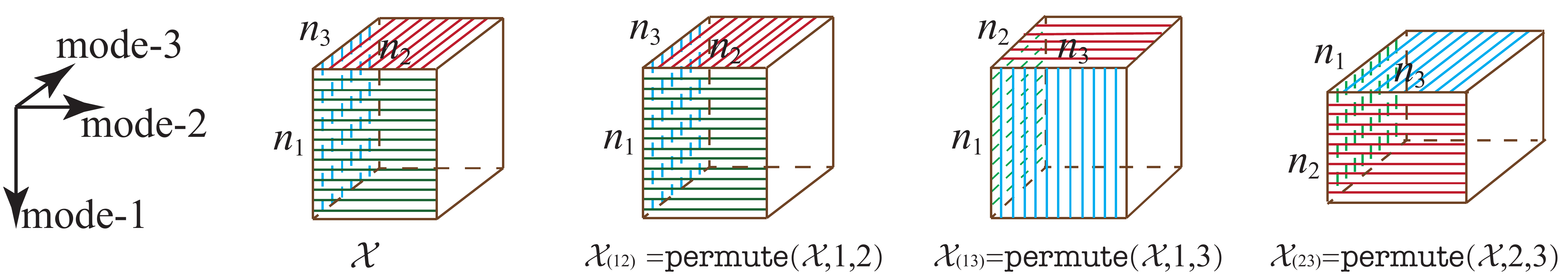}\vspace{-0.2cm}
\caption{Illustration the permutation tensor of an $n_1\times n_2\times n_3$ tensor.}\label{Permute}\vspace{-0.3cm}
\end{center}
\end{figure}

For example, for a four-way tensor $\mathcal{X}\in \mathbb{R}^{2\times 3 \times 3\times 2}$, its mode-$24$ unfolding tensor $\mathcal{X}_{(24)}\in \mathbb{R}^{3\times2\times6}$ can be expressed as
\begin{equation*}
\begin{aligned}
\setlength{\arraycolsep}{0.6pt}
\mathcal{X}_{(24)}(:,:,1)=
\begin{pmatrix}
\mathcal{X}(1,\mathbf{1},1,\mathbf{1})~&~\mathcal{X}(1,\mathbf{1},1,\mathbf{2})\\
\mathcal{X}(1,\mathbf{2},1,\mathbf{1})~&~\mathcal{X}(1,\mathbf{2},1,\mathbf{2})\\
\mathcal{X}(1,\mathbf{3},1,\mathbf{1})~&~\mathcal{X}(1,\mathbf{3},1,\mathbf{2})\\
\end{pmatrix},~
\mathcal{X}_{(24)}(:,:,4)=
\begin{pmatrix}
\mathcal{X}(1,\mathbf{1},2,\mathbf{1})~&~\mathcal{X}(1,\mathbf{1},2,\mathbf{2})\\
\mathcal{X}(1,\mathbf{2},2,\mathbf{1})~&~\mathcal{X}(1,\mathbf{2},2,\mathbf{2})\\
\mathcal{X}(1,\mathbf{3},2,\mathbf{1})~&~\mathcal{X}(1,\mathbf{3},2,\mathbf{2})\\
\end{pmatrix},\\
\setlength{\arraycolsep}{0.6pt}
\mathcal{X}_{(24)}(:,:,2)=
\begin{pmatrix}
\mathcal{X}(2,\mathbf{1},1,\mathbf{1})~&~\mathcal{X}(2,\mathbf{1},1,\mathbf{2})\\
\mathcal{X}(2,\mathbf{2},1,\mathbf{1})~&~\mathcal{X}(2,\mathbf{2},1,\mathbf{2})\\
\mathcal{X}(2,\mathbf{3},1,\mathbf{1})~&~\mathcal{X}(2,\mathbf{3},1,\mathbf{2})\\
\end{pmatrix},~
\mathcal{X}_{(24)}(:,:,5)=
\begin{pmatrix}
\mathcal{X}(2,\mathbf{1},2,\mathbf{1})~&~\mathcal{X}(2,\mathbf{1},2,\mathbf{2})\\
\mathcal{X}(2,\mathbf{2},2,\mathbf{1})~&~\mathcal{X}(2,\mathbf{2},2,\mathbf{2})\\
\mathcal{X}(2,\mathbf{3},2,\mathbf{1})~&~\mathcal{X}(2,\mathbf{3},2,\mathbf{2})\\
\end{pmatrix},\\
\setlength{\arraycolsep}{0.6pt}
\mathcal{X}_{(24)}(:,:,3)=
\begin{pmatrix}
\mathcal{X}(3,\mathbf{1},1,\mathbf{1})~&~\mathcal{X}(3,\mathbf{1},1,\mathbf{2})\\
\mathcal{X}(3,\mathbf{2},1,\mathbf{1})~&~\mathcal{X}(3,\mathbf{2},1,\mathbf{2})\\
\mathcal{X}(3,\mathbf{3},1,\mathbf{1})~&~\mathcal{X}(3,\mathbf{3},1,\mathbf{2})\\
\end{pmatrix},~
\mathcal{X}_{(24)}(:,:,6)=
\begin{pmatrix}
\mathcal{X}(3,\mathbf{1},2,\mathbf{1})~&~\mathcal{X}(3,\mathbf{1},2,\mathbf{2})\\
\mathcal{X}(3,\mathbf{2},2,\mathbf{1})~&~\mathcal{X}(3,\mathbf{2},2,\mathbf{2})\\
\mathcal{X}(3,\mathbf{3},2,\mathbf{1})~&~\mathcal{X}(3,\mathbf{3},2,\mathbf{2})\\
\end{pmatrix}.
\end{aligned}
\end{equation*}

\begin{figure}[!t]
\begin{center}
\includegraphics[width=0.95\textwidth]{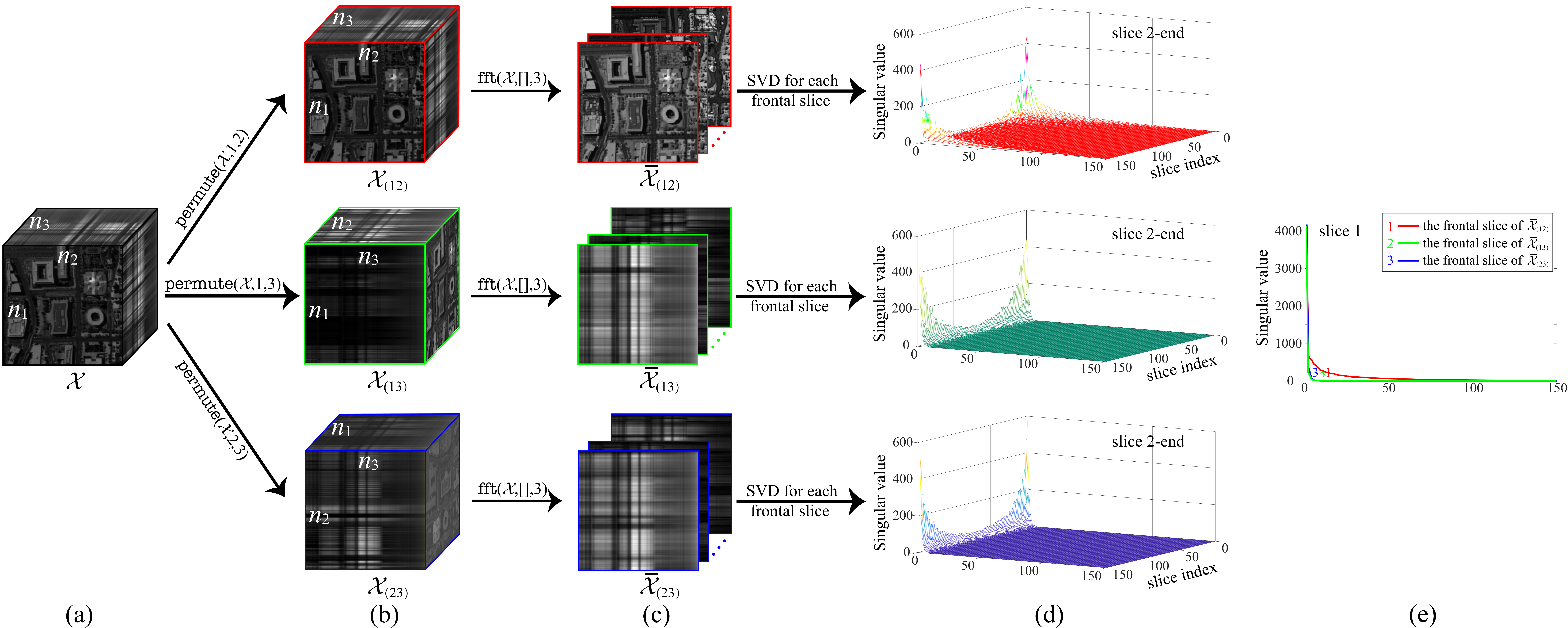}
\caption{Illustration of the low $N$-tubal rank prior of a HSI. (a) The HSI \emph{Washington DC Mall} of size $150\times150\times150$. (b) The mode-$k_1k_2$ permutation tensors of $\mathcal{X}$.
(c) The generated tensors $\bar{\mathcal{X}}_{(k_1k_2)}$ by performing the DFT along each tube of $\mathcal{X}_{(k_1k_2)}$. (d) Singular value curves from the second to the end frontal slices of $\bar{\mathcal{X}}_{(k_1k_2)}$. (e) Singular value curves of the first frontal slices of $\bar{\mathcal{X}}_{(k_1k_2)}$.}
\label{lowrank}
\end{center}
\vspace{-0.5cm}
\end{figure}

Especially, for a three-way tensor, the proporsed tensor unfolding operation does not involve dimensional-reduction, while corresponds to a permutation operation, i.e.,
$$\mathcal{X}(i,j,s)=\mathcal{X}_{(12)}(i,j,s)=\mathcal{X}_{(13)}(i,s,j)=\mathcal{X}_{(23)}(j,s,i).$$
Therefore, in this case, we use $\tt permute$ and $\tt ipermute$ to replace ${\tt t\text{-}unfold}$ and ${\tt t\text{-}fold}$, respectively.
Fig. \ref{Permute} illustrates the mode-$k_1k_2$ permutation (unfolding) tensor of an $n_1\times n_2\times n_3$ tensor.

As pointed out in Section \ref{sec:Int}, the framework of t-SVD and the corresponding tubal rank are applicable only to three-way tensors and lack of flexibility to handle different correlations along different modes.
To handle these two issues, we perform the t-SVD on each mode-$k_1k_2$ unfolding tensor and propose the following tensor $N$-tubal rank.

\begin{myDef}[$\bm{N}$-tubal rank]\label{Ntubal}
\emph{The $N$-tubal rank of an $N$-way tensor $\mathcal{X} \in \mathbb{R}^{n_1\times n_2 \times \cdots \times n_N}$ is defined as a vector, whose elements contain the tubal rank of all mode-$k_1k_2$ unfolding tensors, i.e.,
\begin{equation*}
\begin{aligned}
N\text{-rank}_\text{t}(\mathcal{X})=\big(\text{rank}_\text{t}(\mathcal{X}_{(12)}),&\text{rank}_\text{t}(\mathcal{X}_{(13)}),\cdots,\text{rank}_\text{t}(\mathcal{X}_{(1N)}),\\&
\text{rank}_\text{t}(\mathcal{X}_{(23)}),\cdots,\text{rank}_\text{t}(\mathcal{X}_{(2N)}),\\&~~~~~~~~~~~~~~~~~~~~~\cdots,\text{rank}_\text{t}(\mathcal{X}_{(N-1N)})\big)\in \mathbb{R}^{N(N-1)/2}.
\end{aligned}
\end{equation*}}
\end{myDef}

Clearly, for a three-way tensor, the tensor tubal rank is actually the first element of the tensor $N$-tubal rank.
By taking the HSI \emph{Washington DC Mall} shown in Fig. \ref{lowrank} as an example, the low $N$-tubal rank prior of it can be evidently observed both quantitatively and visually. Especially, the proposed $N$-tubal rank combines the advantages of Tucker rank and tubal rank. On the one hand, compared with the mode-$k_1$ unfolding matrix, the mode-$k_1k_2$ unfolding tensor avoids the destruction to the
structures information along the $k_2$-th mode. On the other hand, as shown in Fig. \ref{lowrank}, the tubal rank of each mode-$k_1k_2$ unfolding (permutation) tensor $\mathcal{X}_{(k_1k_2)}$ more directly depicts the correlation of the the $k_1$-th and the $k_2$-th modes, i.e., it lacks of direct characterization for the correlation along other modes. As all mode-$k_1k_2$ unfolding tensors are considered simultaneously, the proposed $N$-tubal rank can effectively exploit the correlations along all modes. The following theorem reveals the relationship between tensor $N$-tubal rank and CP rank.

\begin{theorem}[$\bm{N}$-tubal rank and CP rank]\label{CPfiber} \emph{Assuming the CP rank of an $N$-way tensor $\mathcal{X}\in \mathbb{R}^{n_1\times n_2 \times \cdots \times n_N}$ is $r$, and its CP decomposition is
\begin{equation*}
\mathcal{X}=\sum_{i=1}^r \mathbf{a}_i^{1} \circ \mathbf{a}_i^{2} \circ  \cdots \circ \mathbf{a}_i^{N},~\mathbf{a}_i^{k}\in \mathbb{R}^{n_k}, k=1,2,\cdots,N.
\end{equation*}
Then the $N$-tubal rank of $\mathcal{X}$ is at most $r\times {\tt ones}(N(N-1)/2,1)$\footnote{${\tt ones}(n,1)\in \mathbb{R}^{n}$ is a vector whose elements are all 1\label{ones}.}.
Especially, we define vector sets
\begin{equation*}
\begin{matrix}
\mathbb{V}_1=\{\mathbf{a}_i^{1}|i=1,2,\cdots,r\},\\
\mathbb{V}_2=\{\mathbf{a}_i^{2}|i=1,2,\cdots,r\},\\
\vdots\\
\mathbb{V}_N=\{\mathbf{a}_i^{N}|i=1,2,\cdots,r\},
\end{matrix}
\end{equation*}
and
\begin{equation*}
\mathbf{c}_i={\tt vec}(\mathcal{C}_i)\in\mathbb{R}^{\prod_{s\neq k_1,k_2}n_s}, ~~i=1,2,\cdots,r,
\end{equation*}
where $\mathcal{C}_i=\mathbf{a}_i^{1} \circ \mathbf{a}_i^{2} \circ  \cdots \circ \mathbf{a}_i^{k_1-1} \circ \mathbf{a}_i^{k_1+1} \circ \cdots \circ \mathbf{a}_i^{k_2-1}\circ \mathbf{a}_i^{k_2+1}\circ \cdots \circ\mathbf{a}_i^{N}$. If each vector set $\mathbb{V}_i$ is linearly independent and each element of $\bar{\mathbf{c}}_i={\tt fft}(\mathbf{c}_i)$ is non-zero, the $N$-tubal rank is equal to $r\times {\tt ones}(N(N-1)/2,1)$.}
\end{theorem}

\textbf{Proof.} The $\mathcal{X}_{(k_1k_2)}$ has the following CP decomposition
\begin{equation*}
\mathcal{X}_{(k_1k_2)}=\sum_{i=1}^r \mathbf{a}_i^{k_1} \circ \mathbf{a}_i^{k_2} \circ \mathbf{c}_i,
\end{equation*}
Letting $\bar{\mathcal{X}}_{(k_1k_2)}={\tt fft}(\mathcal{X}_{(k_1k_2)},[],3)$, then $\bar{\mathcal{X}}_{(k_1k_2)}$ has the the following CP decomposition
\begin{equation*}
\bar{\mathcal{X}}_{(k_1k_2)}=\sum_{i=1}^r \mathbf{a}_i^{k_1} \circ \mathbf{a}_i^{k_2} \circ \bar{\mathbf{c}}_i,
\end{equation*}
where $\bar{\mathbf{c}}_i={\tt fft}(\mathbf{c}_i)$. Letting $\bar{\mathbf{c}}_i=(\bar{c}_i^1, \bar{c}_i^2\cdots \bar{c}_i^{d})$, then the $j$-th frontal slice of $\bar{\mathcal{X}}_{(k_1k_2)}$ can be expressed as
\begin{equation*}
\begin{aligned}
\bar{X}_{(k_1k_2)}^{(j)} = \bar{c}_1^j \mathbf{a}_1^{k_1} (\mathbf{a}_1^{k_2})^{\mathrm{T}}+\bar{c}_2^j \mathbf{a}_2^{k_1} (\mathbf{a}_2^{k_2})^{\mathrm{T}}+\cdots +\bar{c}_r^j \mathbf{a}_r^{k_1} (\mathbf{a}_r^{k_2})^{\mathrm{T}}.
\end{aligned}
\end{equation*}
This implies that the rank of each frontal slice of $\bar{\mathcal{X}}_{(k_1k_2)}$ is at most $r$, and it is equal to $r$ if the vector sets $\mathbb{V}_{k_1}$ or $\mathbb{V}_{k_2}$ is linearly independent and the $j$-th element of each $\bar{\mathbf{c}}_i$ is non-zero. Thus, the tubal rank of $\mathcal{X}_{(k_1k_2)}$ (the $(k_1,k_2)$-th element of the $N$-tubal rank of $\mathcal{X}$) is at most $r$, and it is equal to $r$ if the aforementioned conditions are satisfied. \hfill $\Box$ \vspace{0.3cm}

To effectively minimize the tensor $N$-tubal rank, we propose the following WSTNN as its convex relaxation.

\begin{myDef}[the weighted sum of tensor nuclear norm]\label{DeWSTNN}
\emph{The weighted sum of tensor nuclear norm (WSTNN) of an $N$-way tensor $\mathcal{X} \in \mathbb{R}^{n_1\times n_2 \times \cdots \times n_N}$, denote as $\|\mathcal{X}\|_{\text{WSTNN}}$, is defined as the weighted sum of the TNN of each mode-$k_1k_2$ unfolding tensor, i.e.,
$$\|\mathcal{X}\|_{\text{WSTNN}}:=\sum_{1\leq k_1<k_2\leq N}\alpha_{k_1k_2}\|\mathcal{X}_{(k_1k_2)}\|_{\text{TNN}},$$
where $\alpha_{k_1k_2} \geq 0~(1\leq k_1<k_2\leq N,k_1,k_2\in\mathbb{Z})$ and $\sum_{1\leq k_1<k_2\leq N}\alpha_{k_1k_2}=1$.}
\end{myDef}

The weight $\alpha=(\alpha_{11},\alpha_{12},\cdots,\alpha_{1N},\alpha_{23},\cdots,\alpha_{2N},\cdots,\alpha_{N-1N})$ is an important parameter for WSTNN. For the choice of the weight $\alpha$, we consider the following three cases.

\textbf{Case 1:} The tensor $N$-tubal rank of the underlying tensor is unknown and can not estimated empirically, such as the MRI data, the weight $\alpha$ is chosen to be
$$\alpha=\frac{(1,1,\cdots, 1)}{N(N-1)/2}=\frac{2(1,1,\cdots, 1)}{N(N-1)}.$$

\textbf{Case 2:} The tensor $N$-tubal rank of the underlying tensor $\mathcal{X}\in \mathbb{R}^{n_1\times n_2 \times \cdots \times n_N}$ is known, i.e.,
\begin{equation*}
\begin{aligned}
N\text{-rank}_\text{t}(\mathcal{X})=(r_{11},r_{12},\cdots,r_{1N},r_{23},\cdots,r_{2N},\cdots,r_{N-1N}).
\end{aligned}
\end{equation*}
Since $\alpha_{k_1k_2}$ stands for the contribution of the TNN of the mode-$k_1k_2$ unfolding tensor $\mathcal{X}_{({k_1k_2})}$, the value of $\alpha_{k_1k_2}$ should be dependent on the tubal rank of $\mathcal{X}_{({k_1k_2})}$ ($r_{k_1k_2}$) and the size of the first two modes of $\mathcal{X}_{({k_1k_2})}$ $(n_{k_1}~\text{and}~n_{k_2})$. Specially, larger (or smaller) ratio of $r_{k_1k_2}$ to $\min(n_{k_1},n_{k_2})$ corresponds to smaller (or larger) value of $\alpha_{k_1k_2}$. Therefore, the following strategy is considered to choose the weight $\alpha$, i.e.,
$$\alpha_{k_1k_2}=\frac{e^{\frac{\eta \hat{r}_{k_1k_2}}{R}}}{\sum_{1\leq k_1<k_2\leq N}e^{\frac{\eta \hat{r}_{k_1k_2}}{R}}},~\text{with}~
R=\sum_{1\leq k_1<k_2\leq N}\hat{r}_{k_1k_2},
~~1\leq k_1<k_2\leq N,k_1,k_2\in\mathbb{Z},$$
where $\hat{r}_{k_1k_2}=\frac{\min(n_{k_1},n_{k_2})-r_{k_1k_2}}{\min(n_{k_1},n_{k_2})}$ and $\eta$ is a balance parameter.

\textbf{Case 3:} Particularly, for HSIs/MSIs, although their exact $N$-tubal rank is unknown, the correlation along their spectral mode should be much stronger than those along
their spatial modes. It implies that the value of the first element of the $N$-tubal rank should be much larger than the values of the second and the third elements of it. Thus, in this case, we empirically choose the weights $\alpha$ as $(\theta ,1,1)/(2+\theta)$, and $\theta$ is a balance parameter.

\section{WSTNN-based models and solving algorithms}\label{LRTCandRCA}

In this section, we apply WSTNN to LRTC and TRPCA problems, and propose the WSTNN-based LRTC and WSTNN-based TRPCA models with ADMM-based solving schemes.

\subsection{WSTNN-based LRTC model}

Tensor completion aims at estimating the missing elements from an incomplete observation tensor. Considering an $N$-way tensor $\mathcal{X}\in \mathbb{R}^{n_1\times n_2\times \cdots\times n_N}$, the proposed WSTNN-based LRTC model is formulated as
\begin{equation}
\begin{aligned}
    \min_{\mathcal{X}}&~~\|\mathcal{X}\|_{\text{WSTNN}}\\
    \text{s.t.}&~~\mathcal{P}_{\Omega}(\mathcal{X}-\mathcal{F})=0,
\end{aligned} \label{3DTNNLRTC}
\end{equation}
where $\mathcal{X}$ is the underlying tensor, $\mathcal{F}$ is the observed tensor, $\Omega$ is the index set for the known entries, and $\mathcal{P}_{\Omega}(\mathcal{X})$ is the projection operator that keeps the entries of $\mathcal{X}$ in $\Omega$ and sets others to zero. Let
\begin{equation}
\iota_{\mathbb{S}}(\mathcal{X}):=
\left\{
\begin{aligned}
0,~~&\text{if}~\mathcal{X}\in \mathbb{S},\\
\infty,~&\text{otherwise},
\end{aligned}
\right.
\end{equation}
where $\mathbb{S}:=\{\mathcal{X}\in \mathbb{R}^{n_1\times n_2\times \cdots\times n_N},~\mathcal{P}_{\Omega}(\mathcal{X}-\mathcal{F})=0\}$. Then (\ref{3DTNNLRTC}) can be rewritten as
\begin{equation}
\begin{aligned}
    \min_{\mathcal{X}}&~~\sum_{1\leq k_1<k_2\leq N}\alpha_{k_1k_2}\|\mathcal{X}_{(k_1k_2)}\|_{\text{TNN}}+\iota_{\mathbb{S}}(\mathcal{X}),\\
\end{aligned} \label{3DTNNLRTC1}
\end{equation}
where $\alpha_{k_1k_2} \geq 0~(1\leq k_1<k_2\leq N, k_1,k_2\in\mathbb{Z})$ and $\sum_{1\leq k_1<k_2\leq N}\alpha_{k_1k_2}=1$.

\begin{algorithm}[t]
\caption{ADMM-based optimization algorithm for the proposed WSTNN-based LRTC model~(\ref{3DTNNLRTC}).}
\begin{algorithmic}[1]
\renewcommand{\algorithmicrequire}{\textbf{Input:}}
\Require
The observed tensor $\mathcal{F}$,
index set $\Omega$,
weight $\alpha=(\alpha_{11},\alpha_{12},\cdots,\alpha_{1N},\alpha_{23},\cdots,\alpha_{2N},\cdots,\alpha_{N-1N})$, $\beta=(\beta_{11},\beta_{12},\cdots,\beta_{1N},\beta_{23},\cdots,\beta_{2N},\cdots,\beta_{N-1N})$, $\beta_{\text{max}}=(10^{10},10^{10}, \cdots,10^{10})$, and $\gamma=1.1$.
\renewcommand{\algorithmicrequire}{\textbf{Initialization:}}
\Require
$\mathcal{X}_{\Omega}^{(0)}=\mathcal{F}_{\Omega}$, $\mathcal{X}_{\Omega^c}^{(0)}=0$, $\mathcal{Y}_{k_1k_2}^{(0)}=0$, $\mathcal{M}_{k_1k_2}^{(0)}=0$, $p=0$, and $p_{\text{max}}=500$.
\While {not converged and $p<p_{\text{max}}$}
\State Update $\mathcal{Y}_{k_1k_2}^{(p+1)}$ via (\ref{solveYk}),~$1\leq k_1<k_2\leq N, k_1,k_2\in\mathbb{Z}$.
\State Update $\mathcal{X}^{(p+1)}$ via (\ref{solveX}).
\State Update $\mathcal{M}_{k_1k_2}^{(p+1)}$ via (\ref{solve}),~$1\leq k_1<k_2\leq N, k_1,k_2\in\mathbb{Z}$.
\State $\beta=\text{min}(\gamma\beta,\beta_{\text{max}})$ and $p=p+1$.
\EndWhile
\Ensure
The completed tensor $\mathcal{X}$.
\end{algorithmic}
\label{algorithmsTC}
\end{algorithm}

Next, we use ADMM to solve (\ref{3DTNNLRTC1}). We rewrite (\ref{3DTNNLRTC1}) as the following equivalent constrained problem
\begin{equation}
\begin{aligned}
    \min_{\mathcal{X},\mathcal{Y}_{k_1k_2}}&~~\sum_{1\leq k_1<k_2\leq N}\alpha_{k_1k_2}\big\|(\mathcal{Y}_{k_1k_2})_{(k_1k_2)}\big\|_{\text{TNN}}+\iota_{\mathbb{S}}(\mathcal{X})\\
    \text{s.t.}&~~\mathcal{X}-\mathcal{Y}_{k_1k_2}=0,~1\leq k_1<k_2\leq N, k_1,k_2\in\mathbb{Z}.
\end{aligned} \label{3DTNNLRTC2}
\end{equation}
The concise form of the augmented Lagrangian function of (\ref{3DTNNLRTC2}) can be expressed as
\begin{equation}
\begin{aligned}
L_{\beta_{k_1k_2}}(\mathcal{Y}_{k_1k_2}, \mathcal{X}, \mathcal{M}_{k_1k_2})=&
\sum_{1\leq k_1<k_2\leq N}\bigg\{\alpha_{k_1k_2}\big\|(\mathcal{Y}_{k_1k_2})_{(k_1k_2)}\big\|_{\text{TNN}}\\
    &+\frac{\beta_{k_1k_2}}{2}\bigg\|\mathcal{X}-\mathcal{Y}_{k_1k_2}+\frac{\mathcal{M}_{k_1k_2}}{\beta_{k_1k_2}}\bigg\|_F^2\bigg\}+\iota_{\mathbb{S}}(\mathcal{X})+\mathcal{C},
\end{aligned} \label{Lag2}
\end{equation}
where $\mathcal{M}_{k_1k_2}~(1\leq k_1<k_2\leq N, k_1,k_2\in\mathbb{Z})$ are the Lagrange multipliers, $\beta_{k_1k_2}$ $(1\leq k_1<k_2\leq N, k_1,k_2\in\mathbb{Z})$ are the penalty parameters, and $\mathcal{C}$ is a variable, which is independent of $\mathcal{X}$ and $\mathcal{Y}_{k_1k_2}$.
Within the framework of ADMM, $\mathcal{Y}_{k_1k_2}$, $\mathcal{X}$, and $\mathcal{M}_{k_1k_2}$ are alternately updated as
\begin{equation}
\left\{
\begin{aligned}
&\text{Step~1}:~\mathcal{Y}_{k_1k_2}^{(p+1)} = \argmin_{\mathcal{Y}_{k_1k_2}}L_{\beta_{k_1k_2}}(\mathcal{Y}_{k_1k_2}, \mathcal{X}^{(p)}, \mathcal{M}_{k_1k_2}^{(p)}),\\
&\text{Step~2}:~\mathcal{X}^{(p+1)} = \argmin_{\mathcal{X}}L_{\beta_{k_1k_2}}(\mathcal{Y}_{k_1k_2}^{(p+1)}, \mathcal{X}, \mathcal{M}_{k_1k_2}^{(p)}),\\
&\text{Step~3}:~\mathcal{M}_{k_1k_2}^{(p+1)}=\mathcal{M}_{k_1k_2}^{(p)} +\beta_{k_1k_2}(\mathcal{X}^{(p+1)}-\mathcal{Y}_{k_1k_2}^{(p+1)}).\\
\end{aligned}
\right.
\label{solve}
\end{equation}

In Step~1, the $\mathcal{Y}_{k_1k_2}~(1\leq k_1<k_2\leq N, k_1,k_2\in\mathbb{Z})$ subproblems are
\begin{equation}
\begin{aligned}
\mathcal{Y}_{k_1k_2}^{(p+1)} = &\argmin_{\mathcal{Y}_{k_1k_2}} \alpha_{k_1k_2}\big\|(\mathcal{Y}_{k_1k_2})_{(k_1k_2)}\big\|_{\text{TNN}}\\
&+\frac{\beta_{k_1k_2}}{2}\bigg\|(\mathcal{X}_{(k_1k_2)})^{(p)}-(\mathcal{Y}_{k_1k_2})_{(k_1k_2)} +\frac{\big((\mathcal{M}_{k_1k_2})_{(k_1k_2)}\big)^{(p)}}{\beta_{k_1k_2}}\bigg\|_F^2.
\end{aligned}\label{Yk1}
\end{equation}
To solve (\ref{Yk1}), we introduce the following theorem \cite{zhang2017exact}.
\begin{theorem}
\emph{Assuming that $\mathcal{Z}\in \mathbb{R}^{n_1\times n_2\times n_3}$ is a three-way tensor, a minimizer to
$$
\min_{\mathcal{Y}} \tau \|\mathcal{Y}\|_{\text{TNN}}+\frac{1}{2}\|\mathcal{Y}-\mathcal{Z}\|_F^2,
$$
is given by the tensor singular value thresholding (t-SVT)
$$\mathcal{Y}=\mathcal{D}_{\tau}(\mathcal{Z}):=\mathcal{U}\ast\mathcal{S}_{\tau}\ast\mathcal{V}^{\mathrm{T}},$$
where $\mathcal{Z}=\mathcal{U}\ast\mathcal{S}\ast\mathcal{V}^{\mathrm{T}}$ and $\mathcal{S}_{\tau}$ is an $n_1\times n_2\times n_3$ tensor which satisfies
$$\mathcal{\bar{\mathcal{S}}}_{\tau}(i,i,s)=\max(\mathcal{\bar{S}}(i,i,s)-\tau,0),$$
where $\mathcal{\bar{S}}={\tt fft}(\mathcal{S},[],3)$, and $\tau$ is a threshold.}
\label{Ytheorem}
\end{theorem}
Via Theorem \ref{Ytheorem}, $\mathcal{Y}_k$ $(1\leq k_1<k_2\leq N, k_1,k_2\in\mathbb{Z})$ can be updated as
\begin{equation}
\begin{aligned}
\mathcal{Y}_{k_1k_2}^{(p+1)}={\tt t\text{-}fold}
\Bigg(\mathcal{D}_{\frac{\alpha_{k_1k_2}}{\beta_{k_1k_2}}}\bigg((\mathcal{X}_{(k_1k_2)})^{(p)}+\frac{\big((\mathcal{M}_{k_1k_2})_{(k_1k_2)}\big)^{(p)}}{\beta_{k_1k_2}}\bigg),k_1,k_2\Bigg).
\end{aligned}\label{solveYk}
\end{equation}
In Step~2, we solve the following problem
\begin{equation}
\begin{aligned}
\mathcal{X}^{(p+1)} \in \argmin_{\mathcal{X}} \sum_{1\leq k_1<k_2\leq N}\frac{\beta_{k_1k_2}}{2}\bigg\|\mathcal{X}-\mathcal{Y}_{k_1k_2}^{(p+1)}+\frac{\mathcal{M}_{k_1k_2}^{(p)}}{\beta_{k_1k_2}}\bigg\|_F^2+\iota_{\mathbb{S}}(\mathcal{X}),
\end{aligned}\label{X}
\end{equation}
which is differentiable and has a closed-form solution
\begin{equation}
\begin{aligned}
\mathcal{X}^{(p+1)}=\mathcal{P}_{\Omega^c}\Bigg(\frac{\sum_{1\leq k_1<k_2\leq N}\beta_{k_1k_2}\big(\mathcal{Y}_{k_1k_2}^{(p+1)}-\frac{\mathcal{M}_{k_1k_2}^{(p)}}{\beta_{k_1k_2}}\big)}{\sum_{1\leq k_1<k_2\leq N}\beta_{k_1k_2}}\Bigg)+\mathcal{P}_{\Omega}(\mathcal{F}).
\end{aligned}\label{solveX}
\end{equation}
The pseudocode of the developed algorithm is described in Algorithm \ref{algorithmsTC}.
The computational cost at each iteration is $$\mathcal{O}\bigg(D\sum_{1\leq k_1<k_2\leq N}\Big(\log(d_{k_1k_2})+\min(n_{k_1},n_{k_2})\Big)\bigg),$$
where $D=\prod_{k=1}^{N}n_k$ and $d_{k_1k_2}=D/(n_{k_1}n_{k_2})$.

\subsection{WSTNN-based TRPCA model}

The TRPCA aims to exactly recover a low-rank tensor corrupted by sparse noise. Considering an $N$-way tensor $\mathcal{X}\in \mathbb{R}^{n_1\times n_2\times \cdots \times n_N}$, the proposed WSTNN-based TRPCA model can be formulated as
\begin{equation}
\begin{aligned}
    \min_{\mathcal{L},\mathcal{E}}&~~\|\mathcal{L}\|_{\text{WSTNN}}+\lambda\|\mathcal{E}\|_{1}\\
    \text{s.t.}&~~\mathcal{X}=\mathcal{L}+\mathcal{E},
\end{aligned} \label{3DTNNRPCA}
\end{equation}
where $\mathcal{X}$ is the corrupted observation tensor, $\mathcal{L}$ is the low-rank component, $\mathcal{E}$ is the sparse component, and $\lambda$ is a tuning parameter compromising $\mathcal{L}$ and $\mathcal{E}$. And (\ref{3DTNNRPCA}) can be rewritten as
\begin{equation}
\begin{aligned}
    \min_{\mathcal{L},\mathcal{E}}&~~\sum_{1\leq k_1<k_2\leq N}\alpha_{k_1k_2}\|\mathcal{L}_{(k_1k_2)}\|_{\text{TNN}}+\lambda\|\mathcal{E}\|_{1}\\
    \text{s.t.}&~~~~~~~~~~~~~~~~~~~~~~\mathcal{X}=\mathcal{L}+\mathcal{E}.
\end{aligned} \label{3DTNNRPCA1}
\end{equation}
where $\alpha_{k_1k_2} \geq 0~(1\leq k_1<k_2\leq N, k_1,k_2\in\mathbb{Z})$ and $\sum_{1\leq k_1<k_2\leq N}\alpha_{k_1k_2}=1$.

\begin{algorithm}[t]
\caption{ADMM-based optimization algorithm for the proposed WSTNN-based TRPCA model~(\ref{3DTNNRPCA}).}
\begin{algorithmic}[1]
\renewcommand{\algorithmicrequire}{\textbf{Input:}}
\Require
The corrupted observation tensor $\mathcal{X}$,
weight $\alpha=(\alpha_{11},\alpha_{12},\cdots,\alpha_{1N},\alpha_{23},\cdots,\alpha_{2N},\cdots,\alpha_{N-1N})$, $\beta=(\beta_{11},\beta_{12},\cdots,\beta_{1N},\beta_{23},\cdots,\beta_{2N},\cdots,\beta_{N-1N})$, $\beta_{\text{max}}=(10^{10},10^{10}, \cdots,10^{10})$, $\lambda$, $\rho$, $\rho_{\text{max}}=10^{10}$, and $\gamma=1.2$.
\State Initialization: $\mathcal{L}^{(0)}=0$, $\mathcal{E}^{(0)}=0$, $\mathcal{M}^{(0)}=0$, $\mathcal{Z}_{k_1k_2}^{(0)}=0$, $\mathcal{P}_{k_1k_2}^{(0)}=0$, and $p_{\text{max}}=500$.
\While {not converged and $p<p_{\text{max}}$}
\State Update $\mathcal{Z}_{k_1k_2}^{(p+1)}$ via (\ref{solveRZ}), $1\leq k_1<k_2\leq N$.
\State Update $\mathcal{L}^{(p+1)}$ via (\ref{solveL}).
\State Update $\mathcal{E}^{(p+1)}$ via (\ref{solveRE}).
\State Update $\mathcal{P}_{k_1k_2}^{(p+1)}$ via (\ref{solveRPCA}), $1\leq k_1<k_2\leq N$.
\State Update $\mathcal{M}^{(p+1)}$ via (\ref{solveRPCA}).
\State $\beta=\text{min}(\gamma\beta,\beta_{\text{max}})$, $\rho=\text{min}(\gamma\rho,\rho_{\text{max}})$, and $p=p+1$.
\EndWhile
\Ensure
The low-rank component $\mathcal{L}$ and the sparse component $\mathcal{E}$.
\end{algorithmic}
\label{algRPCA}
\end{algorithm}

Next, we use ADMM to solve (\ref{3DTNNRPCA1}). We rewrite (\ref{3DTNNRPCA1}) as
\begin{equation}
\begin{aligned}
    \min_{\mathcal{L},\mathcal{E},\mathcal{Z}_{k_1k_2}}&~~\sum_{1\leq k_1<k_2\leq N}\alpha_{k_1k_2}\big\|(\mathcal{Z}_{k_1k_2})_{(k_1k_2)}\big\|_{\text{TNN}}+\lambda\|\mathcal{E}\|_{1}\\
    \text{s.t.}&~~\mathcal{X}=\mathcal{L}+\mathcal{E},\\
    &~~\mathcal{L}-\mathcal{Z}_{k_1k_2}=0,~1\leq k_1<k_2\leq N, k_1,k_2\in\mathbb{Z}.
\end{aligned} \label{3DTNNRPCA2}
\end{equation}
The concise form of the augmented Lagrangian function of (\ref{3DTNNRPCA2}) can be expressed as
\begin{equation}
\begin{aligned}
    &L_{\beta_{k_1k_2},\rho}(\mathcal{L},\mathcal{Z}_{k_1k_2},\mathcal{P}_{k_1k_2}, \mathcal{E},\mathcal{M})=
    \sum_{1\leq k_1<k_2\leq N}\bigg\{\alpha_{k_1k_2}\big\|(\mathcal{Z}_{k_1k_2})_{({k_1k_2})}\big\|_{\text{TNN}}\\
    &+\frac{\beta_{k_1k_2}}{2}\bigg\|\mathcal{L}-\mathcal{Z}_{k_1k_2}+\frac{\mathcal{P}_{k_1k_2}}{\beta_{k_1k_2}}\bigg\|_F^2\bigg\}
    +\lambda\|\mathcal{E}\|_{1}+\frac{\rho}{2}\bigg\|\mathcal{X}-\mathcal{L}-\mathcal{E}+\frac{\mathcal{M}}{\rho}\bigg\|_F^2+\mathcal{C},
\end{aligned} \label{RLag2}
\end{equation}
where $\mathcal{P}_{k_1k_2}$ and $\mathcal{M}$ are the Lagrange multipliers, $\beta_{k_1k_2}$ and $\rho$ are the penalty parameters, and $\mathcal{C}$ is a variable, which is independent of $\mathcal{L}$ and $\mathcal{E}$, and $\mathcal{Z}_{k_1k_2}$. To minimize (\ref{RLag2}), we can update $\mathcal{L},\mathcal{Z}_{k_1k_2},\mathcal{P}_{k_1k_2}, \mathcal{E},\mathcal{M}$ ($1\leq k_1<k_2\leq N, k_1,k_2\in\mathbb{Z}$) as
\begin{equation}
\left\{
\begin{aligned}
&\text{Step~1}:~\mathcal{Z}_{k_1k_2}^{(p+1)} = \argmin_{\mathcal{Z}_{k_1k_2}}L_{\beta_{k_1k_2},\rho}(\mathcal{L}^{(p)},\mathcal{Z}_{k_1k_2},\mathcal{P}_{k_1k_2}^{(p)}, \mathcal{E}^{(p)},\mathcal{M}^{(p)}),\\
&\text{Step~2}:~\mathcal{L}^{(p+1)} = \argmin_{\mathcal{L}}L_{\beta_{k_1k_2},\rho}(\mathcal{L},\mathcal{Z}_{k_1k_2}^{(p+1)},\mathcal{P}_{k_1k_2}^{(p)}, \mathcal{E}^{(p)},\mathcal{M}^{(p)}),\\
&\text{Step~3}:~\mathcal{E}^{(p+1)} = \argmin_{\mathcal{E}}L_{\beta_{k_1k_2},\rho}(\mathcal{L}^{(p+1)},\mathcal{Z}_{k_1k_2}^{(p+1)},\mathcal{P}_{k_1k_2}^{(p)}, \mathcal{E},\mathcal{M}^{(p)}),\\
&\text{Step~4}:~\mathcal{P}_{k_1k_2}^{(p+1)}=\mathcal{P}_{k_1k_2}^{(p)} +\beta_{k_1k_2}\big(\mathcal{L}^{(p+1)}-\mathcal{Z}_{k_1k_2}^{(p+1)}\big),\\
&\text{Step~5}:~\mathcal{M}^{(p+1)}=\mathcal{M}^{(p)} +\rho\big(\mathcal{X}-\mathcal{L}^{(p+1)}-\mathcal{E}^{(p+1)}\big).\\
\end{aligned}
\right.
\label{solveRPCA}
\end{equation}

In Step 1, the $\mathcal{Z}_{k_1k_2}$ $(1\leq k_1<k_2\leq N, k_1,k_2\in\mathbb{Z})$ subproblem can be solved as
\begin{equation}
\begin{aligned}
\mathcal{Z}_{k_1k_2}^{(p+1)}={\tt t\text{-}fold}
\Bigg(\mathcal{D}_{\frac{\alpha_{k_1k_2}}{\beta_{k_1k_2}}}\bigg((\mathcal{L}_{({k_1k_2})})^{(p)}+\frac{\big((\mathcal{P}_{k_1k_2})_{(k_1k_2)}\big)^{(p)}}{\beta_{k_1k_2}}\bigg),k_1,k_2\Bigg).
\end{aligned}\label{solveRZ}
\end{equation}
In Step 2, the $\mathcal{L}$ subproblem has the following closed-form solution
\begin{equation}
\begin{aligned}
\mathcal{L}^{(p+1)}=\frac{\rho\Big(\mathcal{X}-\mathcal{E}^{(p)}+\frac{\mathcal{M}^{(p)}}{\rho}\Big)+\sum_{1\leq k_1<k_2\leq N}\beta_{k_1k_2}\Big(\mathcal{Z}_{k_1k_2}^{(p+1)}-\frac{\mathcal{P}_{k_1k_2}^{(p)}}{\beta_{k_1k_2}}\Big)}{\rho+\sum_{1\leq k_1<k_2\leq N}\beta_{k_1k_2}}.
\end{aligned}\label{solveL}
\end{equation}
In Step 3, we solve the following problem
\begin{equation}
\begin{aligned}
\mathcal{E}^{(p+1)} \in \argmin_{\mathcal{E}} \lambda\|\mathcal{E}\|_{1}+\frac{\rho}{2}\bigg\|\mathcal{X}-\mathcal{L}^{(p+1)}-\mathcal{E}+\frac{\mathcal{M}^{(p)}}{\rho}\bigg\|_F^2,
\end{aligned}\label{RE}
\end{equation}
which has the following closed-form solution
\begin{equation}
\begin{aligned}
\mathcal{E}^{(p+1)}=\mathcal{S}_{\frac{\lambda}{\rho}}\bigg(\mathcal{X}-\mathcal{L}^{(p+1)}+\frac{\mathcal{M}^{(p)}}{\rho}\bigg),
\end{aligned}\label{solveRE}
\end{equation}
where $\mathcal{S}_{\xi}(\cdot)$ is the tensor soft thresholding operator with threshold $\xi$, i.e.,
\begin{equation}
\begin{aligned}
\big[\mathcal{S}_{\xi}(\mathcal{X})\big]_{i_1 i_2\cdots i_N}=\mathrm{sgn}(x_{i_1 i_2\cdots i_N})\max(|x_{i_1i_2\cdots i_N}|-\xi,0).
\end{aligned}\label{tsthreshold}
\end{equation}

The pseudocode of the proposed algorithm to solve the proposed WSTNN-based TRPCA model (\ref{3DTNNRPCA}) is described in Algorithm \ref{algRPCA}.
The computational cost at each iteration is $$\mathcal{O}\bigg(D\sum_{1\leq k_1<k_2\leq N}\Big(\log(d_{k_1k_2})+\min(n_{k_1},n_{k_2})\Big)\bigg),$$
where $D=\prod_{k=1}^{N}n_k$ and $d_{k_1k_2}=D/(n_{k_1}n_{k_2})$.

\begin{table}[!t]
\scriptsize
\setlength{\tabcolsep}{10pt}
\renewcommand\arraystretch{1.2}
\caption{Parameters setting in the proposed WSTNN-based LRTC method on different data.}
\begin{center}
\begin{tabular}{c|c|c|c|c}
  \Xhline{1pt}
 Test                  &\multicolumn{2}{c|}{Data}        &$\alpha$             &$\tau$  \\

\hline

\multirow{2}{*}{\tabincell{c}{synthetic data \\ completion}}

                      &\multicolumn{2}{c|}{three-way tensor}         &(1,1,1)/3      & (10,10,10)        \\

                      &\multicolumn{2}{c|}{four-way tensor}          &(1,1,1,1,1,1)/6    & (50,50,50,50,50,50)     \\

                      \hline

 \multirow{4}{*}{\tabincell{c}{real-world\\data completion}}

                      &\multirow{2}{*}{three-way tensor}    & HSI/MSI &(0.001,1,1)/2.001      &\multirow{2}{*}{(100,100,100)}          \\
                      &                                      & MRI     &(1,1,1)/3      &           \\

                      \cline{2-5}

                      &\multirow{2}{*}{four-way tensor}    &  CV     &\multirow{2}{*}{(1,1,1,1,1,1)/6}       &\multirow{2}{*}{(500,500,500,500,500,500)} \\
                      &                                     &  HSV         &      & \\

 \Xhline{1pt}
\end{tabular}
\end{center}\vspace{-0.1cm}
\label{parasetTC}
\end{table}

\begin{figure}[!t]
\setlength{\tabcolsep}{5pt}
\begin{center}
\begin{tabular}{cccc}
\includegraphics[width=0.23\textwidth]{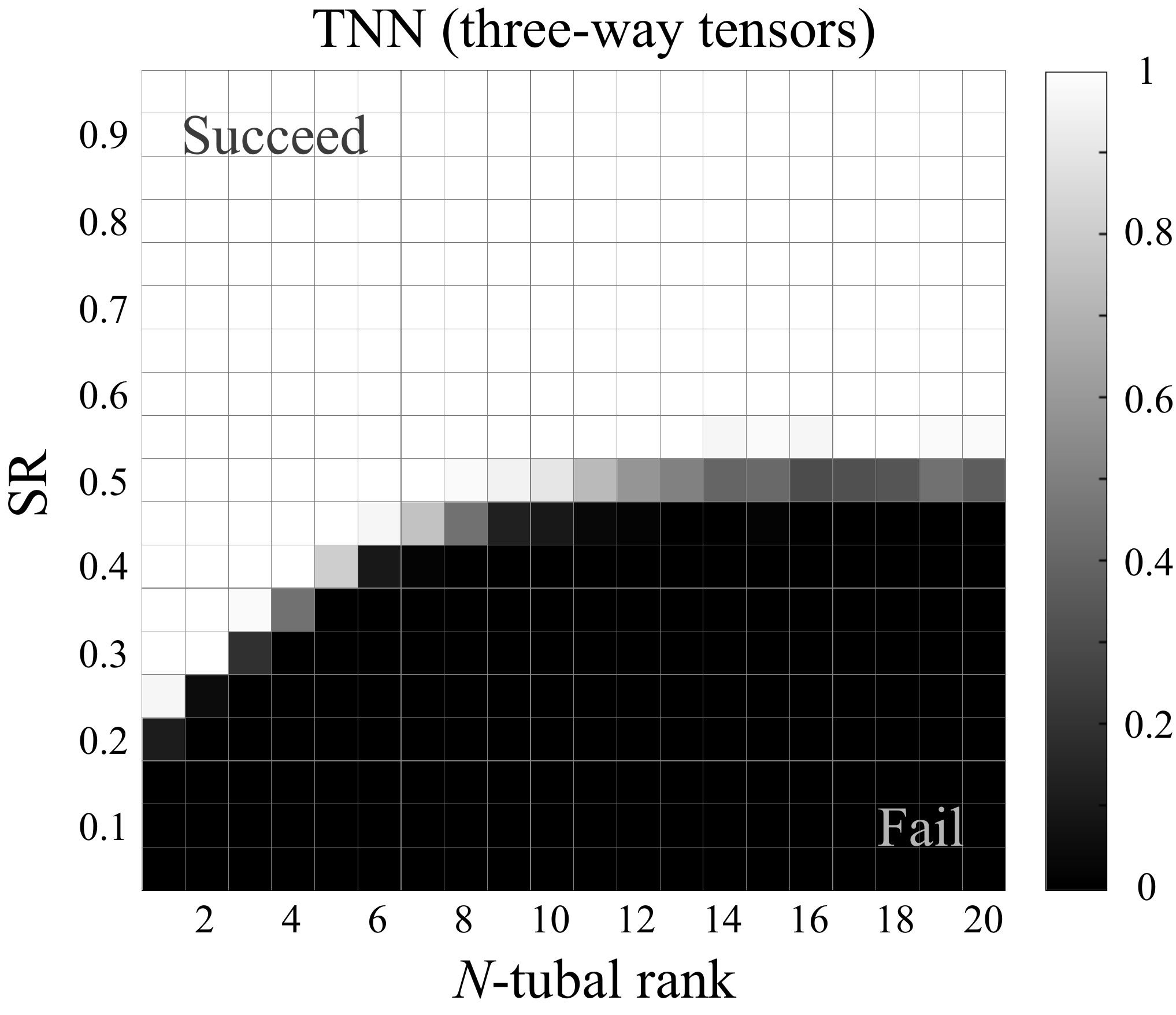}&
\includegraphics[width=0.23\textwidth]{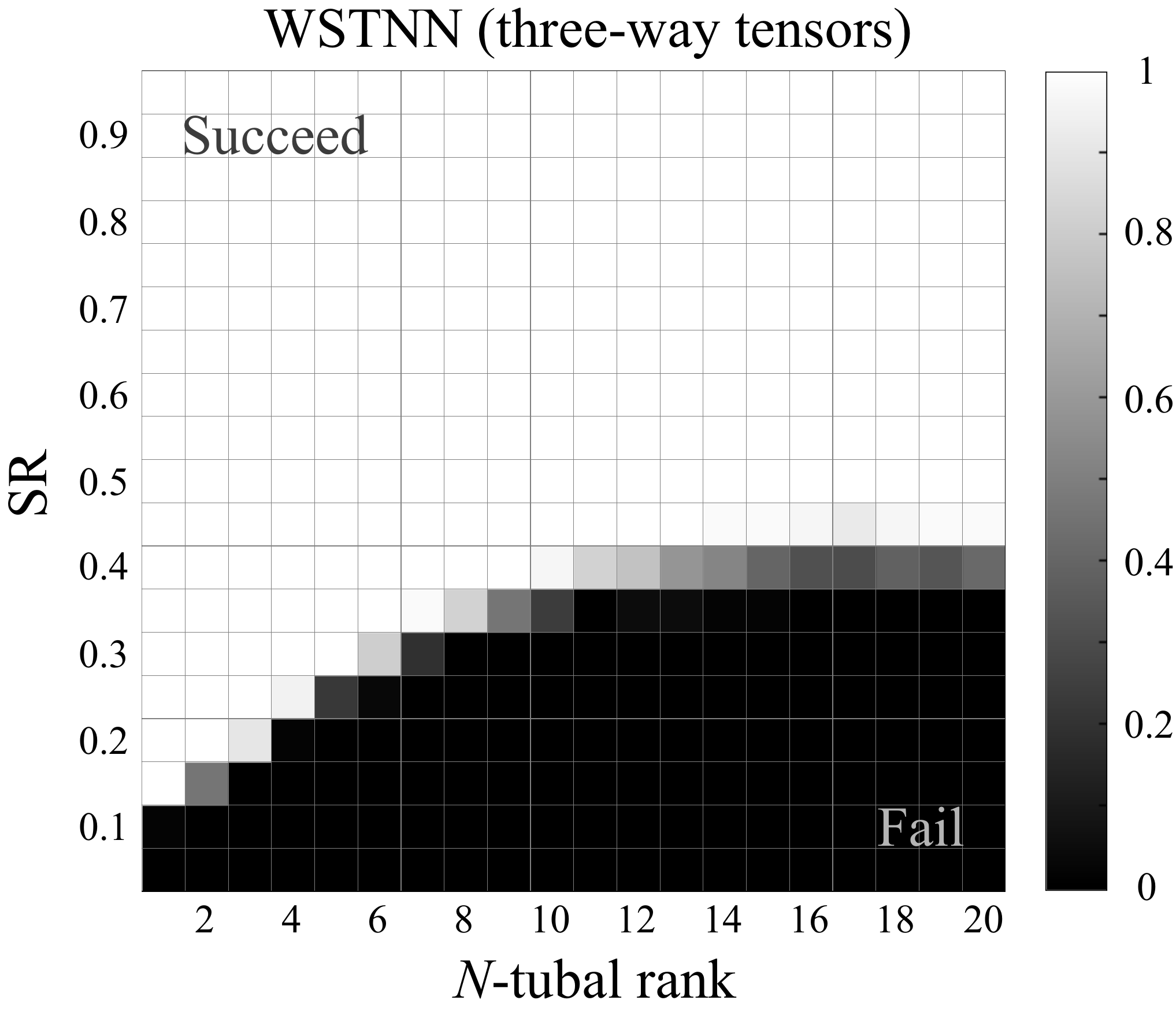}&
\includegraphics[width=0.23\textwidth]{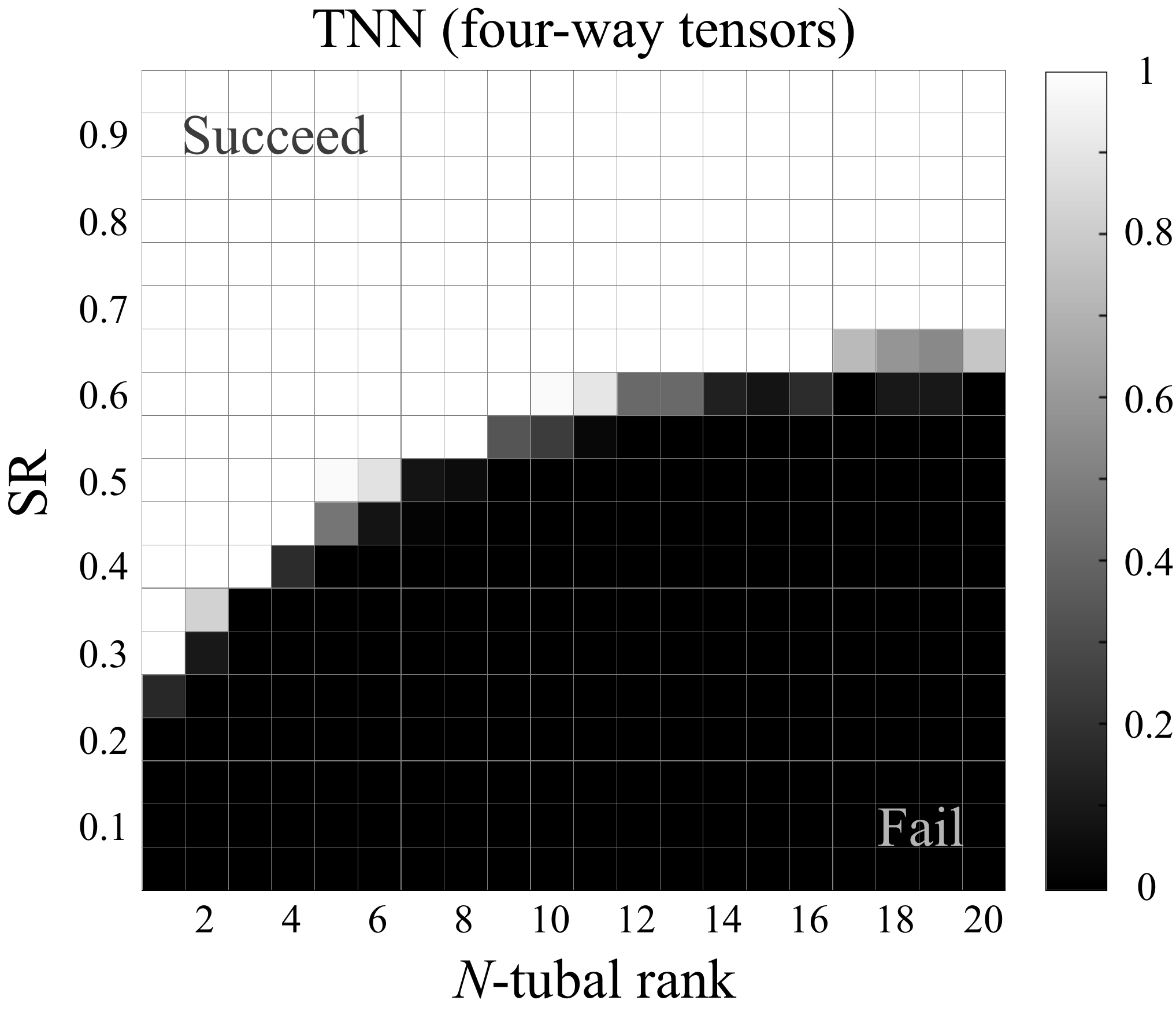}&
\includegraphics[width=0.23\textwidth]{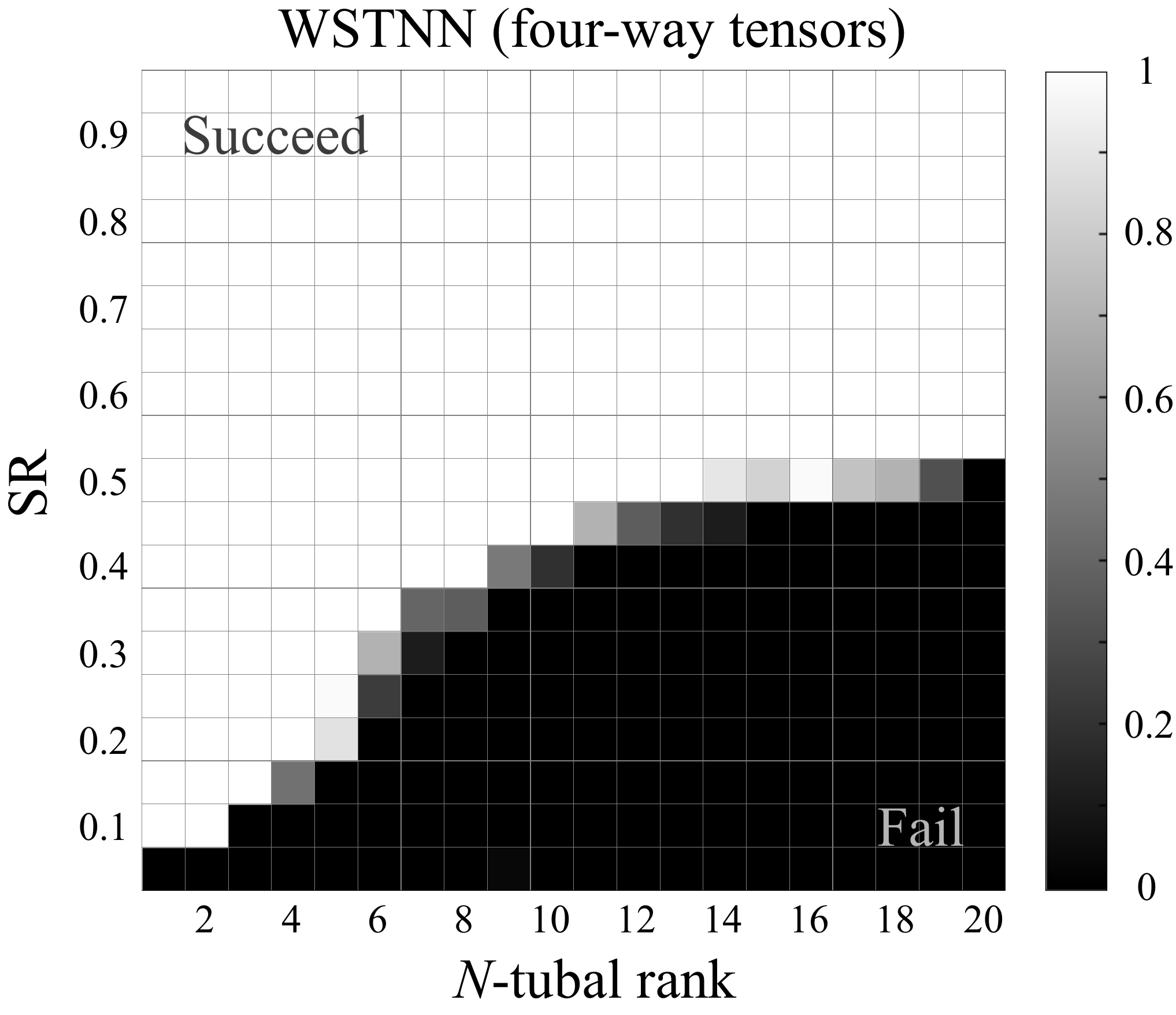}
\end{tabular}
\caption{The success rates for synthetic data with varying $N$-tubal rank and varying SRs. The left two are the results of TNN-based LRTC method \cite{zhang2017exact} and the proposed WSTNN-based LRTC method on three-way tensors.
The right two are the results of TNN-based LRTC method \cite{zhang2017exact} and the proposed WSTNN-based LRTC method on four-way tensors. The gray magnitude represents the success rates.}\vspace{-0.2cm}
\label{sy}
\end{center}\vspace{-0.3cm}
\end{figure}

\section{Numerical experiments} \label{sec:NE}

We evaluate the performance of the proposed WSTNN-based LRTC and TRPCA methods. Both synthetic and real-world data are tested.
We employ the peak signal to noise rate (PSNR), the structural similarity (SSIM) \cite{SSIM}, and the feature similarity (FSIM) \cite{FSIM} to measure the quality of the recovered results. All of them are positive indexes, i.e., the higher value, the better quality.
All tests are implemented on the platform of Windows 7 and MATLAB (R2017b) with an Intel Core i5-4590 3.30GHz and 16GB RAM.

\subsection{Low-rank tensor completion}

In this section, we test both synthetic data and five kinds of real-world data: MSI, HSI, MRI, color video (CV), and hyperspectral video (HSV). If not specified, the methodology for sampling the data is purely random sampling.
The compared LRTC methods include: HaLRTC \cite{Liu2013PAMItensor} and LRTC-TVI \cite{Li2017LowRankTC}, representing state-of-the-arts for Tucker decomposition based method;
BCPF \cite{ZhangCP}, representing state-of-the-arts for CP decomposition based method; logDet \cite{JilogDet}, TNN \cite{zhang2017exact}, PSTNN \cite{jiang2017novel2}, and t-TNN \cite{HutTNN}, representing state-of-the-arts for t-SVD based method. In all four-way tensors tests, as logDet, TNN, PSTNN, and t-TNN, are applicable only to three-way tensors, we first reshape the four-way tensors into three-way tensors, and then perform them.

\textbf{Parameter selection.} In all tests, the stopping criterion lies on the relative change (RelCha) of two successive recovered tensors, i.e.,
$\text{RelCha}=\frac{\|\mathcal{X}^{(p+1)}-\mathcal{X}^{(p)}\|_{F}}{\|\mathcal{X}^{(p)}\|_{F}}< 10^{-4}$. Letting the threshold parameter $\tau=\alpha./\beta$, then $\alpha$ is chosen by the strategy of the weight selection in Section \ref{secmodel}, $\tau$ is set to be $\omega\times{\tt ones}(N(N-1)/2,1)$\textsuperscript{\ref{ones}}, and $\omega$ is empirically selected from a candidate set: $\{1,10,50,100,500,1000,10000\}$. Table \ref{parasetTC} shows the parameters setting in the proposed WSTNN-based LRTC method on different data.

\begin{table}[!t]
\scriptsize
\setlength{\tabcolsep}{8pt}
\renewcommand\arraystretch{1.2}
\caption{The average PSNR, SSIM, and FSIM values of all testing 32 MSIs by eight utilized LRTC methods.}
\begin{center}
\begin{tabular}{c|ccc|ccc|ccc|c}
 \Xhline{1pt}
                      SR        &\multicolumn{3}{c|}{5\%}  &\multicolumn{3}{c|}{10\%}  & \multicolumn{3}{c|}{20\%}         &\multirow{2}{*}{Time(s)} \\
\cline{1-10}
                      Method             &PSNR         & SSIM        &FSIM          &PSNR       & SSIM        &FSIM        &PSNR        & SSIM      &FSIM  \\

                      \hline

                      HaLRTC            &14.90 	      &0.242 	     &0.648 	   &21.43   	&0.537  	  &0.773  	   &32.90   	 &0.892  	 &0.933   &\bf{13.64}\\

                      LRTC-TVI          &23.92        &0.718         &0.812 	   &29.21       &0.868        &0.895       &34.17 	     &0.941 	 &0.953   &472.3\\

                      BCPF              &30.47        &0.785         &0.884 	   &35.66   	&0.903 	      &0.936       &39.62  	     &0.944 	 &0.962   &642.7\\

                      logDet            &16.99 	      &0.309      	 &0.679 	   &31.27 	    &0.780   	  &0.894 	   &40.81 	     &0.968 	 &0.977   &46.31\\

                      TNN               &17.64        &0.332 	     &0.692 	   &30.90 	    &0.780 	      &0.894 	   &39.60  	     &0.962 	 &0.974   &46.14\\

                      PSTNN             &19.56 	      &0.264 	     &0.526 	   &32.95  	    &0.809 	      &0.882 	   &40.77        &0.962    	 &0.973   &63.48\\

                      t-TNN             &28.32 	      &0.779 	     &0.874 	   &35.45 	    &0.942 	      &0.954 	   &42.67 	     &0.985 	 &0.987   &24.79\\

                      WSTNN            &\bf{32.03}  &\bf{0.881}    &\bf{0.930}   &\bf{38.74}  &\bf{0.977}   &\bf{0.979}  &\bf{45.70}   &\bf{0.994}  &\bf{0.994}&75.31\\

 \Xhline{1pt}
\end{tabular}
\end{center}\vspace{-0.2cm}
\label{MSITC}
\end{table}

\begin{figure}[!t]
\begin{center}
\includegraphics[width=0.949\textwidth]{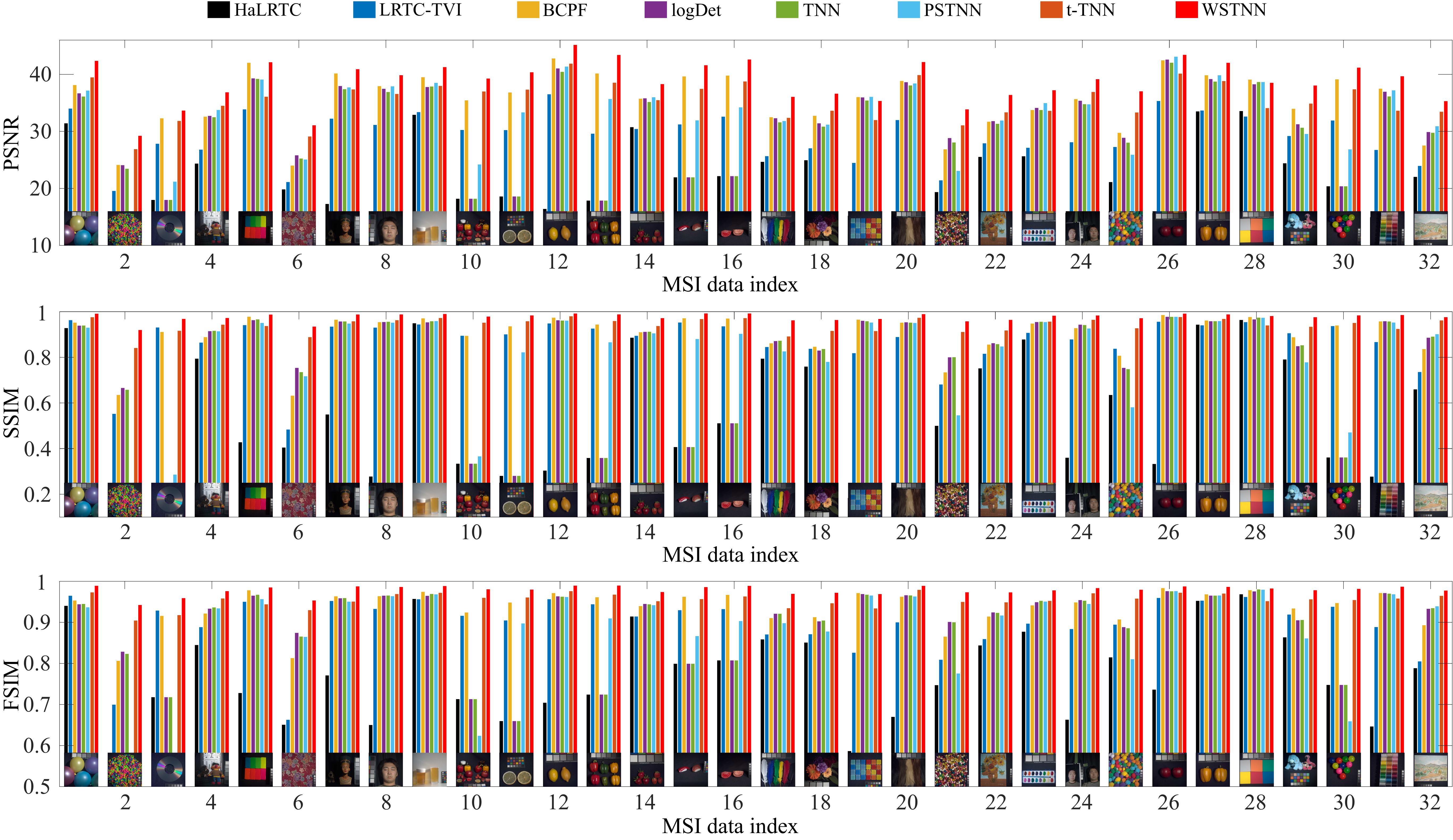}~~\vspace{-0.1cm}
\caption{The PSNR, SSIM, and FSIM values of all 32 testing MSIs in the dataset CAVE output by eight utilized LRTC methods with $\text{SR}=10\%$.}\label{MSIqua}
\end{center}\vspace{-0.6cm}
\end{figure}

\begin{figure}[!t]
\tiny
\setlength{\tabcolsep}{0.9pt}
\begin{center}
\begin{tabular}{cccccccccc}
\includegraphics[width=0.096\textwidth]{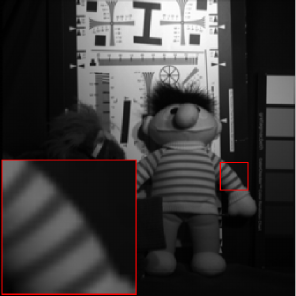}&
\includegraphics[width=0.096\textwidth]{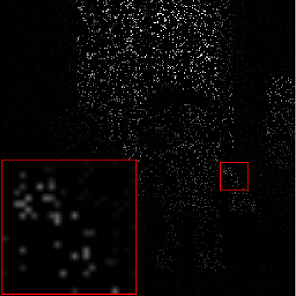}&
\includegraphics[width=0.096\textwidth]{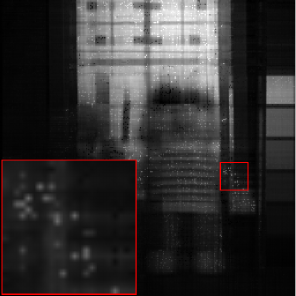}&
\includegraphics[width=0.096\textwidth]{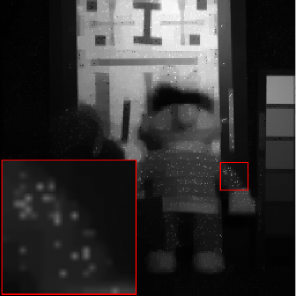}&
\includegraphics[width=0.096\textwidth]{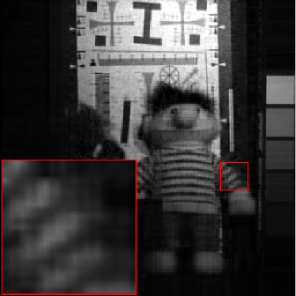}&
\includegraphics[width=0.096\textwidth]{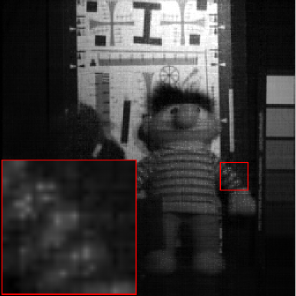}&
\includegraphics[width=0.096\textwidth]{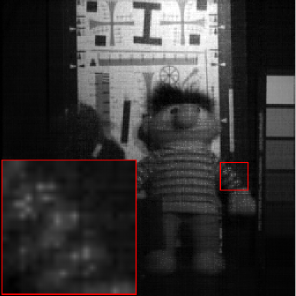}&
\includegraphics[width=0.096\textwidth]{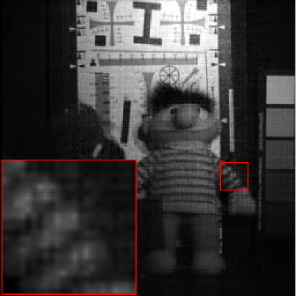}&
\includegraphics[width=0.096\textwidth]{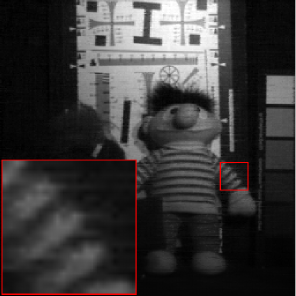}&
\includegraphics[width=0.096\textwidth]{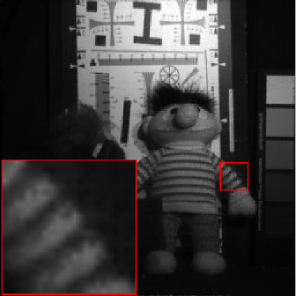}\\

\includegraphics[width=0.096\textwidth]{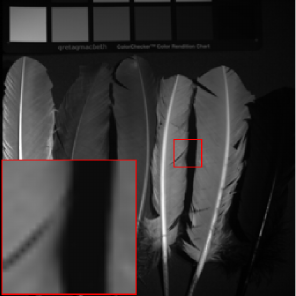}&
\includegraphics[width=0.096\textwidth]{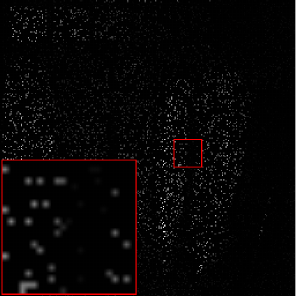}&
\includegraphics[width=0.096\textwidth]{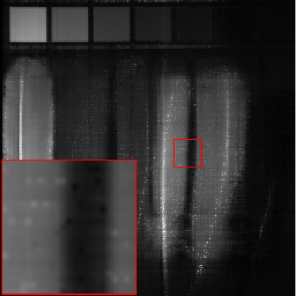}&
\includegraphics[width=0.096\textwidth]{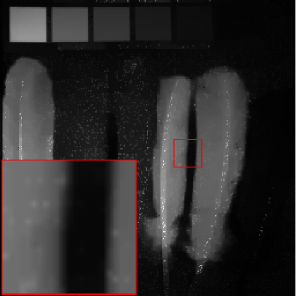}&
\includegraphics[width=0.096\textwidth]{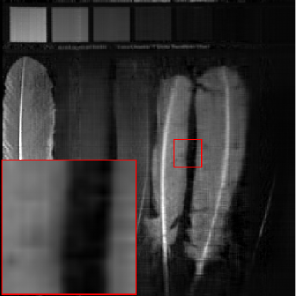}&
\includegraphics[width=0.096\textwidth]{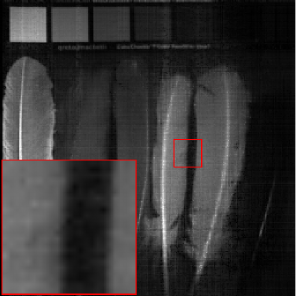}&
\includegraphics[width=0.096\textwidth]{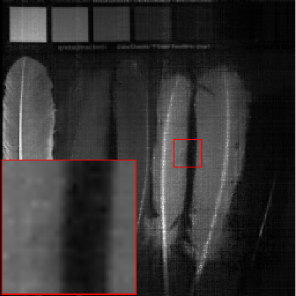}&
\includegraphics[width=0.096\textwidth]{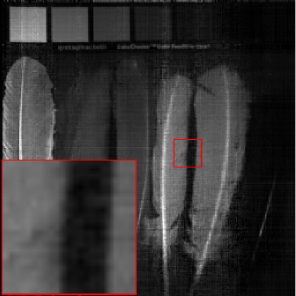}&
\includegraphics[width=0.096\textwidth]{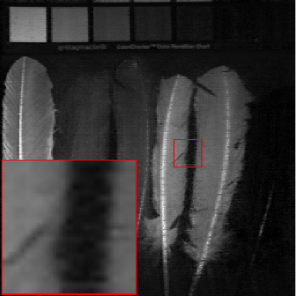}&
\includegraphics[width=0.096\textwidth]{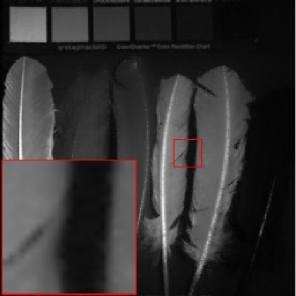}\\

\includegraphics[width=0.096\textwidth]{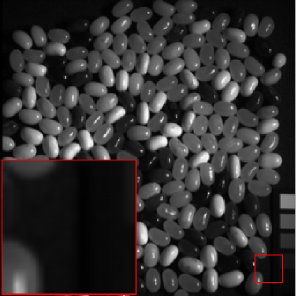}&
\includegraphics[width=0.096\textwidth]{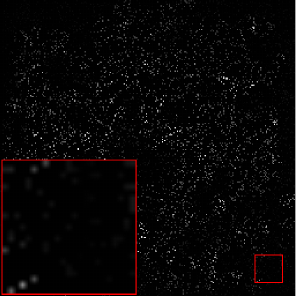}&
\includegraphics[width=0.096\textwidth]{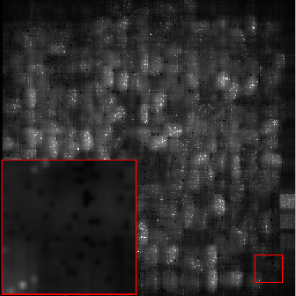}&
\includegraphics[width=0.096\textwidth]{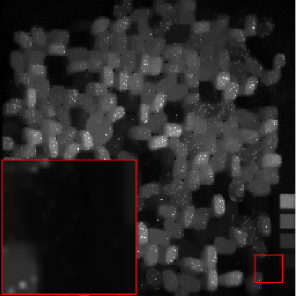}&
\includegraphics[width=0.096\textwidth]{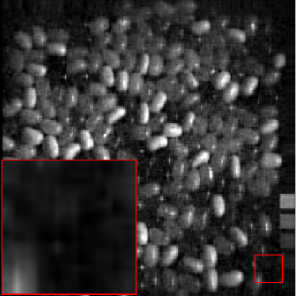}&
\includegraphics[width=0.096\textwidth]{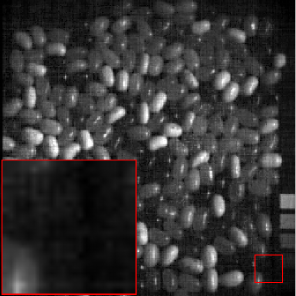}&
\includegraphics[width=0.096\textwidth]{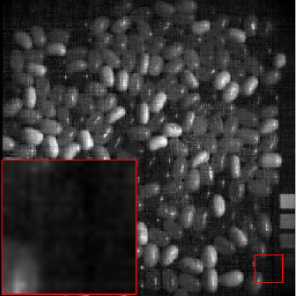}&
\includegraphics[width=0.096\textwidth]{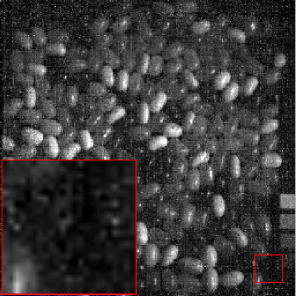}&
\includegraphics[width=0.096\textwidth]{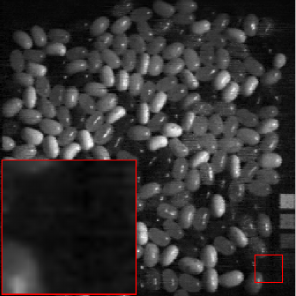}&
\includegraphics[width=0.096\textwidth]{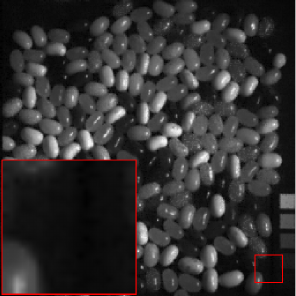}\\

\includegraphics[width=0.096\textwidth]{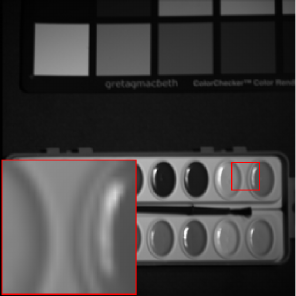}&
\includegraphics[width=0.096\textwidth]{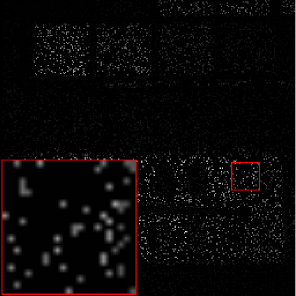}&
\includegraphics[width=0.096\textwidth]{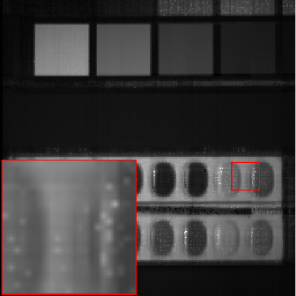}&
\includegraphics[width=0.096\textwidth]{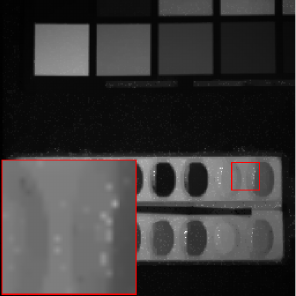}&
\includegraphics[width=0.096\textwidth]{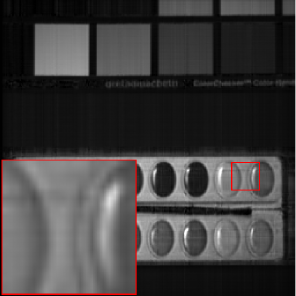}&
\includegraphics[width=0.096\textwidth]{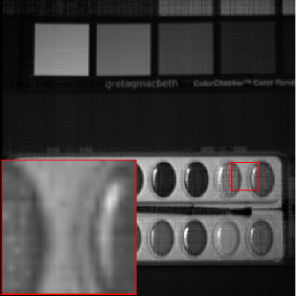}&
\includegraphics[width=0.096\textwidth]{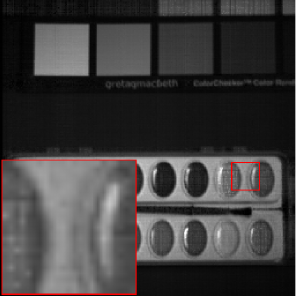}&
\includegraphics[width=0.096\textwidth]{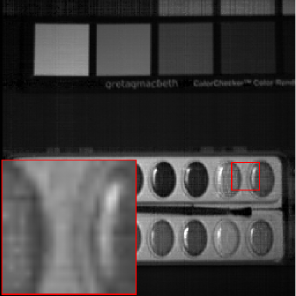}&
\includegraphics[width=0.096\textwidth]{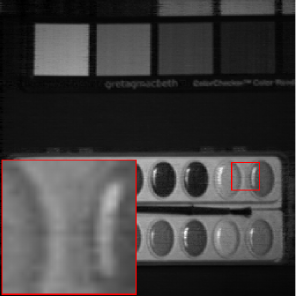}&
\includegraphics[width=0.096\textwidth]{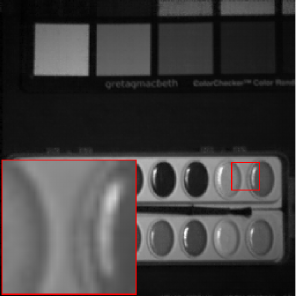}\\

Original& Observed & HaLRTC \cite{Liu2013PAMItensor}& LRTC-TVI \cite{Li2017LowRankTC} &BCPF \cite{ZhangCP} & logDet \cite{JilogDet} & TNN \cite{zhang2017exact}& PSTNN \cite{jiang2017novel2} & t-TNN \cite{HutTNN} & WSTNN
  \end{tabular}
  \caption{The completion results of four selected MSIs with $\text{SR}=10\%$. From top to bottom: the images located at the 31-th band in \emph{chart and stuffed toy}, \emph{feathers}, \emph{jelly beans}, and \emph{paints}, respectively.}\vspace{-0.2cm}
  \label{imageMSIfig}
  \end{center}
\end{figure}

\textbf{Synthetic data completion.} We test both synthetic three-way tensors of size $30 \times 30 \times 30$ and four-way tensors of size $30 \times 30 \times 30\times 30$. The testing synthetic tensors are consisted by the sum of $r$ rank-one tensors, which are generated by performing vector outer product on $N$ ($N=3~\text{or}~4$) random vectors. In practice, the data in each test is regenerated and is confirmed to meet the conditions in Theorem \ref{CPfiber}, i.e., its $N$-tubal rank is $r\times{\tt ones}(N(N-1)/2,1)$. We define the success rate as the ratio of successful times to the total number of times, and one test is successful if the relative square error of the recovered tensor $\hat{\mathcal{X}}$ and the ground truth tensor $\mathcal{X}$, i.e. $\|\hat{\mathcal{X}}-\mathcal{X}\|_F^2/\|\mathcal{X}\|_F^2$, is less than $10^{-3}$.

 We test data with different $N$-tubal rank and the sampling rates (SRs), which is defined as the proportion of the known elements. The $N$-tubal rank are set to be $r\times{\tt ones}(N(N-1)/2,1)~(r=1,2,...,20)$ and the SRs are set to be $0.05\times s~(s=1,2,\cdots,19)$. For each pair of $N$-tubal rank and SR, we conduct 50 independent tests and calculate the success rate. Fig. \ref{sy} shows the success rates for varying $N$-tubal rank and varying SRs, and the results demonstrate that the proposed WSTNN-based LRTC method is more robust and preferable than the TNN-based method \cite{zhang2017exact}.

\begin{figure}[!t]
\tiny
\setlength{\tabcolsep}{0.9pt}
\begin{center}
\begin{tabular}{cccccccccc}
\includegraphics[width=0.096\textwidth]{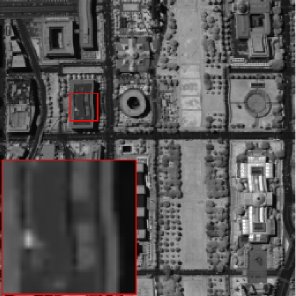}&
\includegraphics[width=0.096\textwidth]{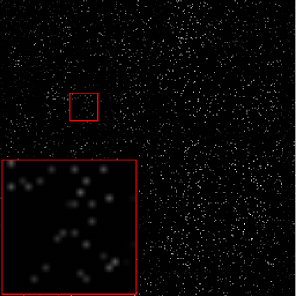}&
\includegraphics[width=0.096\textwidth]{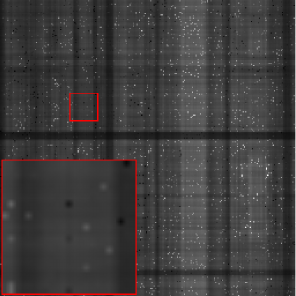}&
\includegraphics[width=0.096\textwidth]{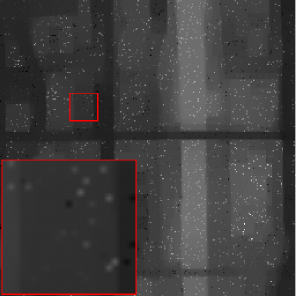}&
\includegraphics[width=0.096\textwidth]{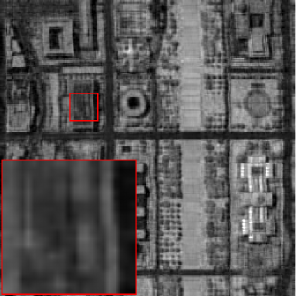}&
\includegraphics[width=0.096\textwidth]{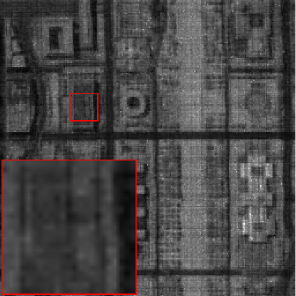}&
\includegraphics[width=0.096\textwidth]{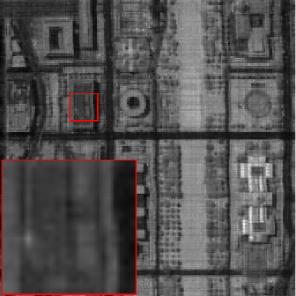}&
\includegraphics[width=0.096\textwidth]{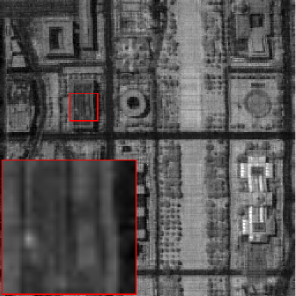}&
\includegraphics[width=0.096\textwidth]{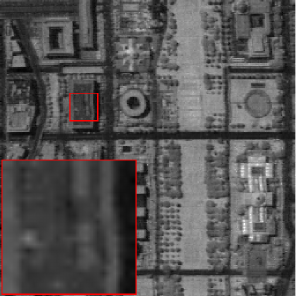}&
\includegraphics[width=0.096\textwidth]{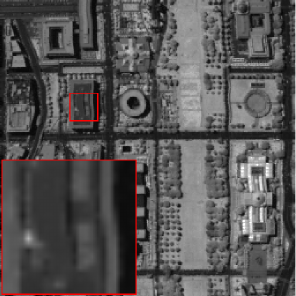}\\
\includegraphics[width=0.096\textwidth]{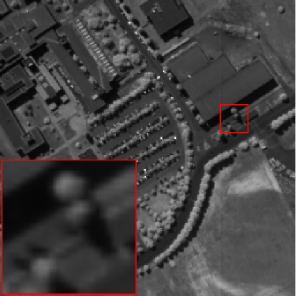}&
\includegraphics[width=0.096\textwidth]{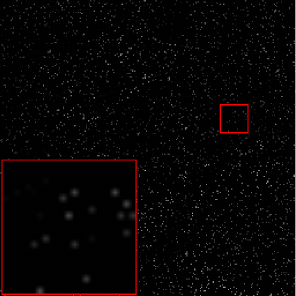}&
\includegraphics[width=0.096\textwidth]{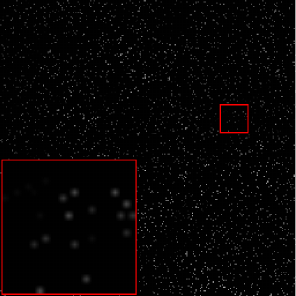}&
\includegraphics[width=0.096\textwidth]{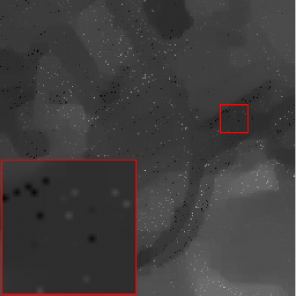}&
\includegraphics[width=0.096\textwidth]{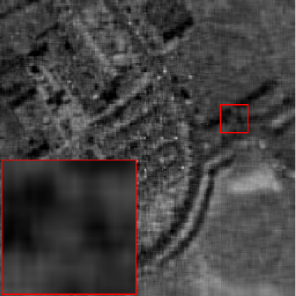}&
\includegraphics[width=0.096\textwidth]{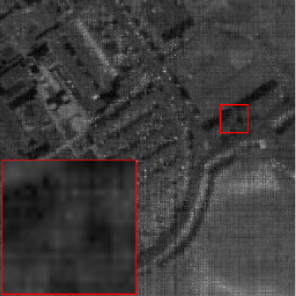}&
\includegraphics[width=0.096\textwidth]{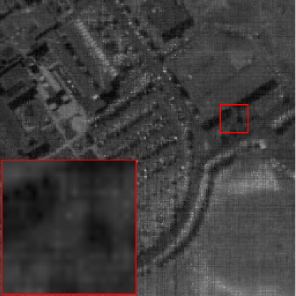}&
\includegraphics[width=0.096\textwidth]{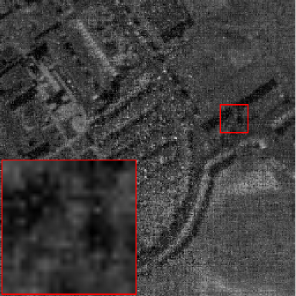}&
\includegraphics[width=0.096\textwidth]{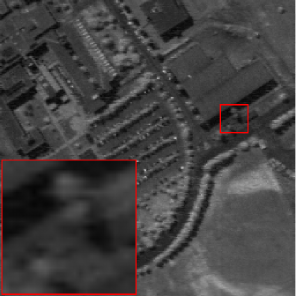}&
\includegraphics[width=0.096\textwidth]{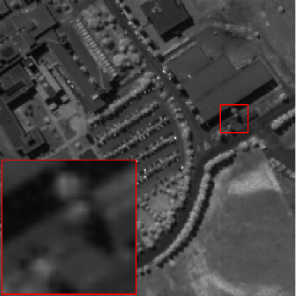}\\
Original& Observed & HaLRTC \cite{Liu2013PAMItensor}& LRTC-TVI \cite{Li2017LowRankTC} &BCPF \cite{ZhangCP} & logDet \cite{JilogDet} & TNN \cite{zhang2017exact}& PSTNN \cite{jiang2017novel2} & t-TNN \cite{HutTNN} & WSTNN
  \end{tabular}
  \caption{The completion results of the HSIs \emph{Washington DC Mall} and \emph{Pavia University} with $\text{SR}=5\%$. Top row: the image located at the 70-th band in \emph{Washington DC Mall}. Bottom row: the image located at the 85-th band in \emph{Pavia University}.}\vspace{-0.2cm}
  \label{imageWDCfig}
  \end{center}
\end{figure}

\begin{table}[!t]
\scriptsize
\setlength{\tabcolsep}{5pt}
\renewcommand\arraystretch{1.2}
\caption{The PSNR, SSIM, and FSIM values output by eight utilized LRTC methods for HSIs.}
\begin{center}
\begin{tabular}{c|c|ccc|ccc|ccc|c}
  \Xhline{1pt}
\multirow{2}{*}{HSI}  &SR                  &\multicolumn{3}{c|}{5\%}                  &\multicolumn{3}{c|}{10\%}             &\multicolumn{3}{c|}{20\%}     &\multirow{2}{*}{Time(s)} \\
\cline{2-11}
                      &Method             &PSNR         & SSIM        &FSIM          &PSNR       & SSIM        &FSIM        &PSNR       & SSIM        &FSIM \\

                      \hline

\multirow{8}{*}{\tabincell{c}{\emph{Washington} \\ \emph{DC Mall}\\$256\times256\times150$}}

                      &HaLRTC            &20.72        & 0.452        & 0.665       &24.74       &0.656        &0.798      & 29.38    & 0.848    & 0.909  &\bf{76.487}   \\

                      &LRTC-TVI          &21.93        & 0.437        & 0.605 	    &25.89       &0.638        &0.759      & 29.11    & 0.824    & 0.893  &2348.2\\

                      &BCPF              &29.07        & 0.820        & 0.895 	    &31.89   	 &0.895 	   &0.934      & 32.77    & 0.911    & 0.943  &2955.9\\

                      &logDet            &25.22        & 0.685        & 0.848       &32.50       &0.911        &0.947      & 37.99    & 0.969    & 0.981  &237.18   \\

                      &TNN               &28.87        & 0.831        & 0.907       &32.41       &0.913        &0.949      & 36.85    & 0.963    & 0.977  &294.46  \\

                      &PSTNN             &28.15        & 0.793        & 0.886       &32.63       &0.911        &0.946      & 37.39    & 0.965    & 0.978   &306.16   \\

                      &t-TNN             &33.23        & 0.932        & 0.959       &43.96       &0.994        &0.996      & 56.99    & 0.997    & 0.998   &184.23   \\

                      &WSTNN            &\bf{40.54}  &\bf{0.988}    &\bf{0.992}   &\bf{50.31}  &\bf{0.999}   &\bf{0.999}   &\bf{58.89}&\bf{0.999}&\bf{0.999}&544.26  \\

                      \hline

 \multirow{8}{*}{\tabincell{c}{\emph{Pavia} \\ \emph{University}\\$256\times256\times87$}}

                      &HaLRTC            &15.01        & 0.043        & 0.517       &24.02       &0.611        &0.736     & 27.59      & 0.788    & 0.861  &\bf{49.745}\\

                      &LRTC-TVI          &23.26        & 0.554        & 0.652 	    &25.80       &0.713        &0.785     & 29.19      & 0.866    & 0.903  &1427.3 \\

                      &BCPF              &27.64        & 0.726        & 0.835 	    &30.39   	 &0.836 	   &0.898     & 32.07      & 0.884    & 0.928  &1603.6 \\

                      &logDet            &26.90        & 0.684        & 0.835       &32.69       &0.876        &0.932     & 39.34      & 0.959    & 0.977  &140.96 \\

                      &TNN               &28.12        & 0.750        & 0.865       &32.15       &0.874        &0.931     & 37.49      & 0.950    & 0.972  &168.44 \\

                      &PSTNN             &23.18        & 0.449        & 0.737       &32.97       &0.872        &0.932     & 38.84      & 0.955    & 0.974  &181.04  \\

                      &t-TNN             &33.38        & 0.928        & 0.957       &41.15       &0.988        &0.993     & 50.83      & 0.997    & 0.998  &101.49  \\

                      &WSTNN            &\bf{37.26}  &\bf{0.976}    &\bf{0.983}   &\bf{44.48}  &\bf{0.995}   &\bf{0.997}  &\bf{53.92} &\bf{0.999}&\bf{0.999}&258.78 \\

 \Xhline{1pt}
\end{tabular}
\end{center}\vspace{-0.5cm}
\label{HSITC}
\end{table}

\textbf{MSI completion.} We test 32 MSIs in the dataset CAVE\footnote{\url{http://www.cs.columbia.edu/CAVE/databases/multispectral}.}. All testing data are of size $256\times256\times31$. Table \ref{MSITC} lists the mean values of PSNR, SSIM, and FSIM of all 32 MSIs recovered by different LRTC methods. As observed, the proposed method can significantly outperform the compared methods in terms of all evaluation indices. In Fig. \ref{MSIqua}, we show the PSNR, SSIM, and FSIM values of each  MSI output by eight utilized LRTC methods with $\text{SR}=10\%$. We can see that in all cases, the proposed method has an overall better performance than the compared ones with respect to all evaluation indices. To illustrate the visual quality, in Fig. \ref{imageMSIfig}, we show one band in four testing data recovered by different methods with $\text{SR}=10\%$. It is obvious that the proposed method is evidently superior to the compared ones, in recovery of both abundant shape structure and texture information.

\begin{figure}[!t]
\setlength{\tabcolsep}{1pt}
\begin{center}
\begin{tabular}{ccc}
\multicolumn{3}{c}{~~\includegraphics[width=0.9\textwidth]{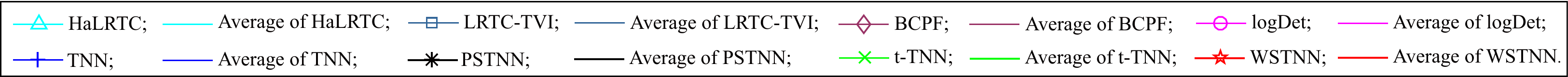}}\vspace{0.1cm}\\
\includegraphics[width=4.8cm, height=3.8cm]{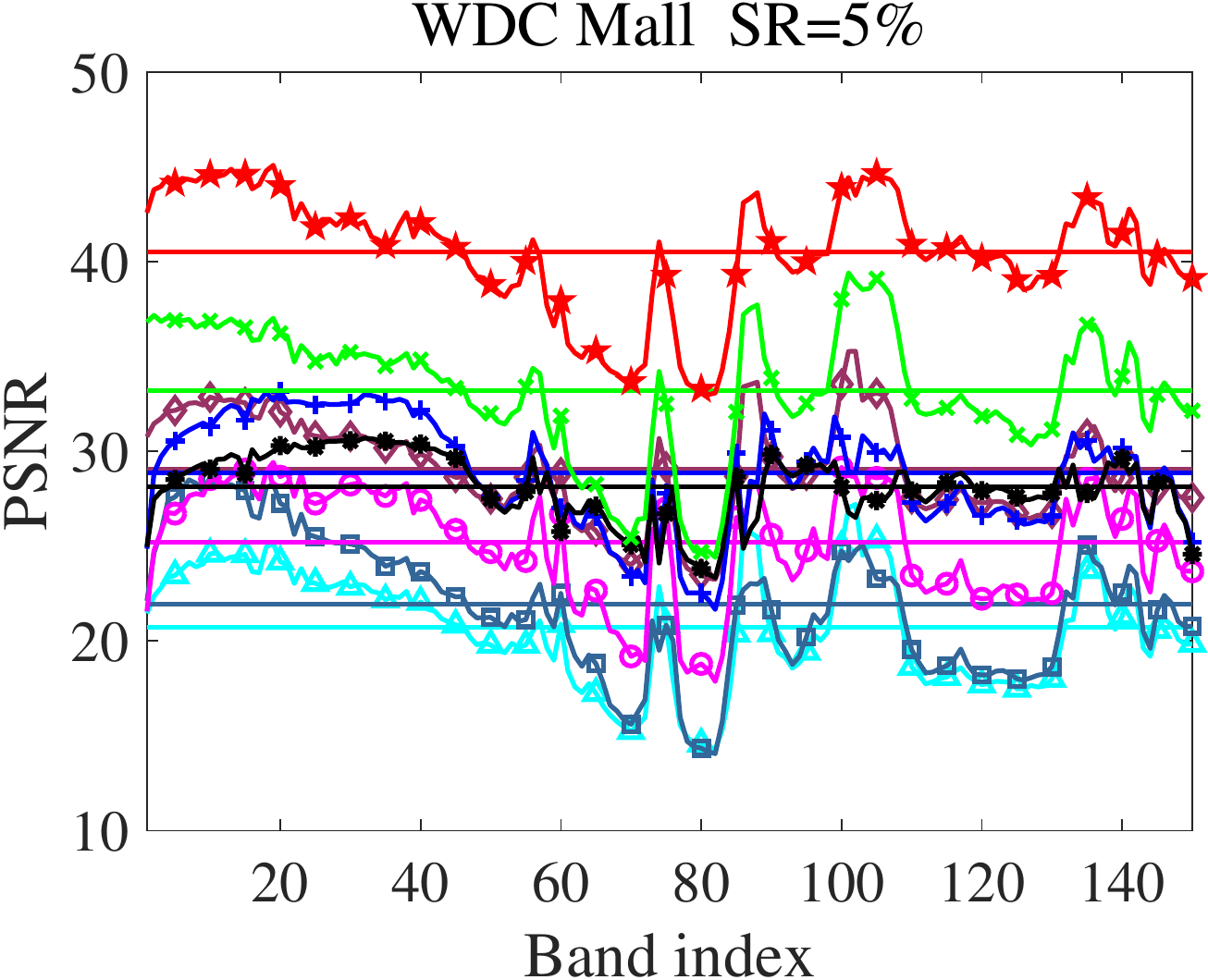}&~~~~~~
\includegraphics[width=4.8cm, height=3.8cm]{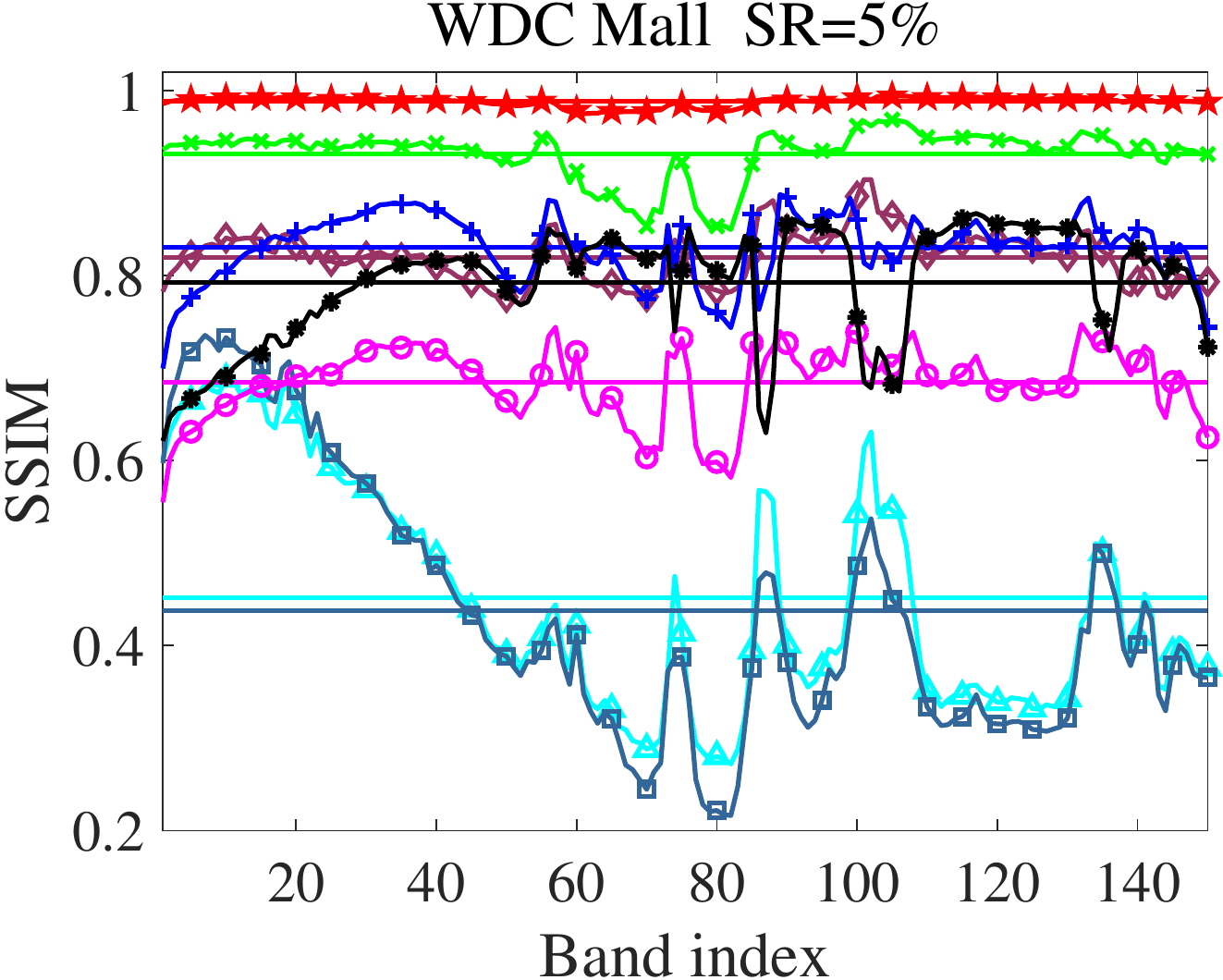}&~~~~~~
\includegraphics[width=4.8cm, height=3.8cm]{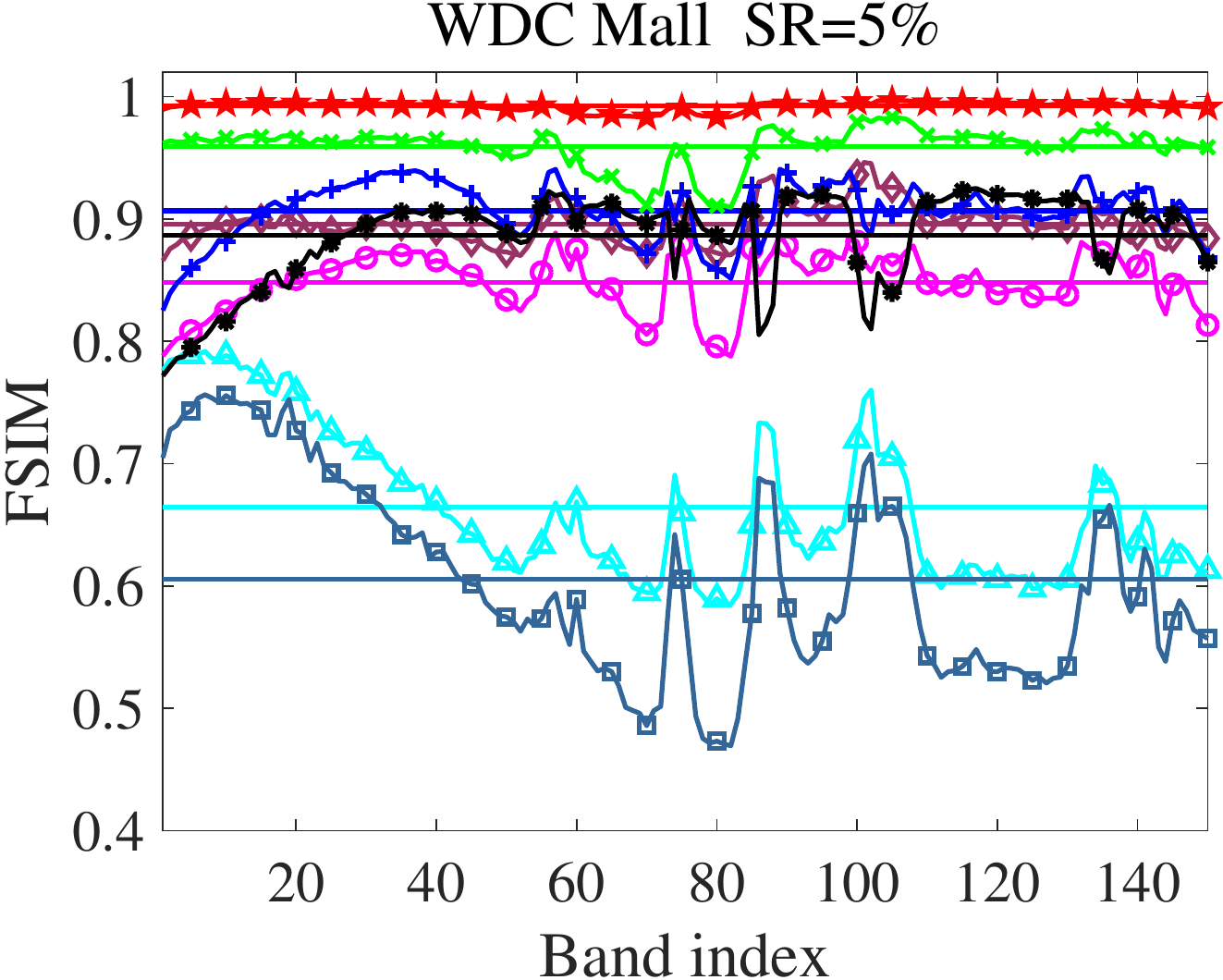}~~
\end{tabular}
\caption{The PSNR, SSIM, and FSIM values of each band of the recovered HSI \emph{Washington DC Mall} output by eight LRTC methods with $\text{SR}=5\%$.}
\label{imageWDCquality}\vspace{-0.5cm}
\end{center}
\end{figure}

\begin{table}[!t]
\scriptsize
\setlength{\tabcolsep}{8.15pt}
\renewcommand\arraystretch{1.2}
\caption{The  PSNR, SSIM, and FSIM values output by eight utilized LRTC methods for MRI.}
\begin{center}
\begin{tabular}{c|ccc|ccc|ccc|c}

 \Xhline{1pt}

                       SR        &\multicolumn{3}{c|}{5\%}  &\multicolumn{3}{c|}{10\%}  & \multicolumn{3}{c|}{20\%}  &\multirow{2}{*}{Time(s)}\\

                      \cline{1-10}

                      Method             &PSNR         & SSIM        &FSIM          &PSNR       & SSIM        &FSIM        &PSNR        & SSIM      &FSIM  \\

                      \hline

                      HaLRTC            & 15.40        & 0.241       & 0.608       & 19.03     & 0.390       & 0.699       & 24.30     & 0.653     & 0.826  & \bf{69.981}\\

                      LRTC-TVI          & 19.36        & 0.597       & 0.702	   &22.84      & 0.748       & 0.805       &28.19      &0.891      & 0.908  & 1473.8\\

                      BCPF              & 22.37        & 0.426       & 0.734 	   &23.81      & 0.495 	     & 0.758       &24.96      &0.552      & 0.779  & 1525.6\\

                      logDet            & 18.32        & 0.283       & 0.654       & 25.36     & 0.596       & 0.791       & 31.22     & 0.823     & 0.892  & 165.90\\

                      TNN               & 22.71        & 0.472       & 0.743       & 26.06     & 0.642       & 0.811       & 29.99     & 0.799     & 0.881  & 165.85\\

                      PSTNN             & 20.39        & 0.288       & 0.629       & 26.45     & 0.621       & 0.802       & 30.71     & 0.805     & 0.885  & 209.19\\

                      t-TNN             & 22.78        & 0.460       & 0.736       & 26.42     & 0.649       & 0.816       & 30.58     & 0.816     & 0.890  & 170.04\\

                      WSTNN            &\bf{25.60}  &\bf{0.714}    &\bf{0.827}   &\bf{29.02}  &\bf{0.835}   &\bf{0.887}    & \bf{33.46}   & \bf{0.931}  & \bf{0.941} &405.01\\

  \Xhline{1pt}
\end{tabular}
\end{center}\vspace{-0.5cm}
\label{MRITC}
\end{table}

\textbf{HSI completion.} We test HSIs \emph{Washington DC Mall}\footnote{\url{http://lesun.weebly.com/hyperspectral-data-set.html}.\label{HSIweb}} and \emph{Pavia University}\textsuperscript{\ref {HSIweb}}. Table \ref{HSITC} lists the values of PSNR, SSIM, and FSIM of these two testing HSIs recovered by different LRTC methods. We observe that compared with other methods, the proposed method consistently achieves the highest values in terms of all evaluation indexes, e.g., when SR is set as $5\%$ or $10\%$, the proposed method achieves around $7$ dB gain in PSNR beyond the second best method in the test on \emph{Washington DC Mall}.
For visual comparison, in Fig. \ref{imageWDCfig}, we show one band in these two testing HSIs recovered by eight utilized LRTC methods with $\text{SR}=5\%$.
As observed, the proposed method is able to produce visually superior results than the compared methods. Fig. \ref{imageWDCquality} shows the PSNR, SSIM and FSIM values of each band of the recovered HSI \emph{Washington DC Mall} obtained by eight compared LRTC methods with $\text{SR}=5\%$. From this figure, it is easy to observe that the proposed method achieves the best performance in all bands among eight LRTC methods.

\begin{figure}[!t]
\tiny
\setlength{\tabcolsep}{0.9pt}
\begin{center}
\begin{tabular}{cccccccccc}
\includegraphics[width=0.096\textwidth]{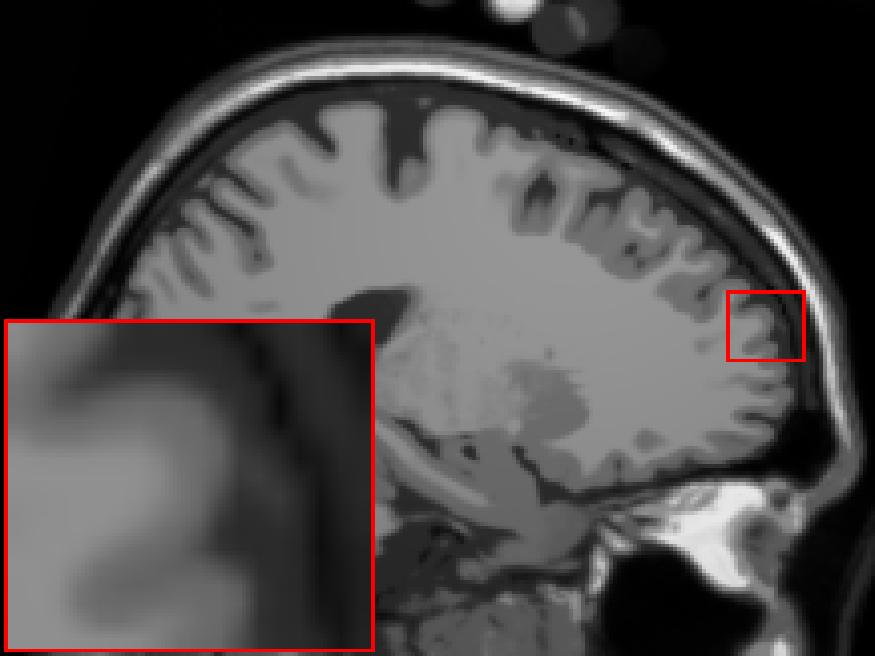}&
\includegraphics[width=0.096\textwidth]{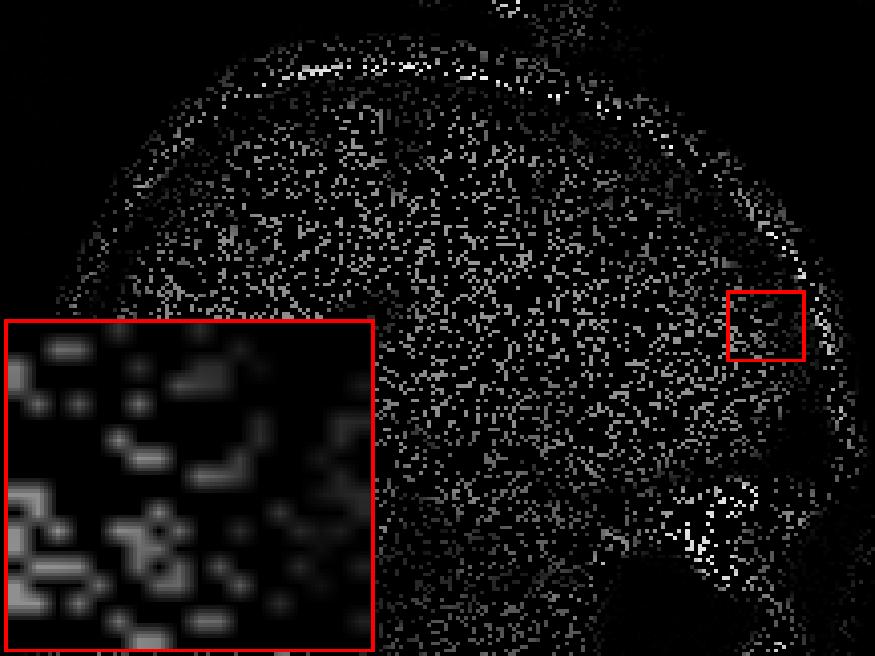}&
\includegraphics[width=0.096\textwidth]{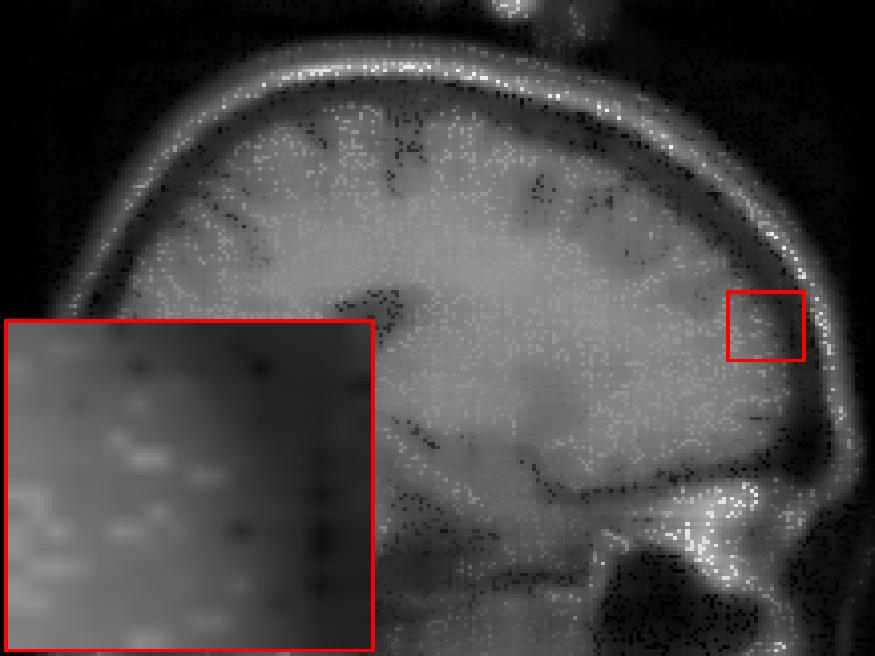}&
\includegraphics[width=0.096\textwidth]{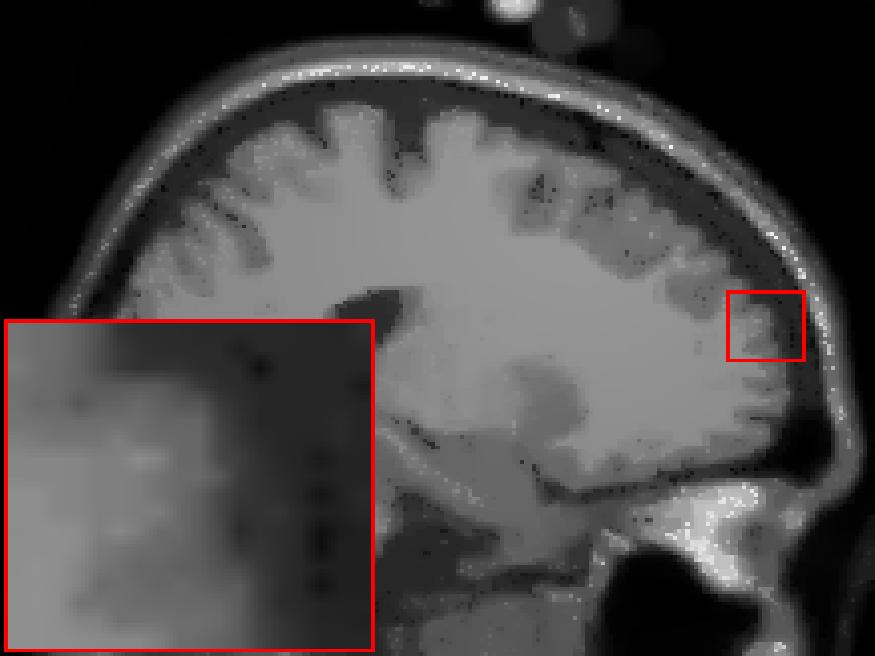}&
\includegraphics[width=0.096\textwidth]{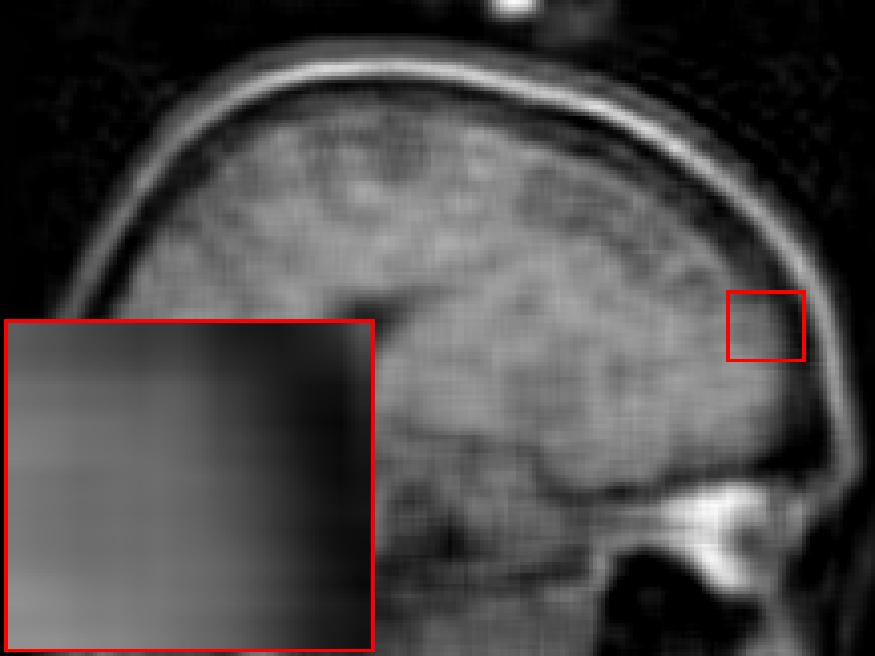}&
\includegraphics[width=0.096\textwidth]{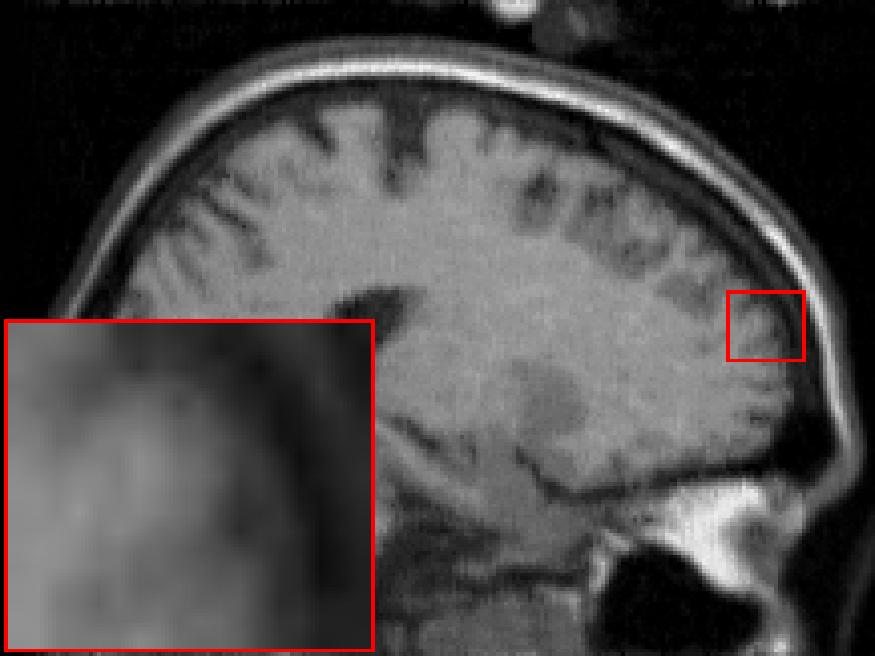}&
\includegraphics[width=0.096\textwidth]{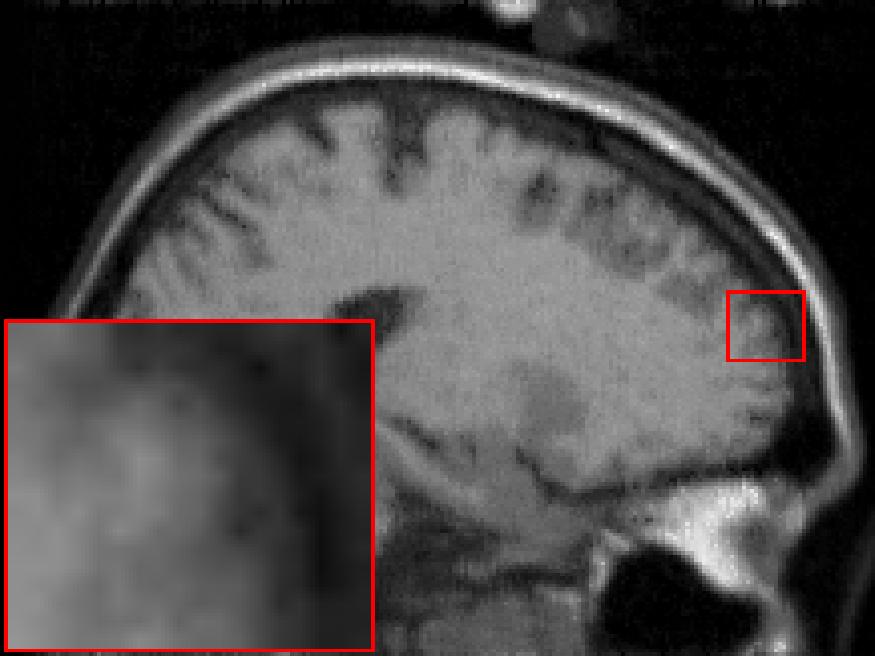}&
\includegraphics[width=0.096\textwidth]{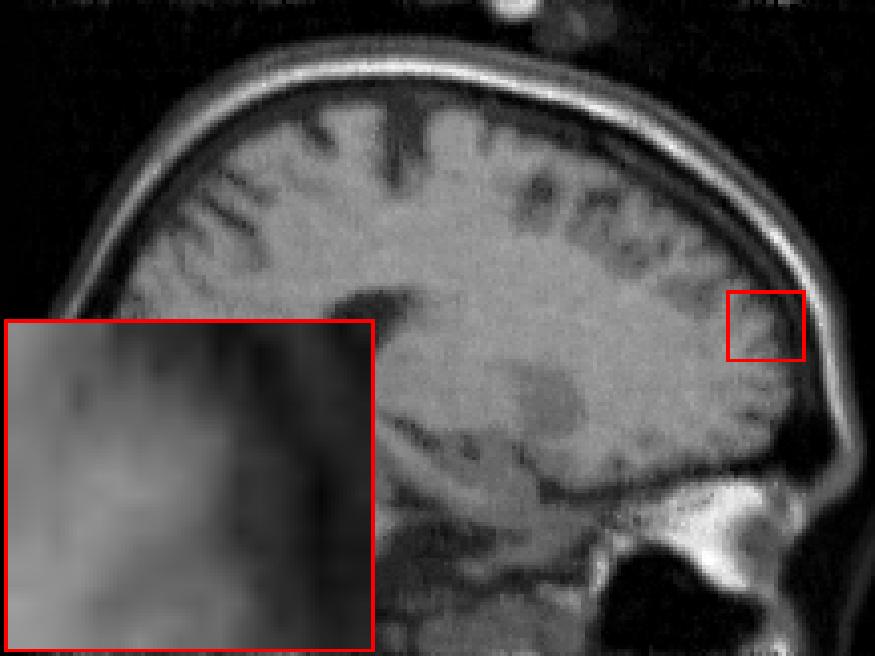}&
\includegraphics[width=0.096\textwidth]{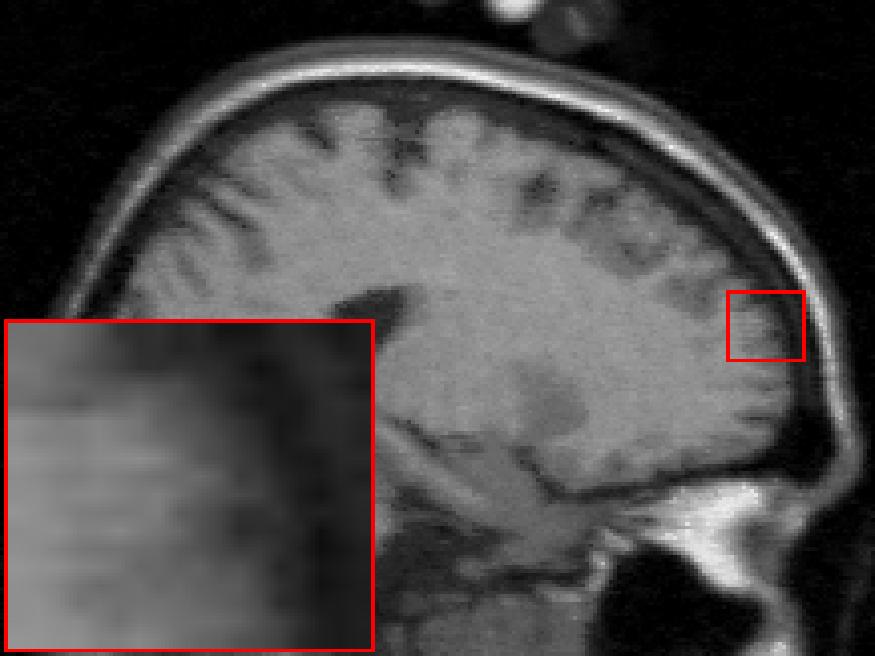}&
\includegraphics[width=0.096\textwidth]{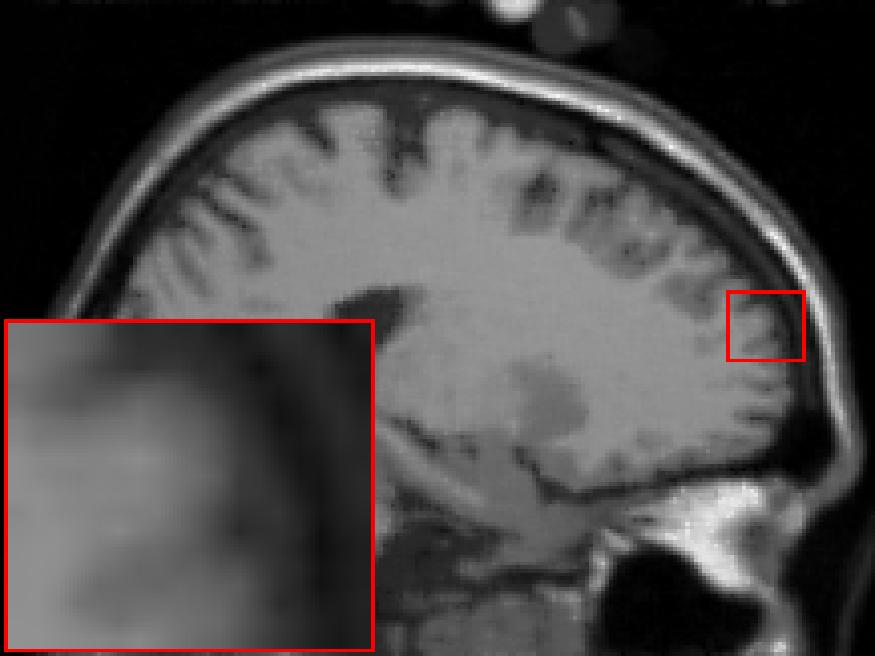}\\
\includegraphics[width=0.096\textwidth]{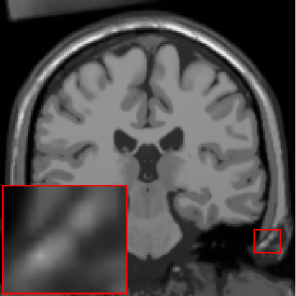}&
\includegraphics[width=0.096\textwidth]{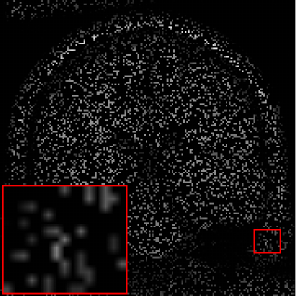}&
\includegraphics[width=0.096\textwidth]{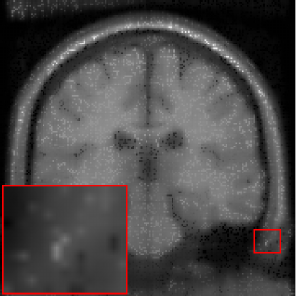}&
\includegraphics[width=0.096\textwidth]{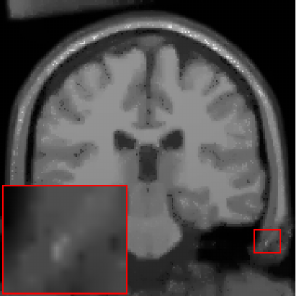}&
\includegraphics[width=0.096\textwidth]{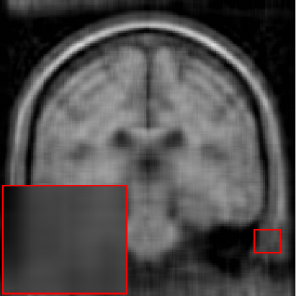}&
\includegraphics[width=0.096\textwidth]{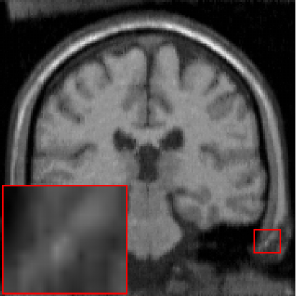}&
\includegraphics[width=0.096\textwidth]{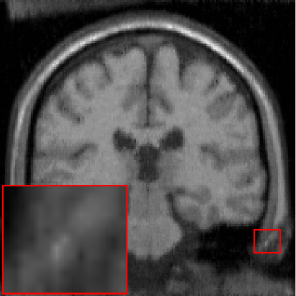}&
\includegraphics[width=0.096\textwidth]{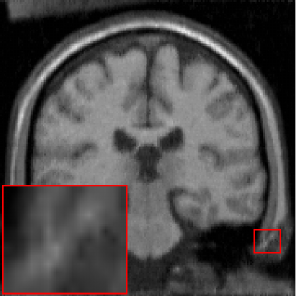}&
\includegraphics[width=0.096\textwidth]{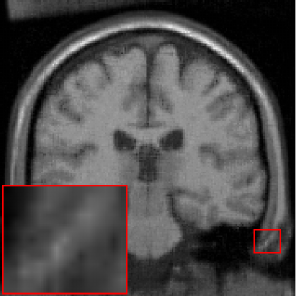}&
\includegraphics[width=0.096\textwidth]{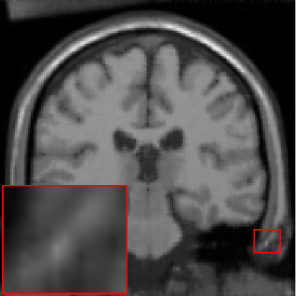}\\
\includegraphics[width=0.096\textwidth]{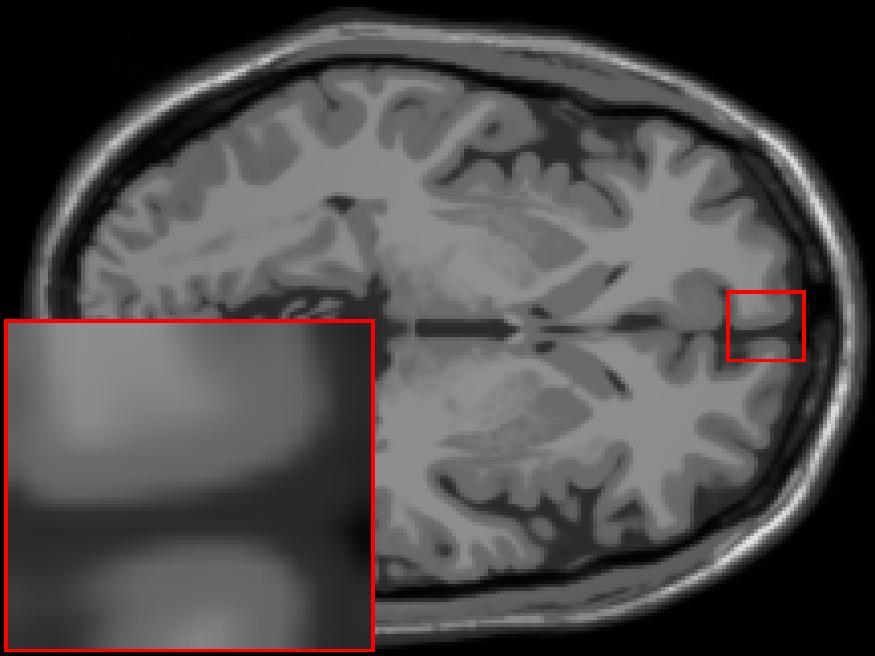}&
\includegraphics[width=0.096\textwidth]{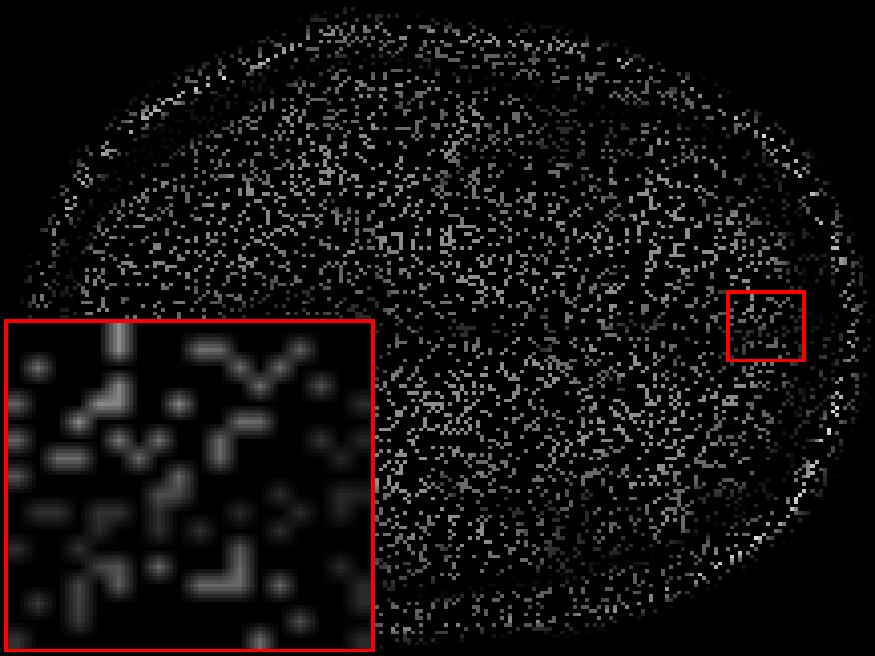}&
\includegraphics[width=0.096\textwidth]{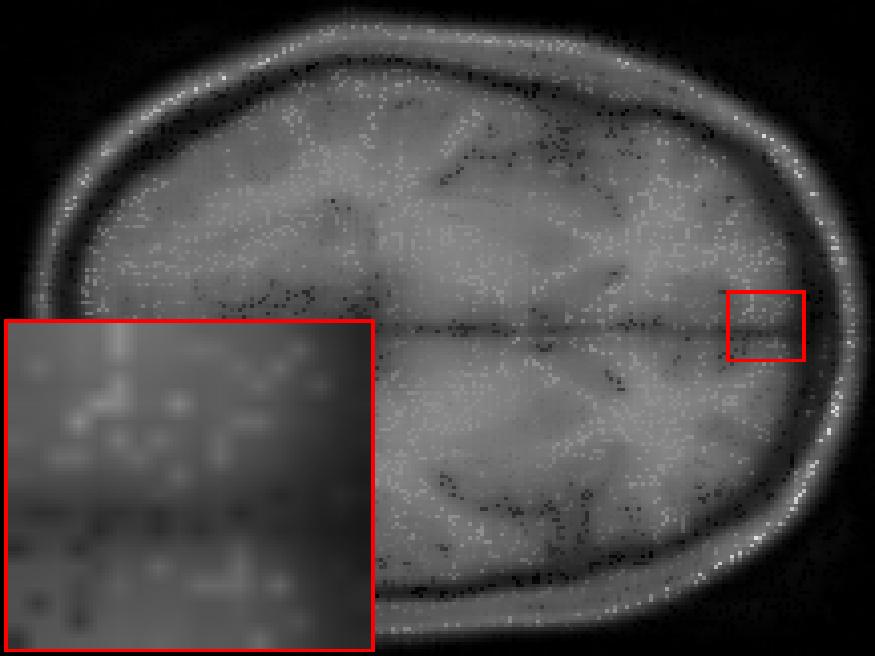}&
\includegraphics[width=0.096\textwidth]{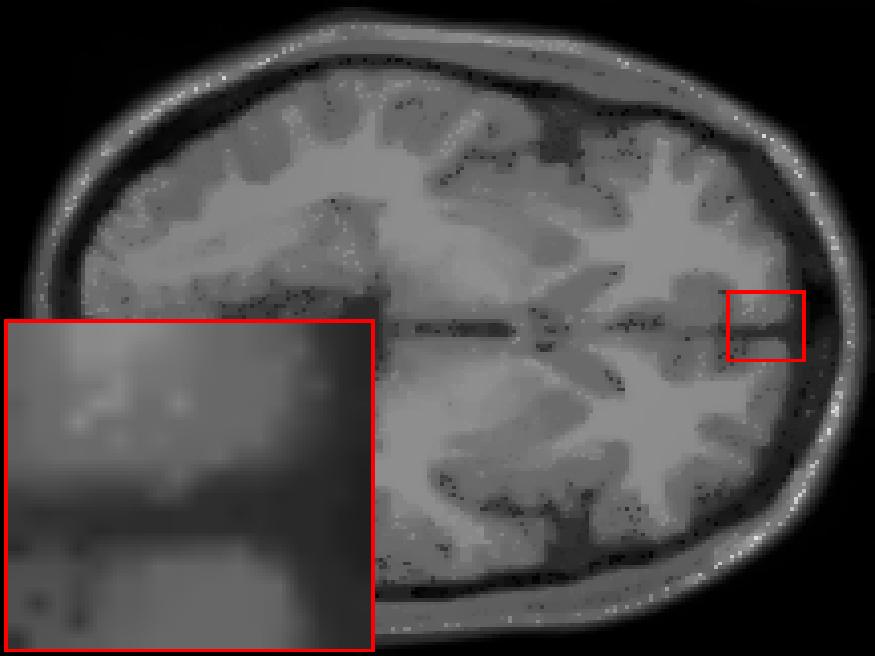}&
\includegraphics[width=0.096\textwidth]{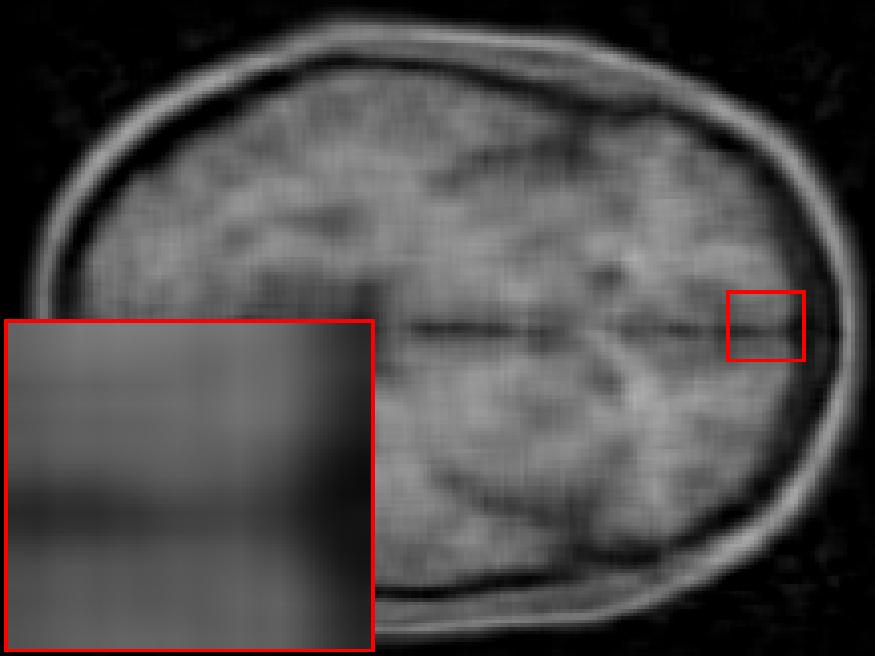}&
\includegraphics[width=0.096\textwidth]{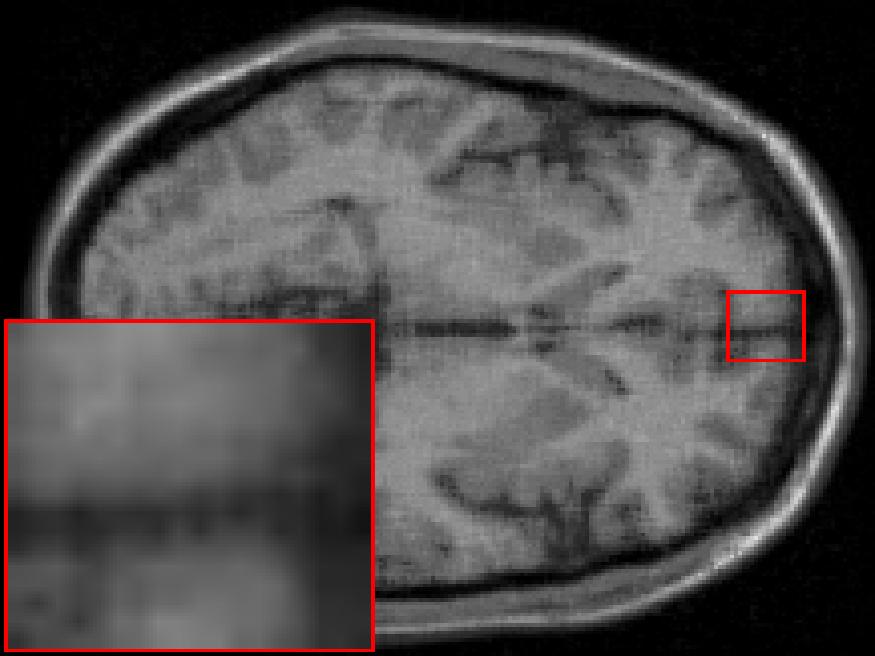}&
\includegraphics[width=0.096\textwidth]{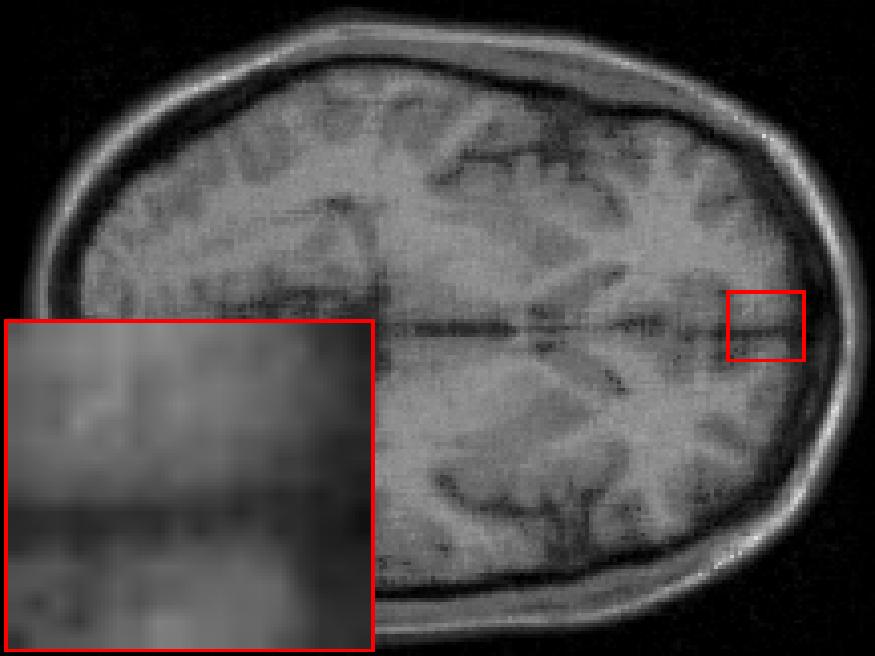}&
\includegraphics[width=0.096\textwidth]{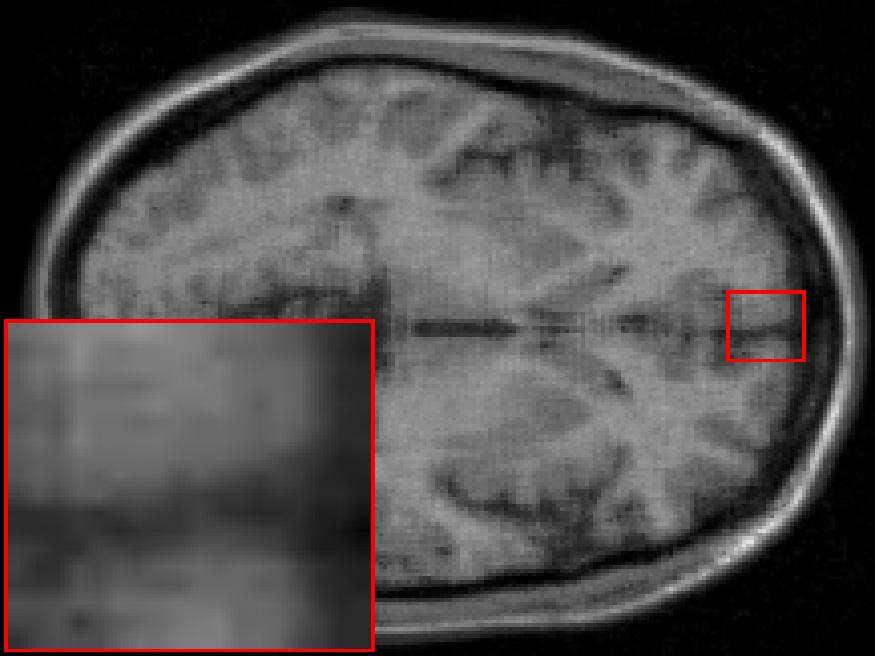}&
\includegraphics[width=0.096\textwidth]{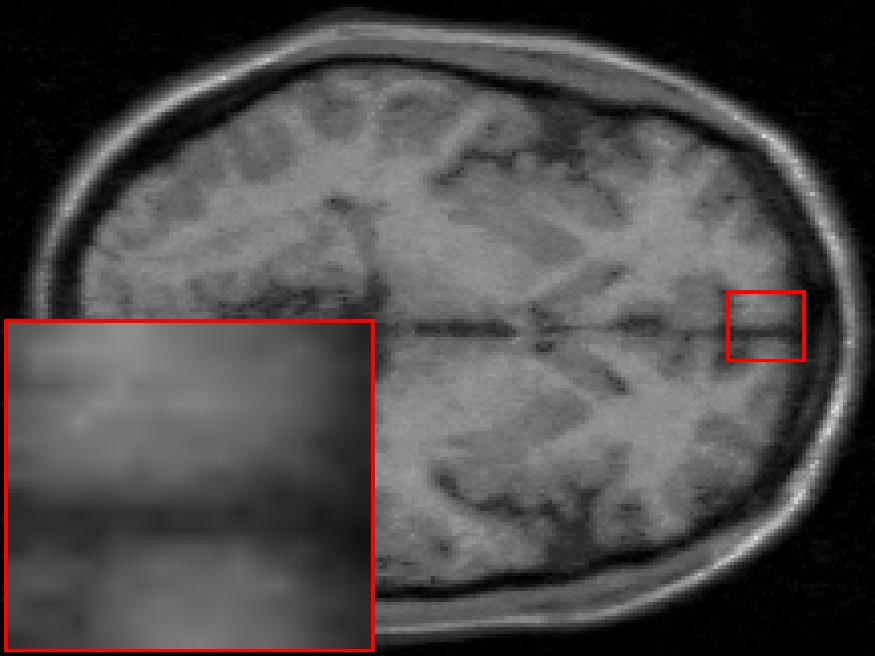}&
\includegraphics[width=0.096\textwidth]{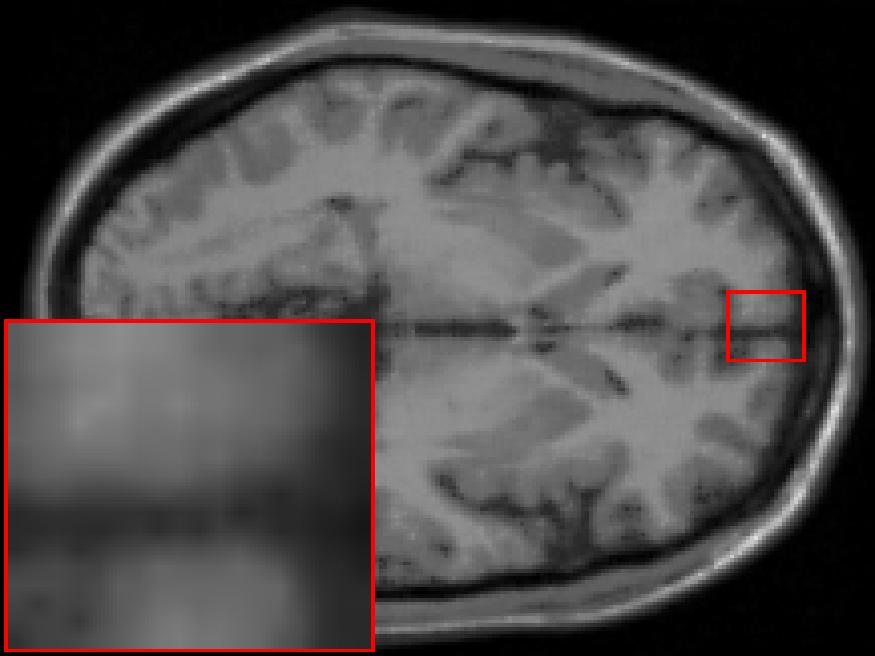}\\
Original& Observed & HaLRTC \cite{Liu2013PAMItensor}& LRTC-TVI \cite{Li2017LowRankTC} & BCPF \cite{ZhangCP} & logDet \cite{JilogDet} & TNN \cite{zhang2017exact}& PSTNN \cite{jiang2017novel2} & t-TNN \cite{HutTNN} & WSTNN
  \end{tabular}
  \caption{The completion results of the MRI data with $\text{SR}=20\%$. From top to bottom: the images located at the 70-th horizonal slice, the 100-th lateral slice, and the 70-th frontal slice, respectively.}
  \label{imageMRIfig}\vspace{-0.2cm}
  \end{center}
\end{figure}

\begin{figure}[!t]
\tiny
\setlength{\tabcolsep}{0.9pt}
\begin{center}
\begin{tabular}{cccccccccc}
\includegraphics[width=0.096\textwidth]{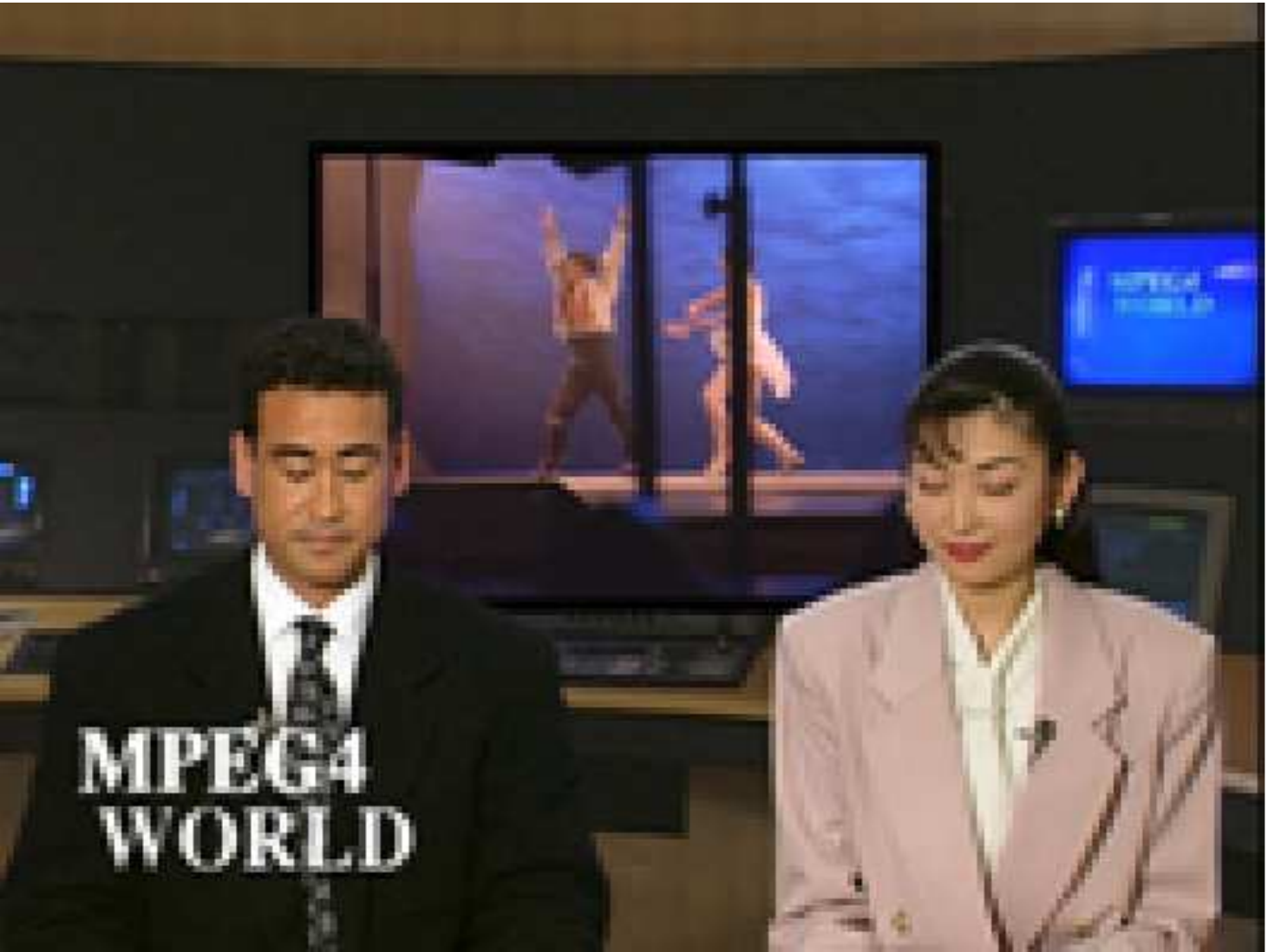}&
\includegraphics[width=0.096\textwidth]{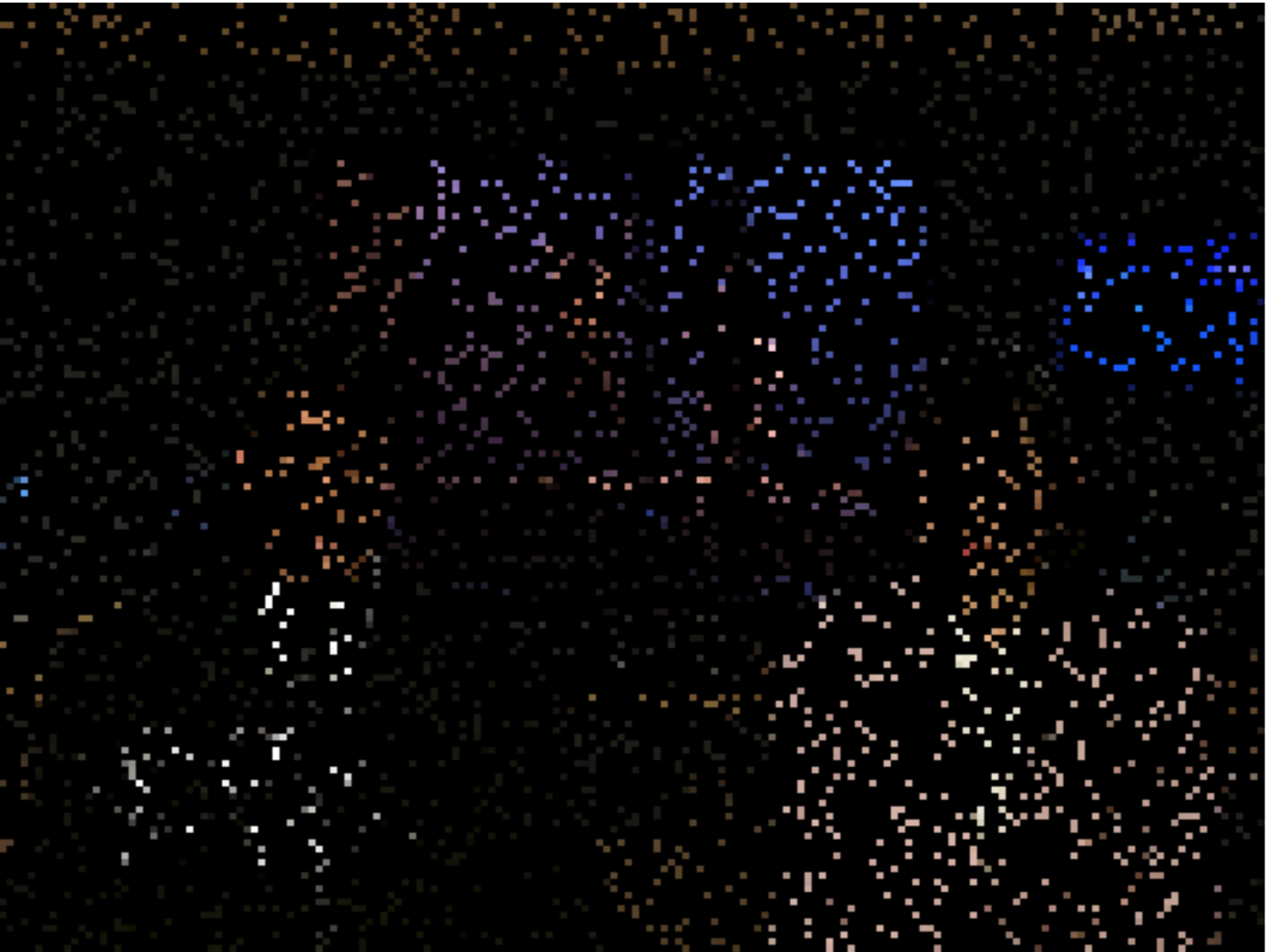}&
\includegraphics[width=0.096\textwidth]{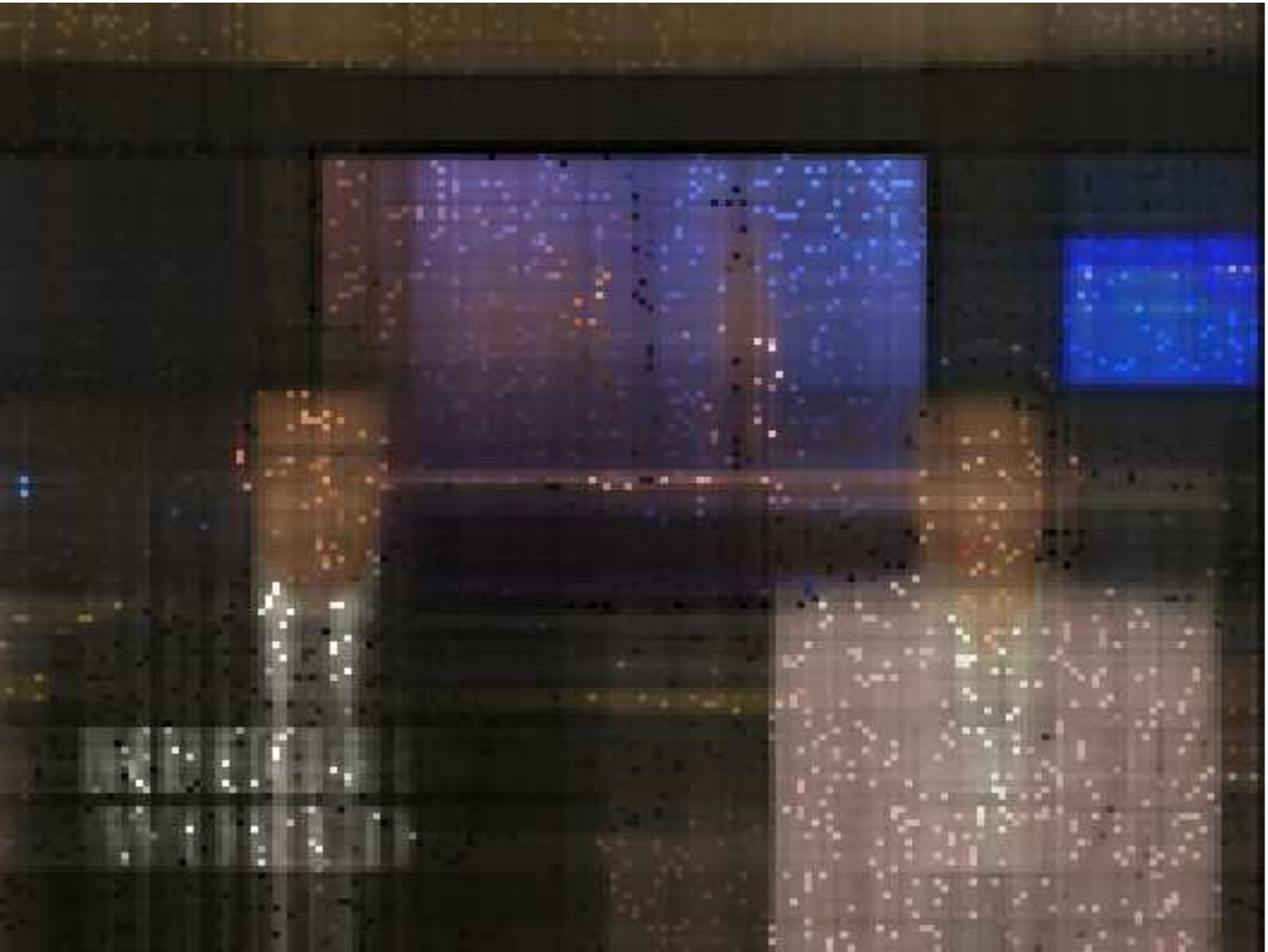}&
\includegraphics[width=0.096\textwidth]{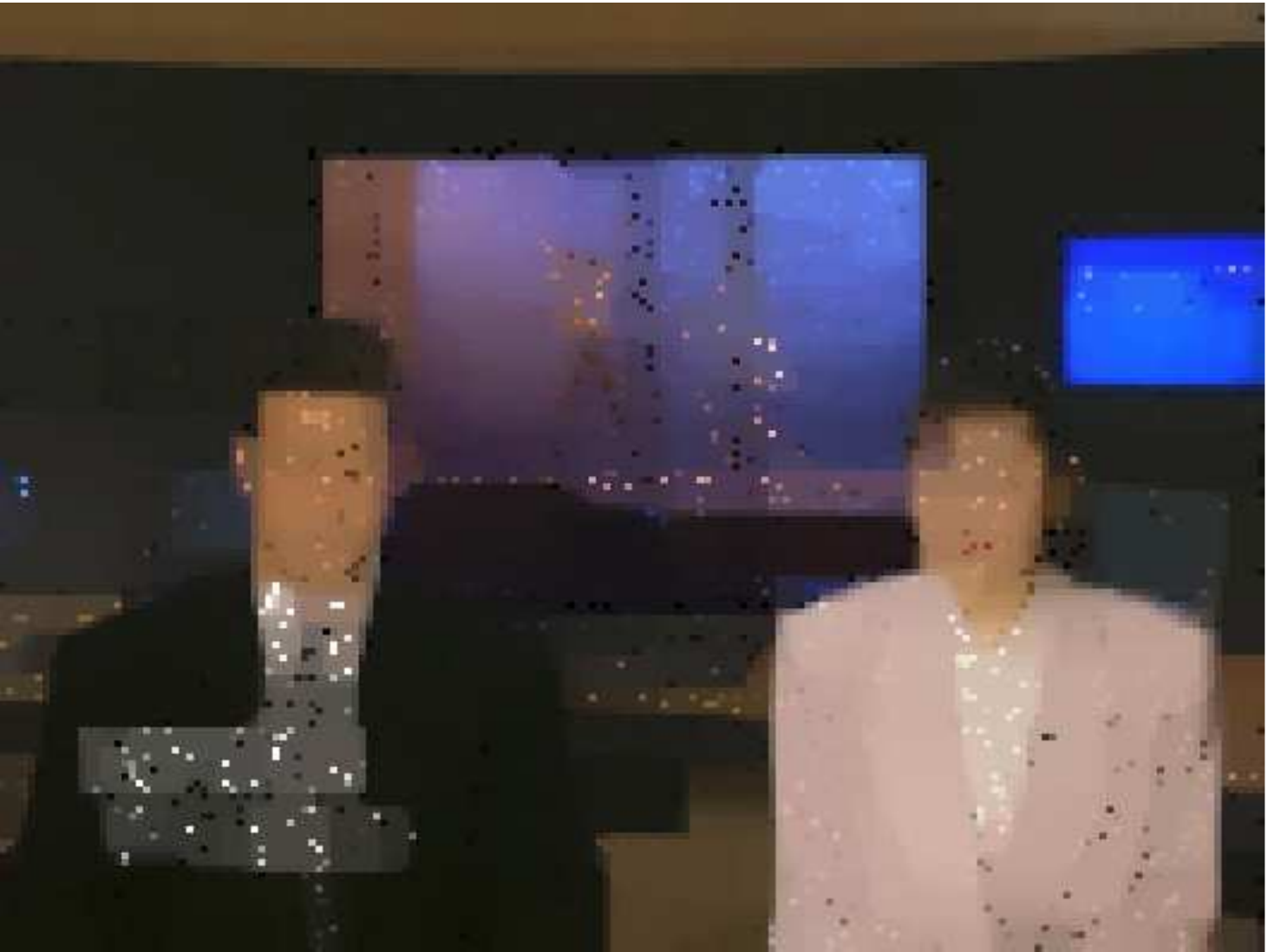}&
\includegraphics[width=0.096\textwidth]{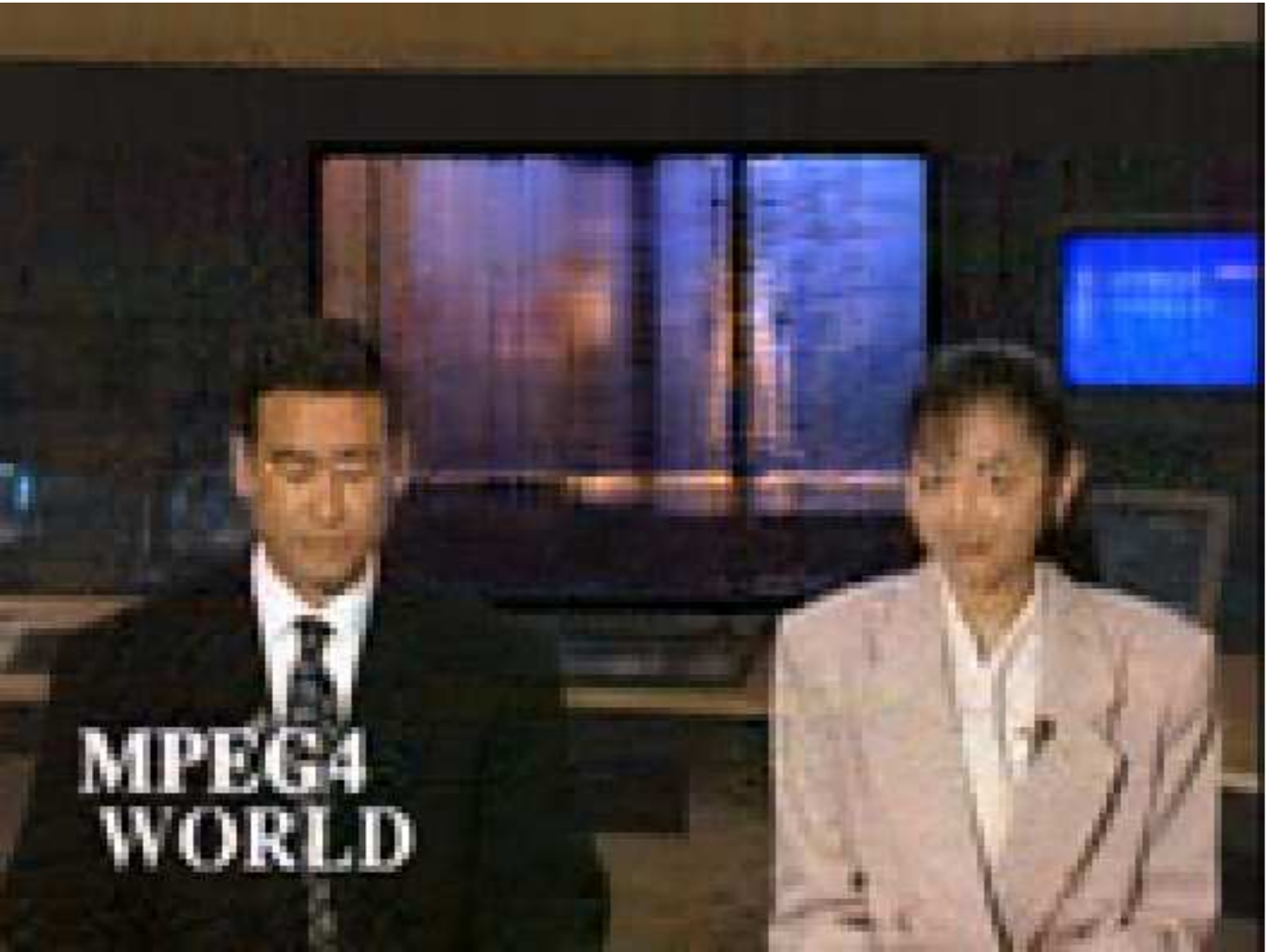}&
\includegraphics[width=0.096\textwidth]{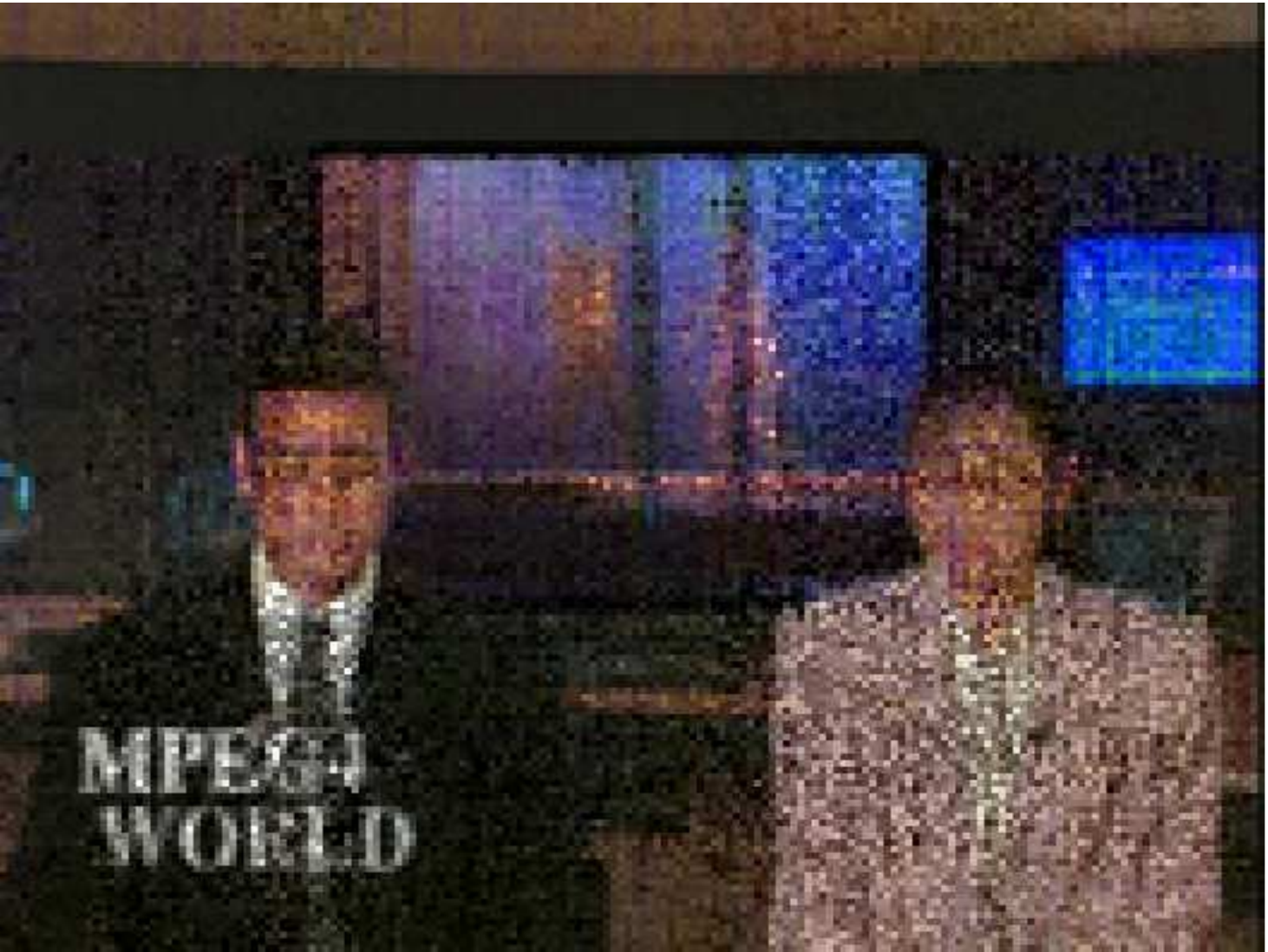}&
\includegraphics[width=0.096\textwidth]{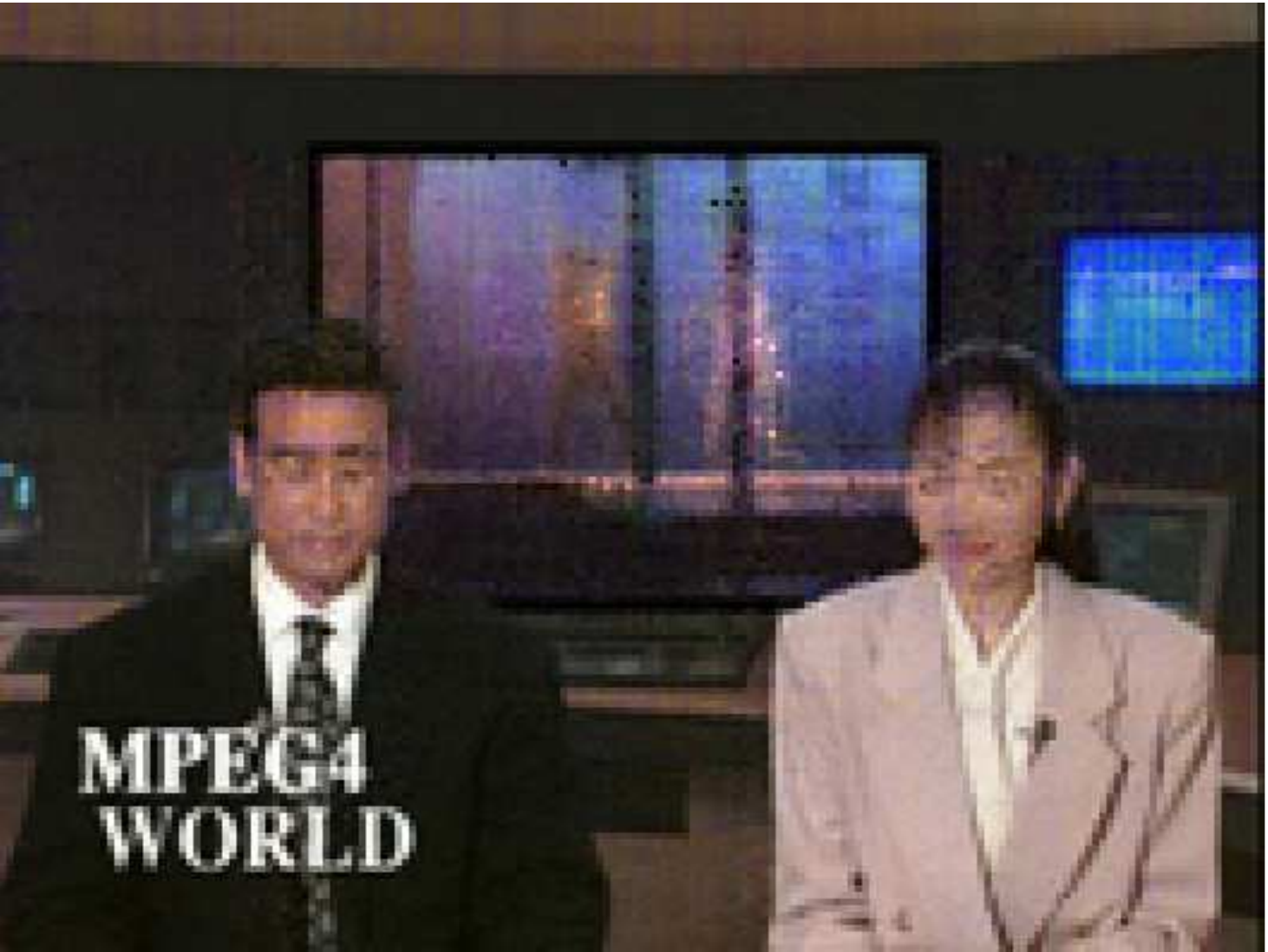}&
\includegraphics[width=0.096\textwidth]{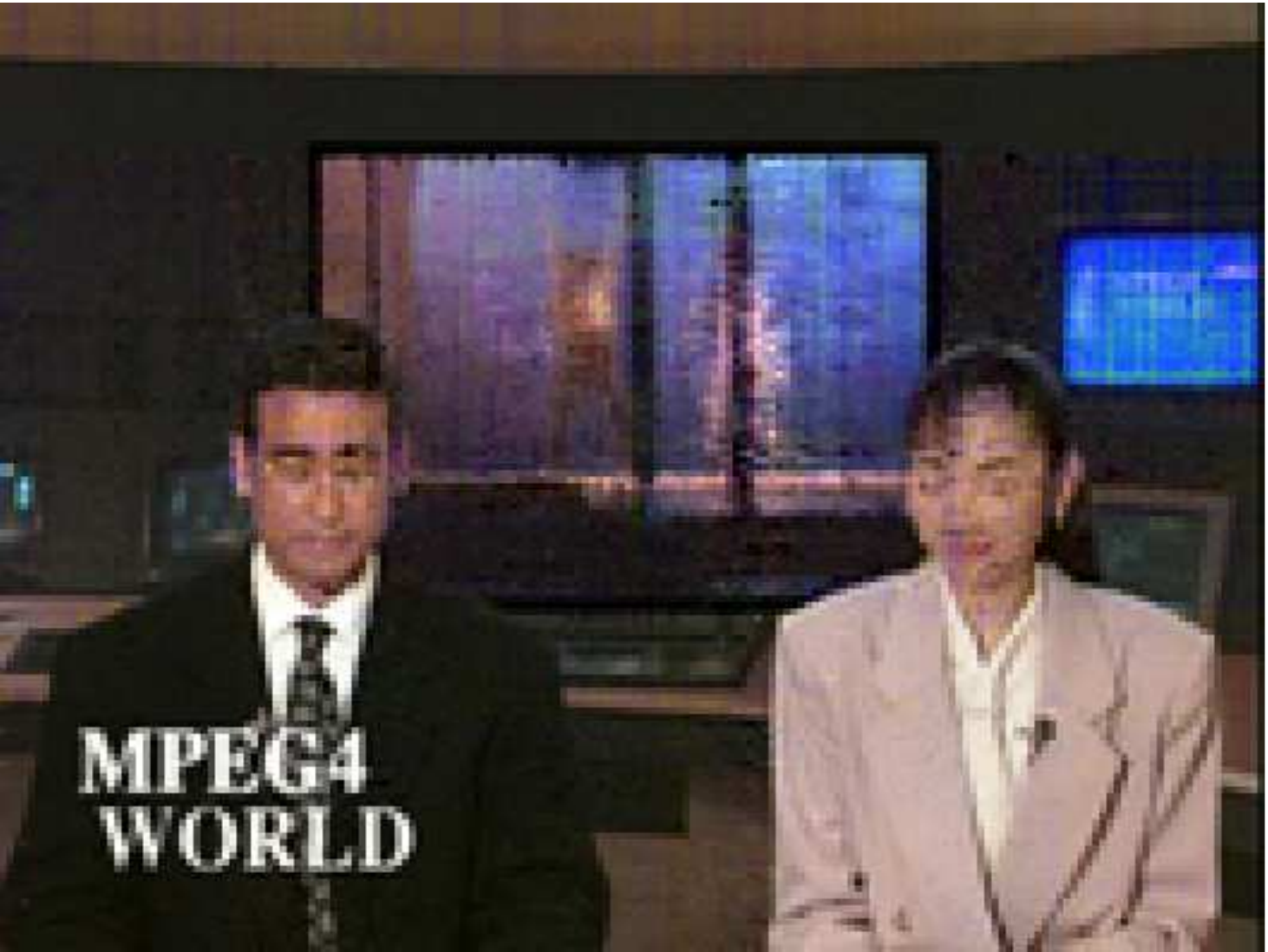}&
\includegraphics[width=0.096\textwidth]{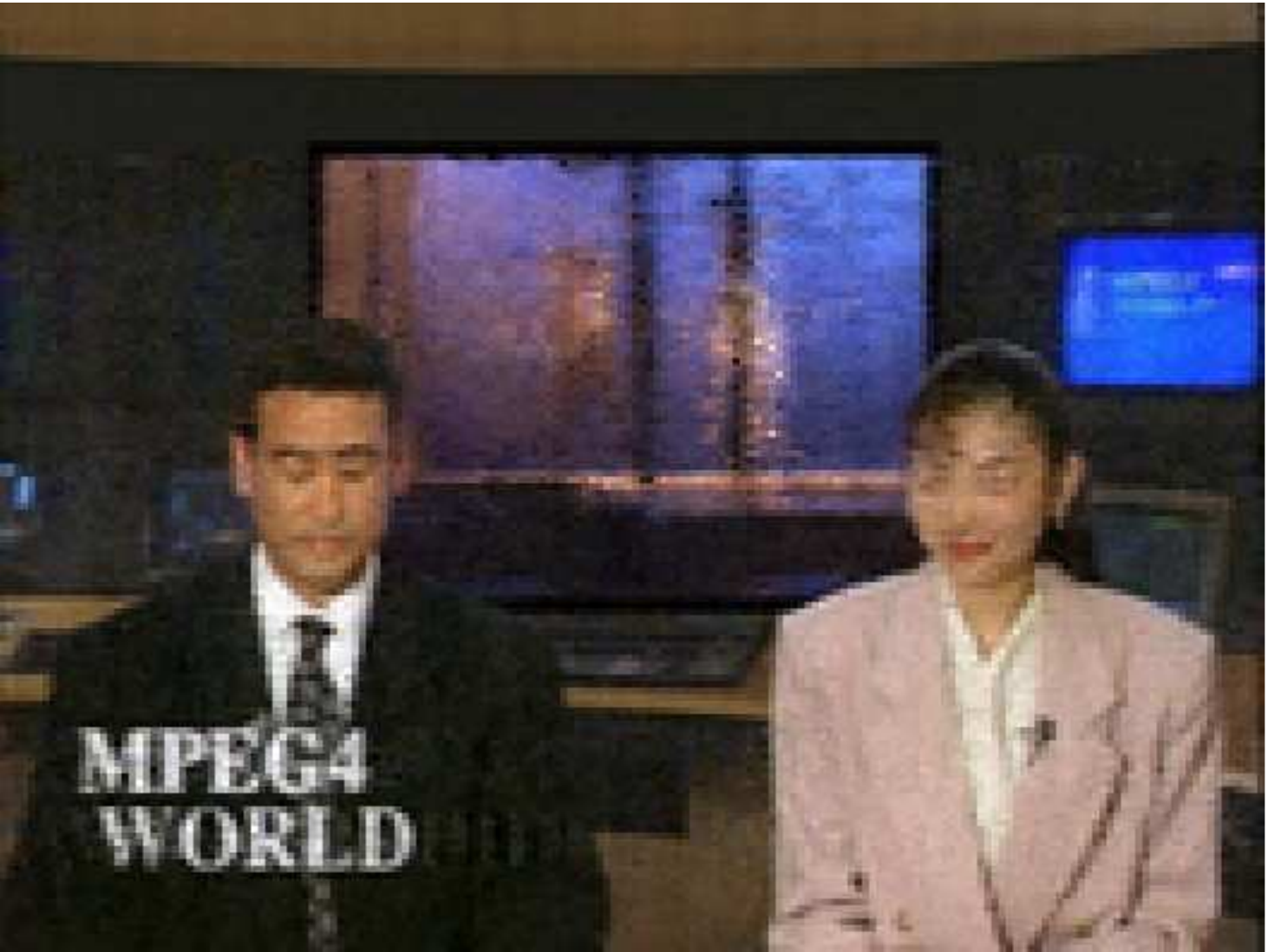}&
\includegraphics[width=0.096\textwidth]{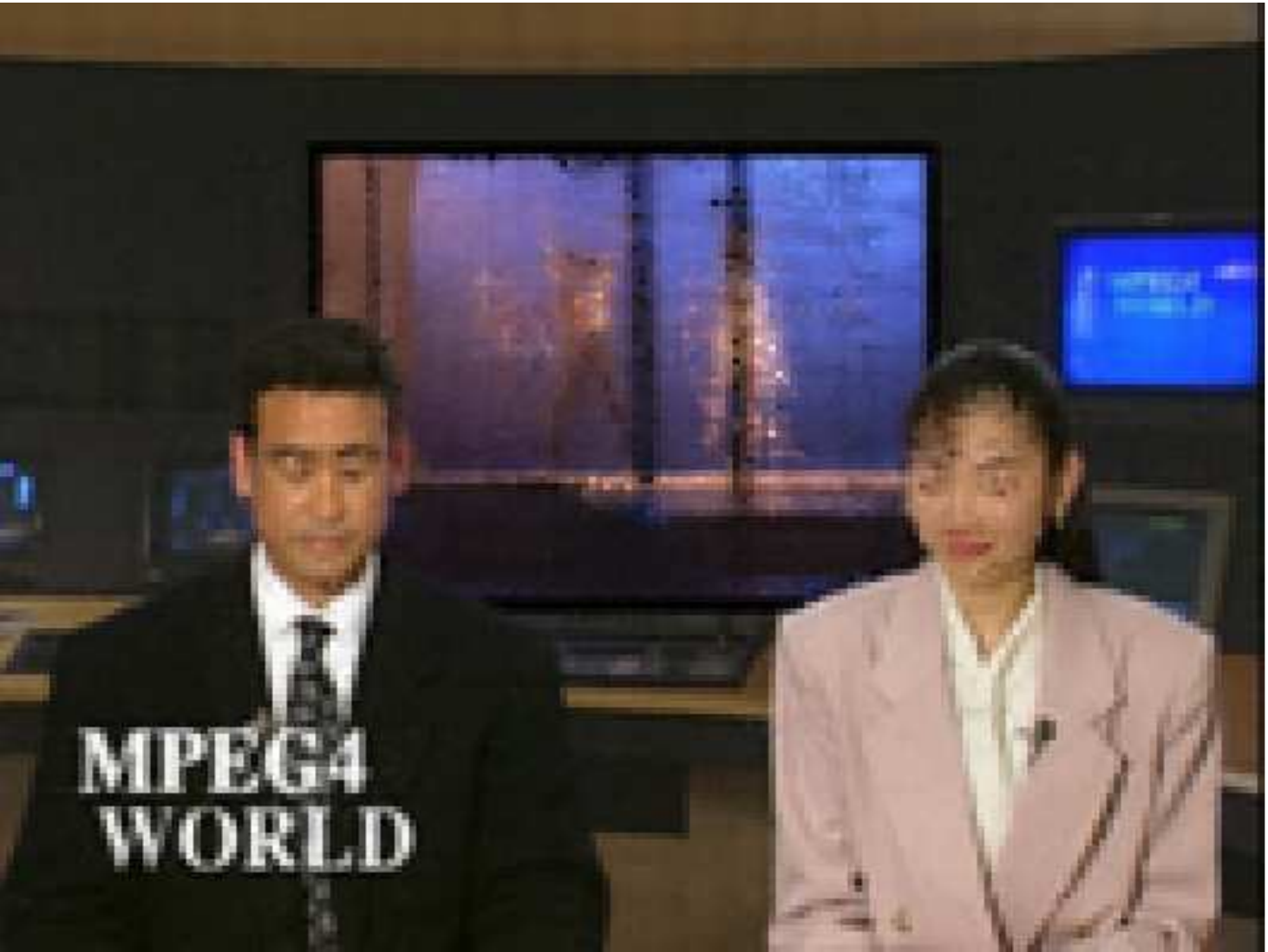}\\
\includegraphics[width=0.096\textwidth]{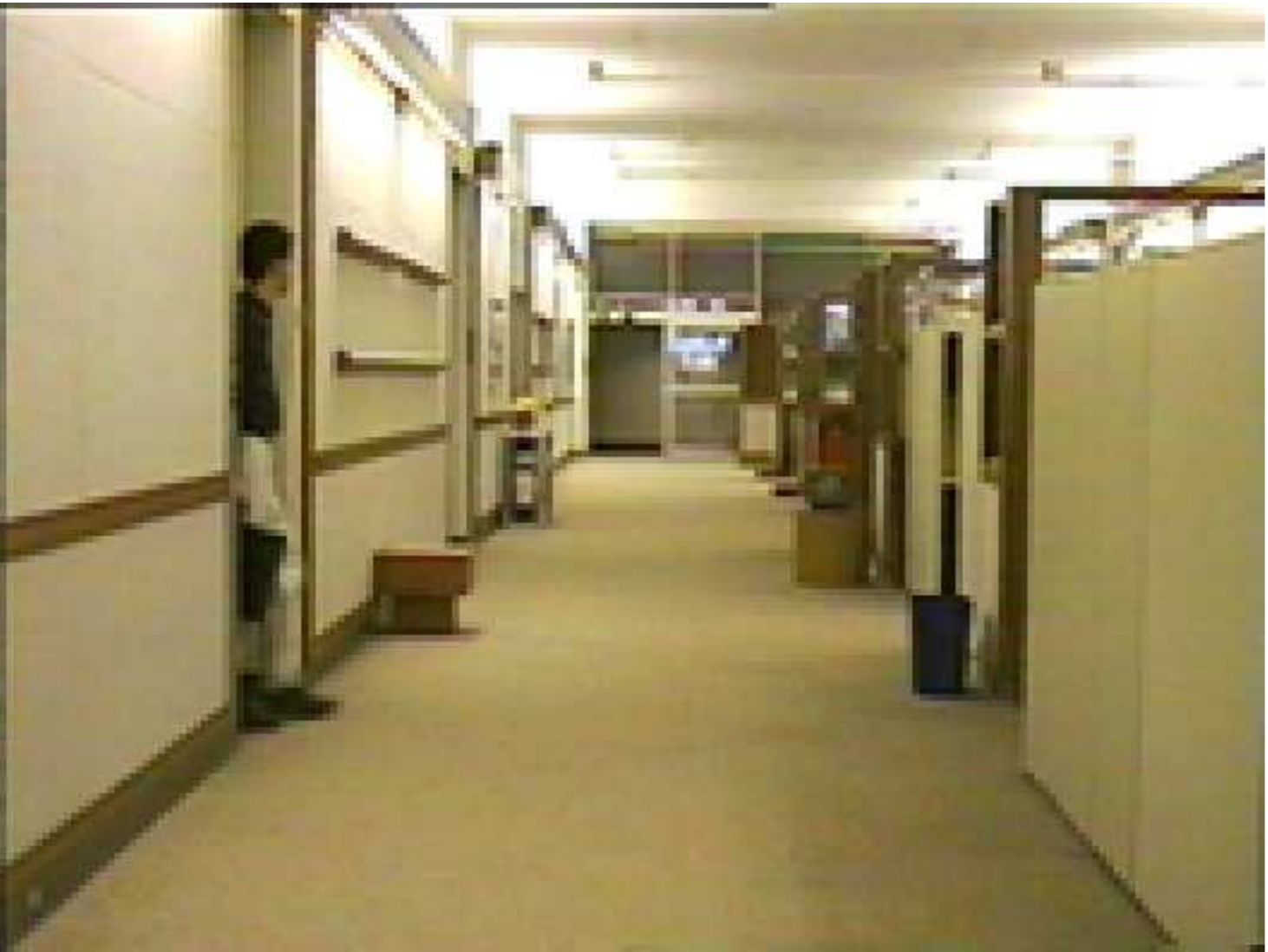}&
\includegraphics[width=0.096\textwidth]{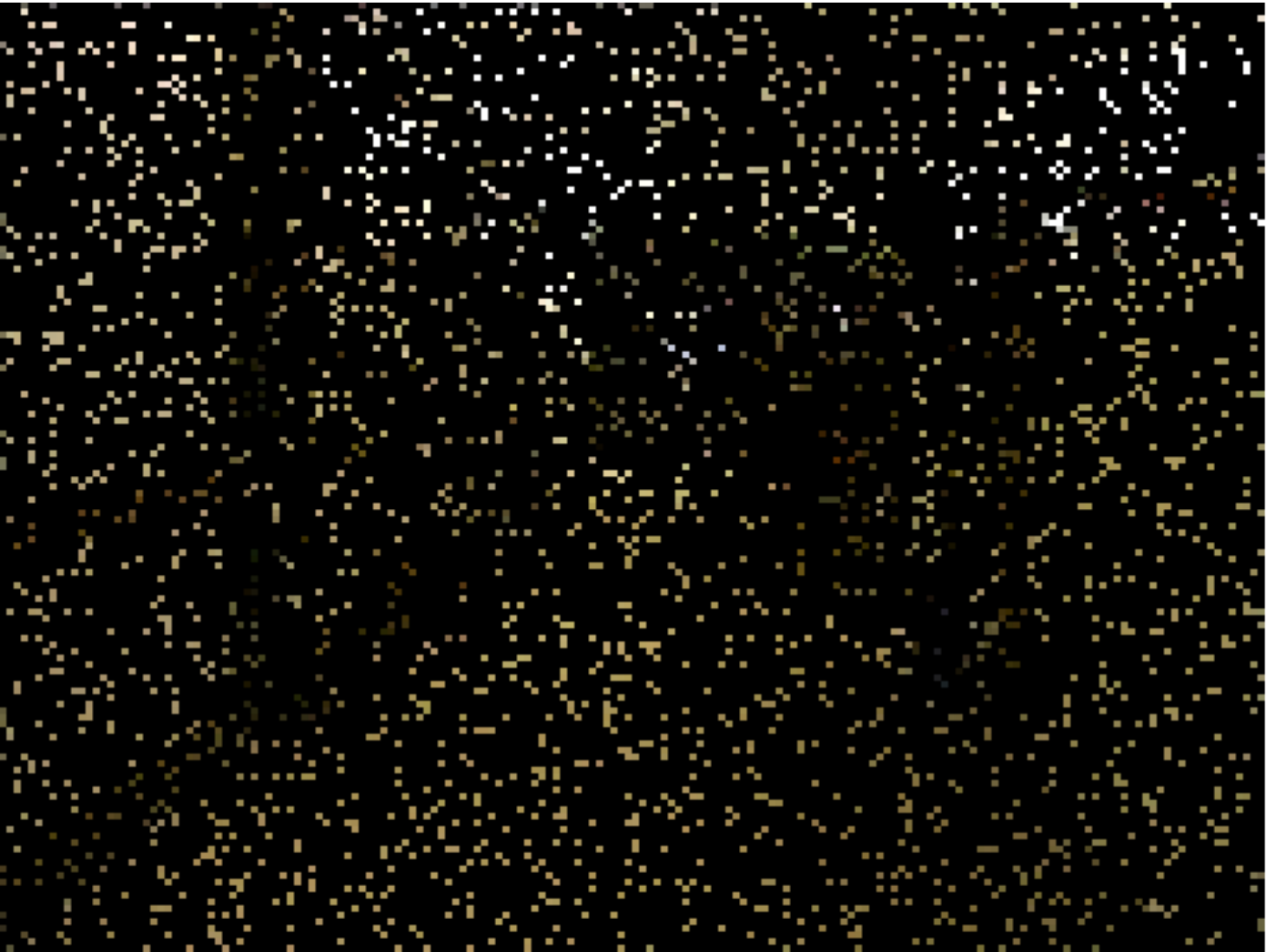}&
\includegraphics[width=0.096\textwidth]{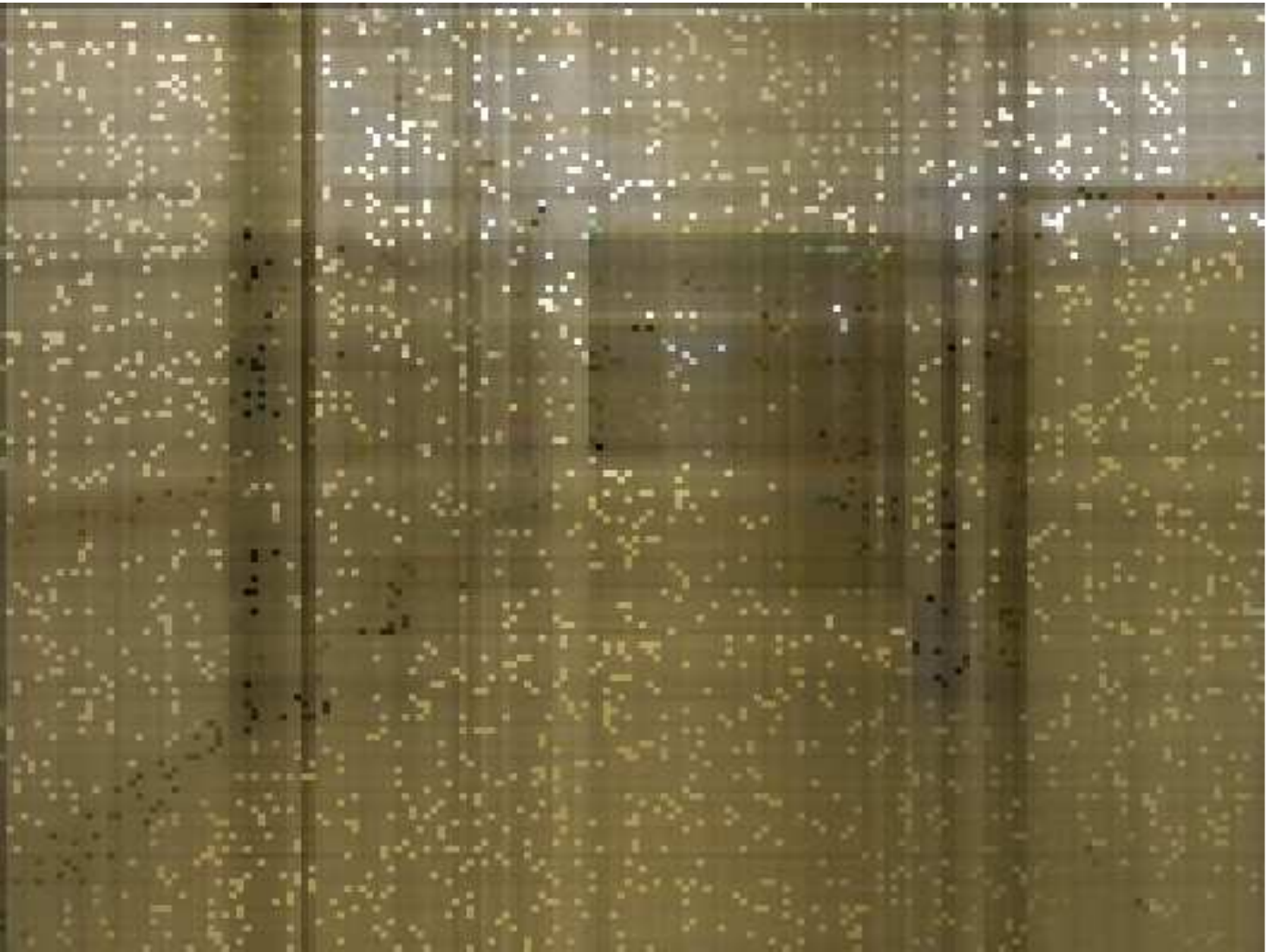}&
\includegraphics[width=0.096\textwidth]{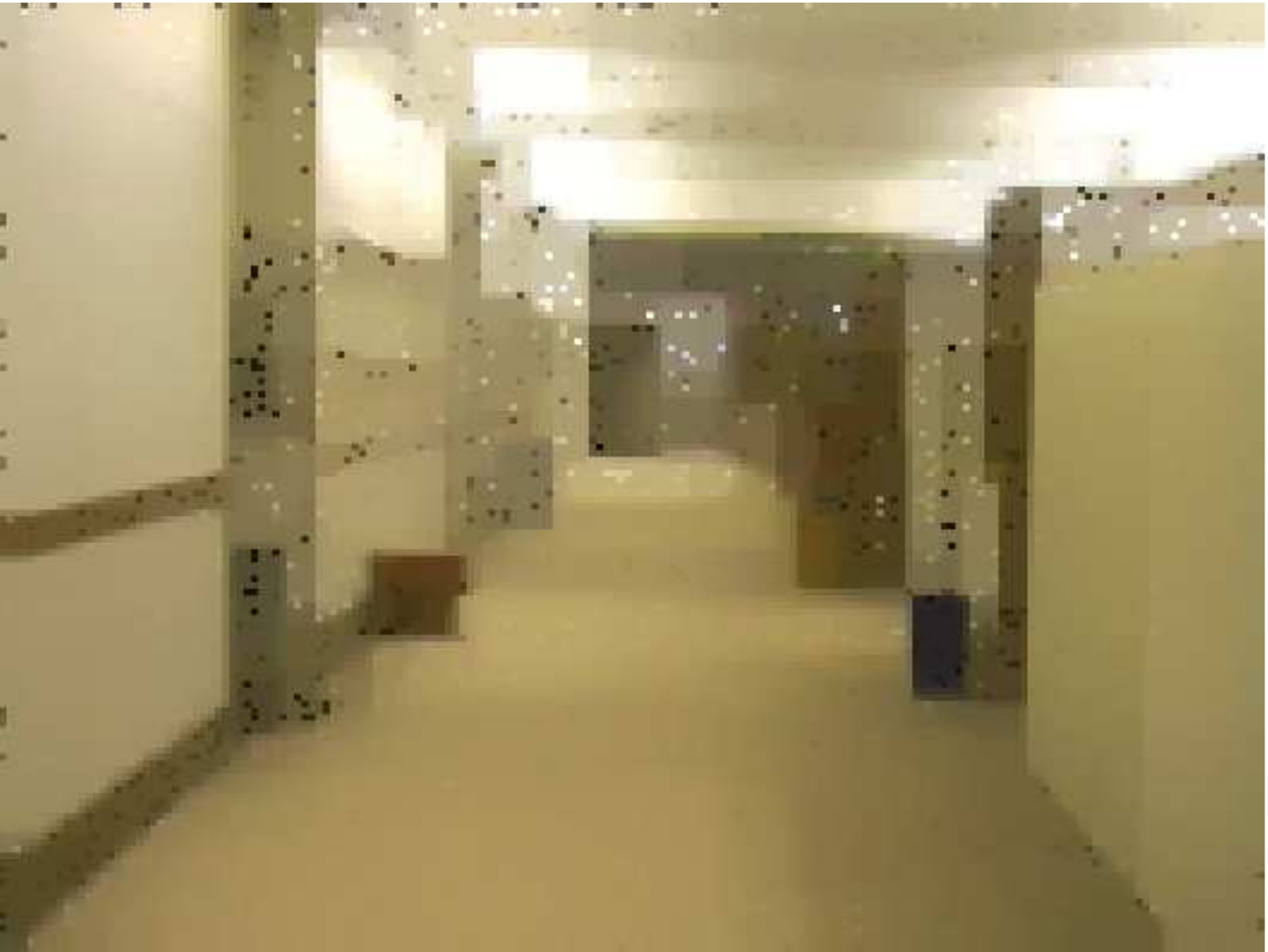}&
\includegraphics[width=0.096\textwidth]{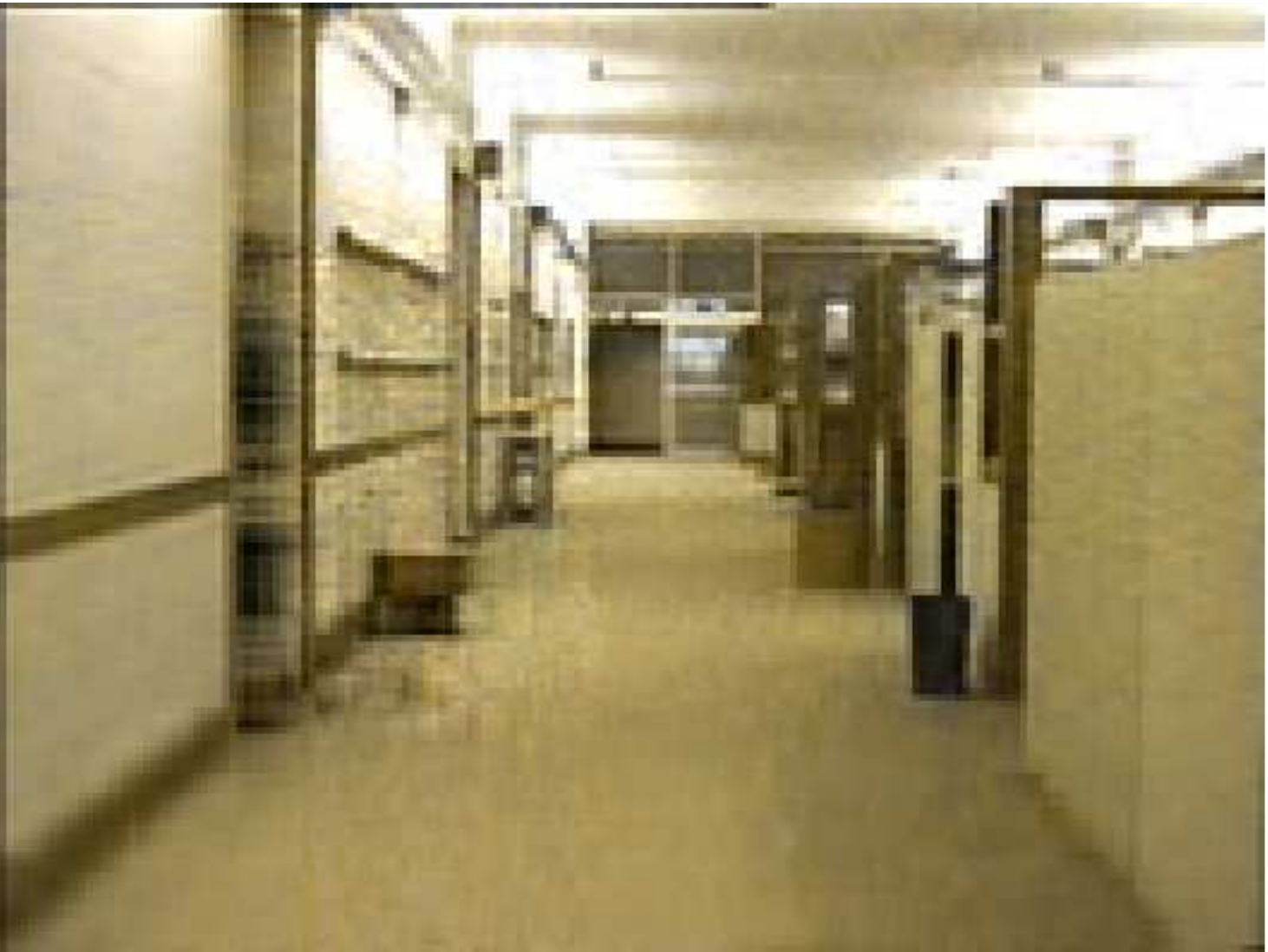}&
\includegraphics[width=0.096\textwidth]{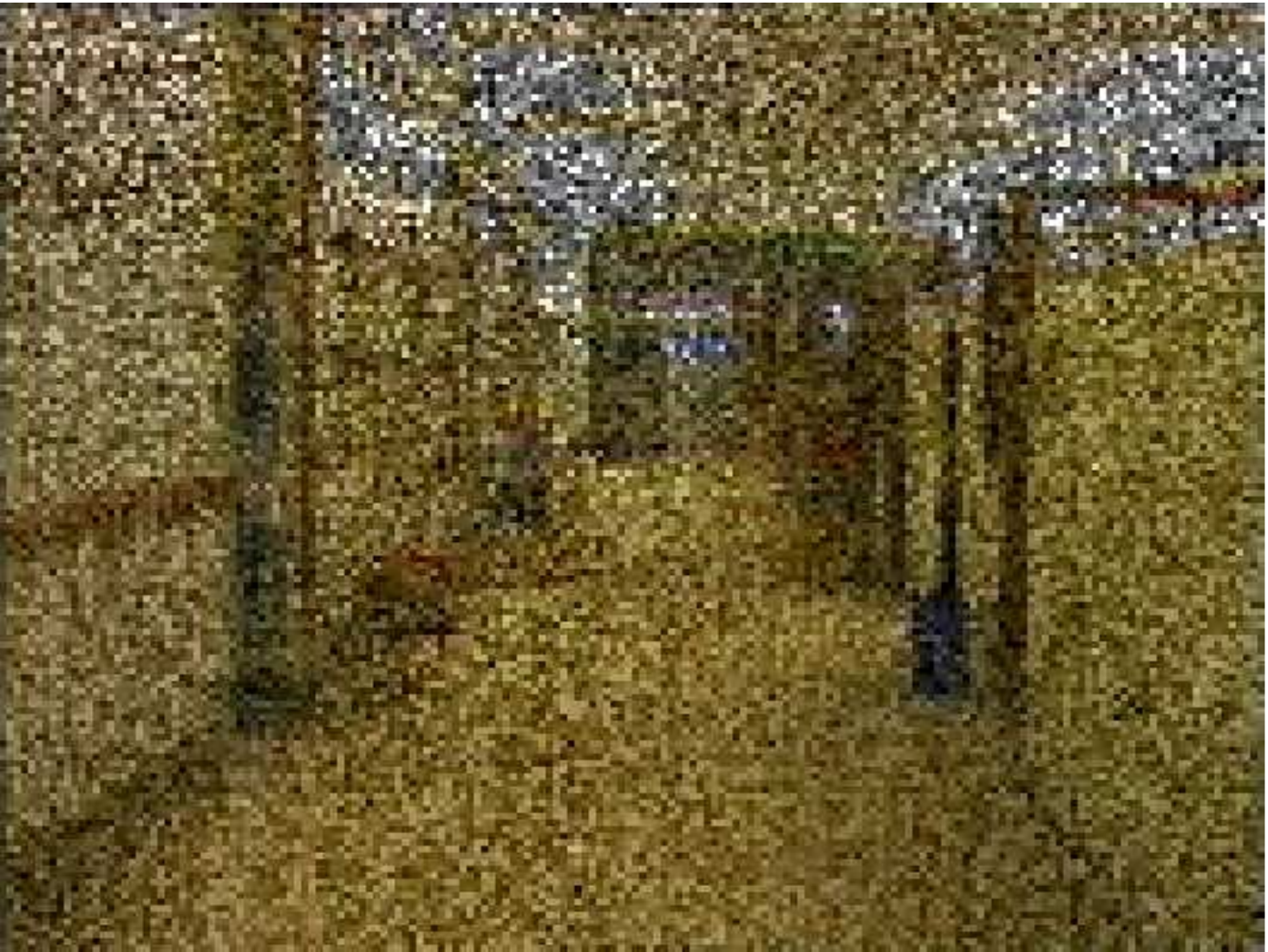}&
\includegraphics[width=0.096\textwidth]{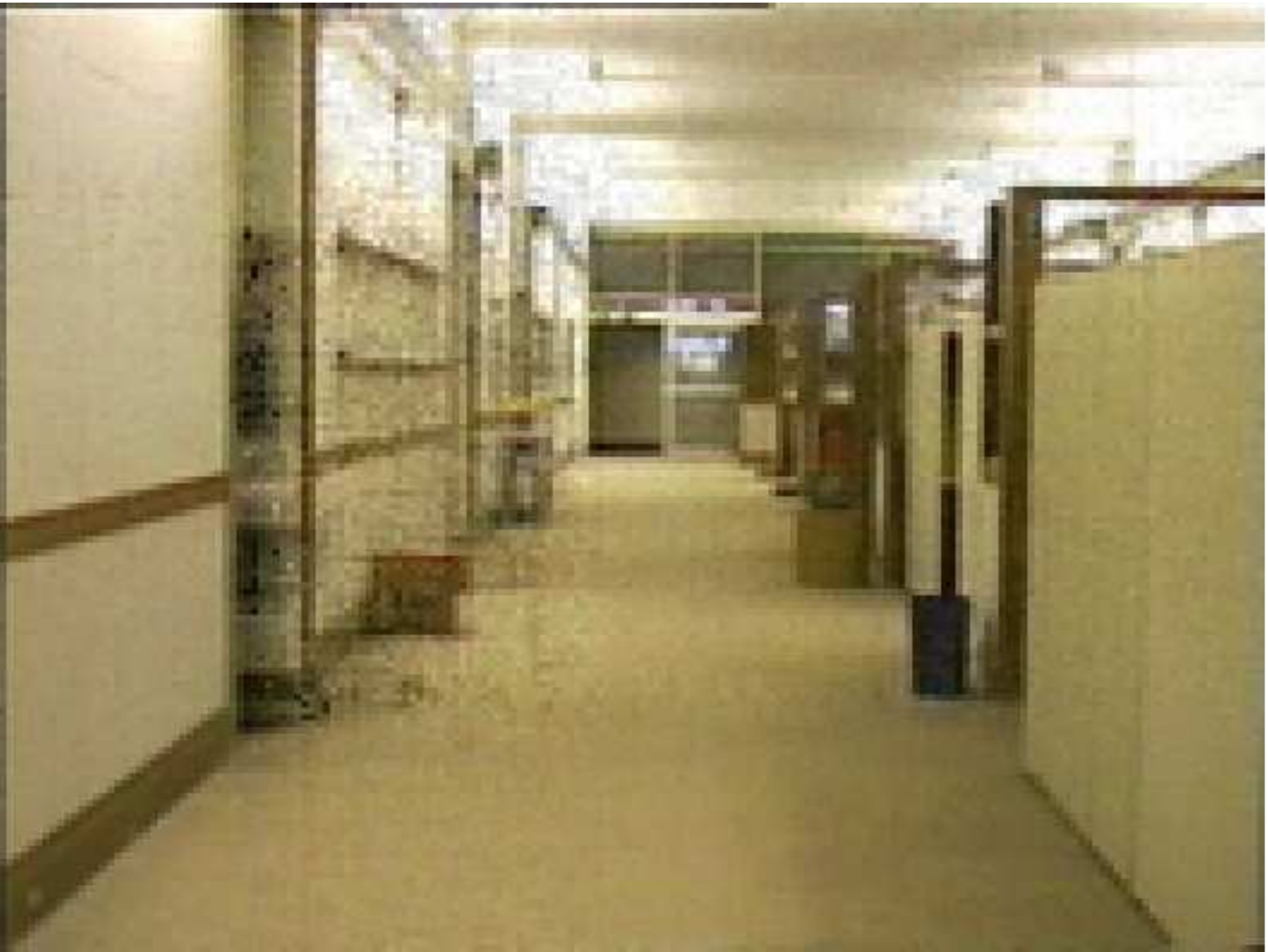}&
\includegraphics[width=0.096\textwidth]{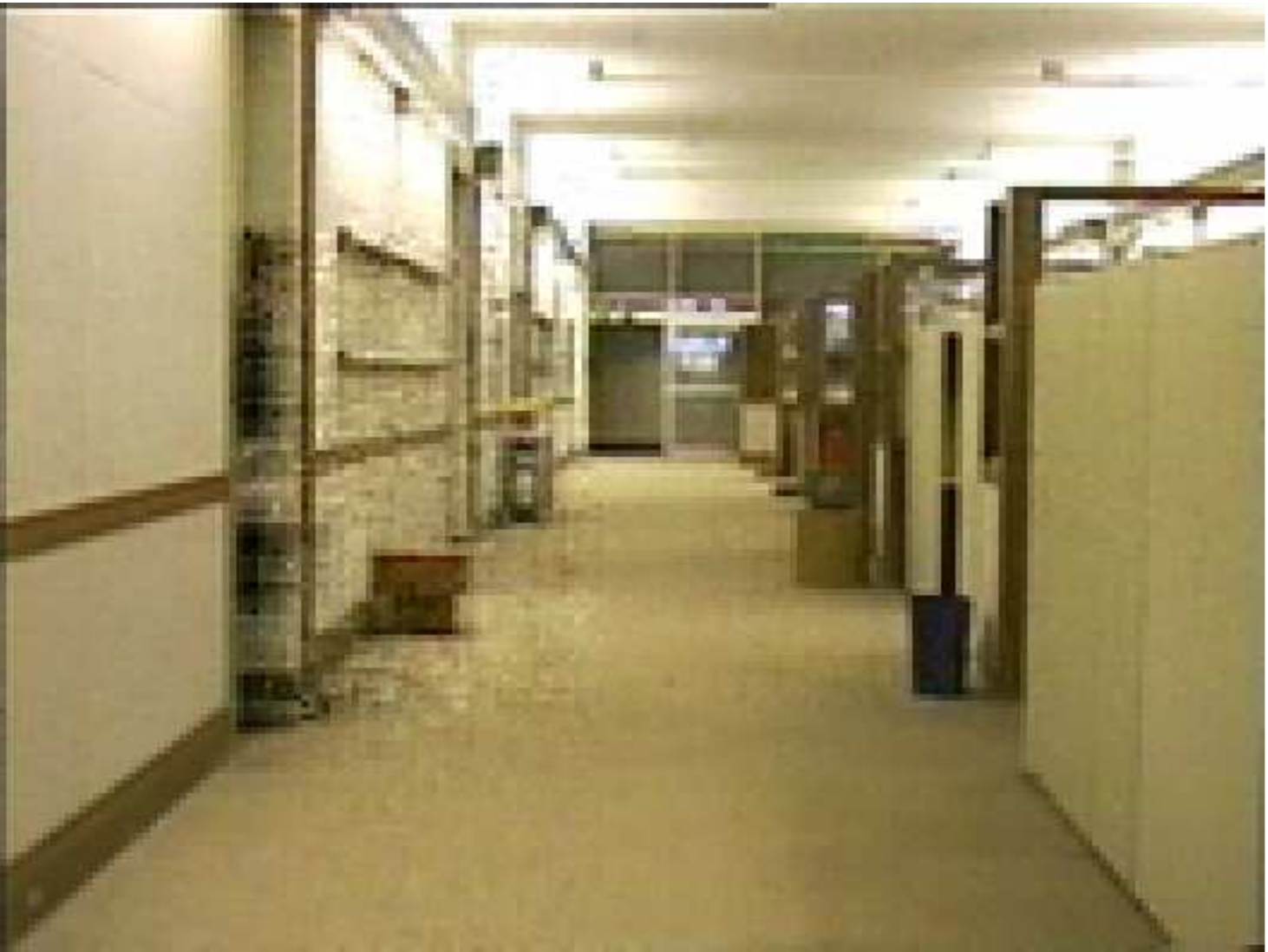}&
\includegraphics[width=0.096\textwidth]{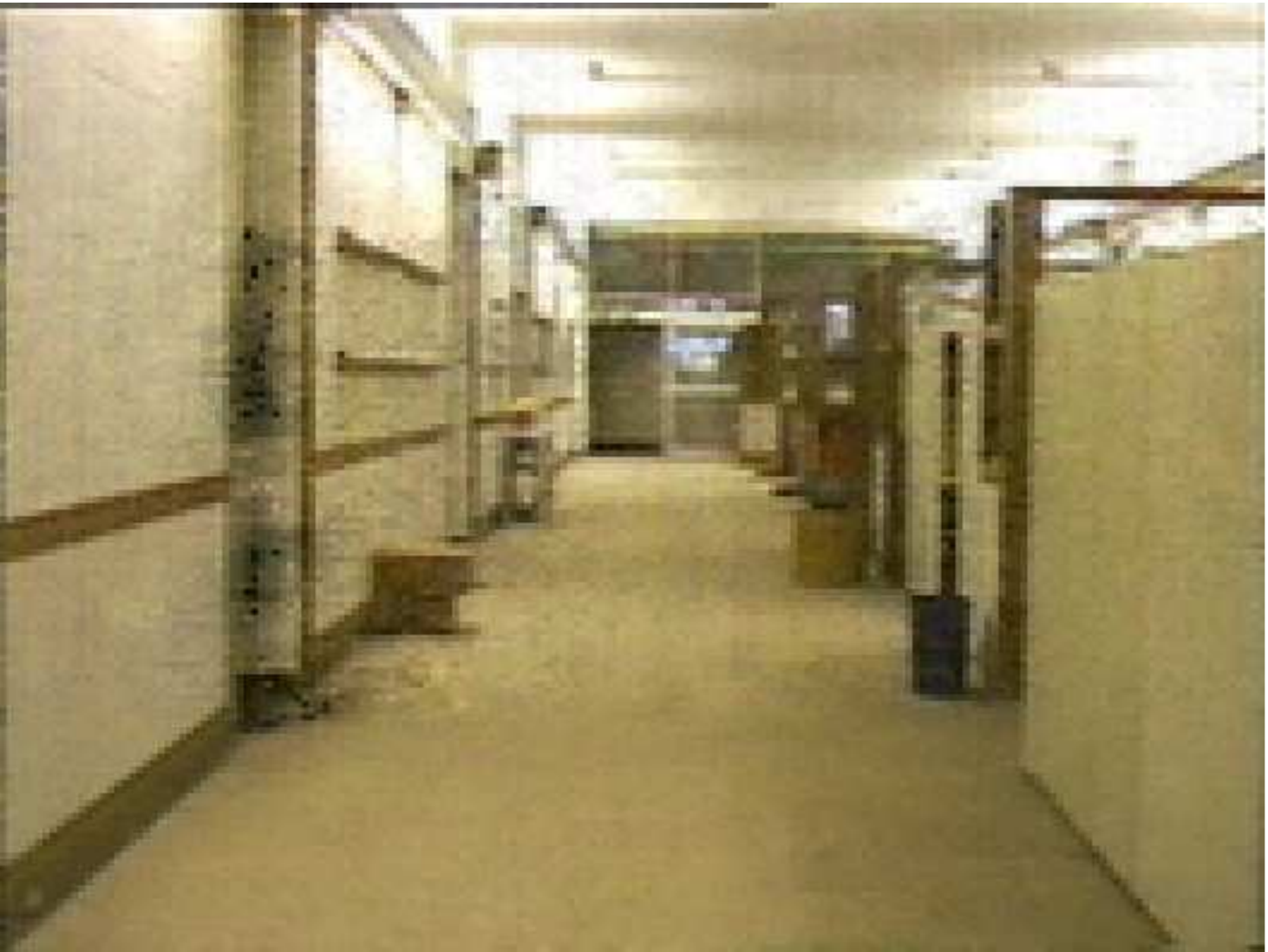}&
\includegraphics[width=0.096\textwidth]{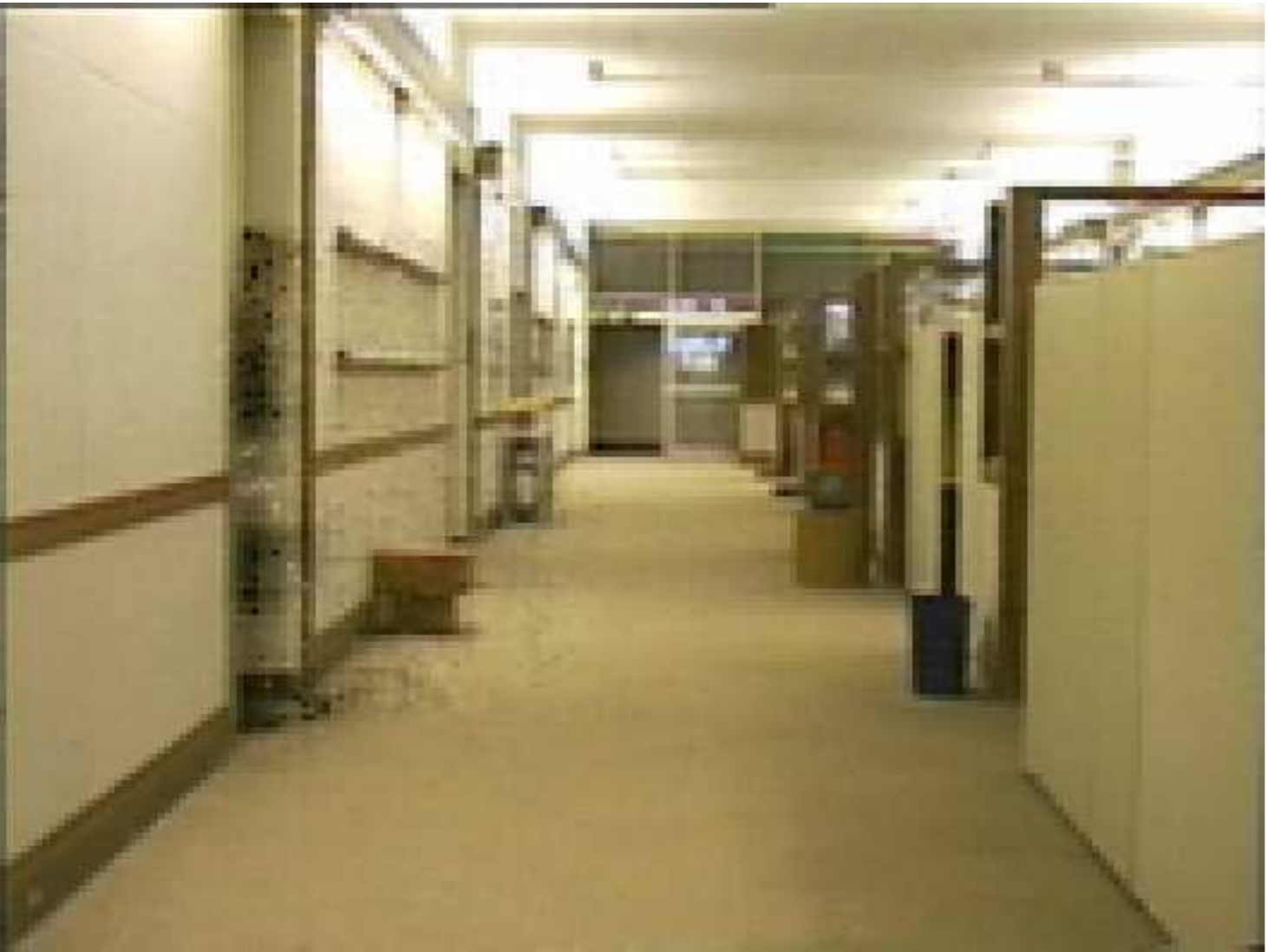}\\
Original& Observed & HaLRTC \cite{Liu2013PAMItensor}& LRTC-TVI \cite{Li2017LowRankTC} &BCPF \cite{ZhangCP} & logDet \cite{JilogDet} & TNN \cite{zhang2017exact}& PSTNN \cite{jiang2017novel2} & t-TNN \cite{HutTNN} & WSTNN
  \end{tabular}
  \caption{The completion results of the CVs \emph{news} and \emph{hall} with $\text{SR}=10\%$. Top row: the image located at the 49-th frame in \emph{news}. Bottom row: the image located at the 20-th frame in \emph{hall}.}
  \label{imagevideofig}
  \end{center}\vspace{-0.6cm}
\end{figure}

\begin{table}[!t]
\scriptsize
\setlength{\tabcolsep}{6pt}
\renewcommand\arraystretch{1.2}
\caption{The PSNR, SSIM, and FSIM values output by eight utilized LRTC methods for CVs.}
\begin{center}
\begin{tabular}{c|c|ccc|ccc|ccc|c}
 \Xhline{1pt}
\multirow{2}{*}{CV}  &SR        &\multicolumn{3}{c|}{5\%}  &\multicolumn{3}{c|}{10\%}  & \multicolumn{3}{c|}{20\%}  & \multirow{2}{*}{Time(s)}\\
\cline{2-11}
                      &Method             &PSNR         & SSIM        &FSIM          &PSNR       & SSIM        &FSIM        &PSNR        & SSIM      &FSIM  \\

                      \hline

\multirow{8}{*}{\emph{news}}

                      &HaLRTC            & 12.59       & 0.413       & 0.649         & 17.67      & 0.596     & 0.767       & 23.92      & 0.816      & 0.886     & \bf{42.53}\\

                      &LRTC-TVI          & 18.31       & 0.640       & 0.731	     & 20.16      & 0.728     & 0.802       & 23.51      & 0.858      & 0.901     & 768.8\\

                      &BCPF              & 25.49       & 0.779       & 0.881 	     & 28.05      & 0.857 	  & 0.919       & 29.87      & 0.897      & 0.939     & 961.3\\

                      &logDet            & 13.69       & 0.288       & 0.836         & 18.03      & 0.534     & 0.782       & 33.11      & 0.944      & 0.969     & 92.16\\

                      &TNN               & 21.23       & 0.659       & 0.832         & 29.12      & 0.893     & 0.940       & 32.75      & 0.943      & 0.968     & 97.32\\

                      &PSTNN             & 23.03       & 0.624       & 0.884         & 29.69      & 0.893     & 0.942       & 33.37      & 0.947      & 0.970     & 98.38\\

                      &t-TNN             & 20.65       & 0.605       & 0.804         & 26.92      & 0.844     & 0.919       & 31.91      & 0.934      & 0.965     & 91.36\\

                      &WSTNN            &\bf{26.92}  &\bf{0.892}    &\bf{0.929}   &\bf{30.67}  &\bf{0.947}   &\bf{0.964}  &\bf{34.61}   &\bf{0.976}  &\bf{0.983}  & 324.2\\

                      \hline

 \multirow{8}{*}{\emph{hall}}

                      &HaLRTC            & 9.179     & 0.180        & 0.532       & 14.21       & 0.348       & 0.640       & 23.82      & 0.784      & 0.858  &\bf{40.68}\\

                      &LRTC-TVI          & 19.02     & 0.606        & 0.679	      & 21.83       & 0.755       & 0.797       & 25.93      & 0.875      & 0.905  & 712.6\\

                      &BCPF              & 25.69     & 0.783        & 0.868 	  & 28.32       & 0.847 	  & 0.905       & 30.72      & 0.884      & 0.931  & 906.5 \\

                      &logDet            & 8.192     & 0.073        & 0.503       & 10.80       & 0.162       & 0.553       & 17.30      & 0.498      & 0.750   & 93.94\\

                      &TNN               & 15.29     & 0.375        & 0.692       & 26.95       & 0.840       & 0.909       & 31.92      & 0.916      & 0.957   & 92.46\\

                      &PSTNN             & 25.31     & 0.765        & 0.874       & 29.39       & 0.876       & 0.935       & 32.42      & 0.920      & 0.959   & 92.27\\

                      &t-TNN             & 17.87     & 0.442        & 0.718       & 27.86       & 0.844       & 0.917       & 32.39      & 0.918      & 0.957   & 84.98\\

                      &WSTNN            &\bf{26.99}  &\bf{0.875}    &\bf{0.920}   &\bf{31.13}  &\bf{0.925}   &\bf{0.956}  &\bf{34.36}   &\bf{0.953}  &\bf{0.973}& 347.6\\

 \Xhline{1pt}
\end{tabular}
\end{center}\vspace{-0.5cm}
\label{VideoTC}
\end{table}

\textbf{MRI completion.} We test MRI\footnote{\url{http://brainweb.bic.mni.mcgill.ca/brainweb/selection\_normal.html}.} data set of size $181 \times 217 \times 181$. Table \ref{MRITC} lists the values of PSNR, SSIM, and FSIM of the testing MRI recovered by different LRTC methods. As observed, the proposed method significantly outperforms the compared methods in terms of all evaluation indices. In Fig. \ref{imageMRIfig}, we show three slices obtained by different directions. It can be observed that no matter which direction they are in, the proposed method is evidently superior to the compared ones, both in recovery of abundant shape structure and texture information.

\begin{table}[!t]
\scriptsize
\setlength{\tabcolsep}{8pt}
\renewcommand\arraystretch{1.2}
\caption{The  PSNR, SSIM, and FSIM values output by eight utilized LRTC methods for HSV.}
\begin{center}
\begin{tabular}{c|ccc|ccc|ccc|c}

 \Xhline{1pt}

                       SR        &\multicolumn{3}{c|}{5\%}  &\multicolumn{3}{c|}{10\%}  & \multicolumn{3}{c|}{20\%}  &\multirow{2}{*}{Time(s)}\\

                      \cline{1-10}

                      Method             &PSNR         & SSIM        &FSIM          &PSNR       & SSIM        &FSIM        &PSNR        & SSIM      &FSIM  \\

                      \hline

                      HaLRTC            & 9.008        & 0.115       & 0.519       & 10.46     & 0.194       & 0.565       & 13.41     & 0.338     & 0.652  & \bf{162.77}\\

                      LRTC-TVI          & 22.09        & 0.686       & 0.791	   & 27.08     & 0.835       & 0.891       & 32.19     & 0.931     & 0.959  & 5121.5\\

                      BCPF              & 27.75        & 0.855       & 0.907 	   & 30.23     & 0.902       & 0.934       & 31.69     & 0.917     & 0.945  & 5840.6\\

                      logDet            & 31.01        & 0.912       & 0.948       & 38.94     & 0.975       & 0.984       & 44.52     & 0.991     & 0.995  & 446.61\\

                      TNN               & 33.68        & 0.946       & 0.968       & 38.02     & 0.974       & 0.984       & 42.94     & 0.989     & 0.993  & 487.95\\

                      PSTNN             & 32.93        & 0.934       & 0.960       & 38.53     & 0.975       & 0.985       & 43.41     & 0.989     & 0.994  & 423.32\\

                      t-TNN             & 29.43        & 0.894       & 0.931       & 34.37     & 0.957       & 0.971       & 40.11     & 0.986     & 0.990  & 391.87\\

                      WSTNN            &\bf{37.61}  &\bf{0.979}    &\bf{0.986}   &\bf{43.67}  &\bf{0.994}   &\bf{0.996}    & \bf{49.11}   & \bf{0.997}  & \bf{0.998} &1228.3\\

  \Xhline{1pt}
\end{tabular}
\end{center}\vspace{-0.1cm}
\label{HSVTC}
\end{table}

\begin{figure}[!t]
\tiny
\setlength{\tabcolsep}{0.9pt}
\begin{center}
\begin{tabular}{cccccccccc}
\includegraphics[width=0.096\textwidth]{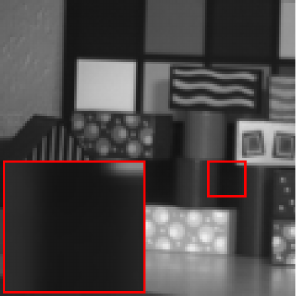}&
\includegraphics[width=0.096\textwidth]{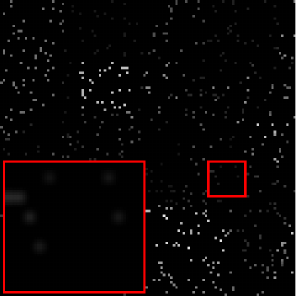}&
\includegraphics[width=0.096\textwidth]{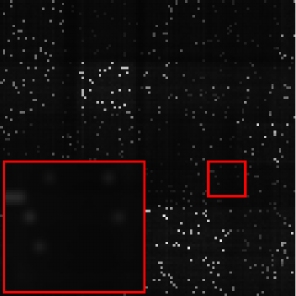}&
\includegraphics[width=0.096\textwidth]{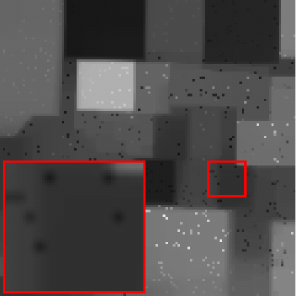}&
\includegraphics[width=0.096\textwidth]{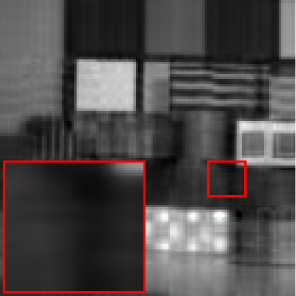}&
\includegraphics[width=0.096\textwidth]{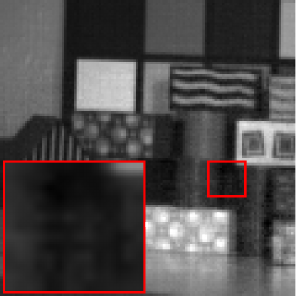}&
\includegraphics[width=0.096\textwidth]{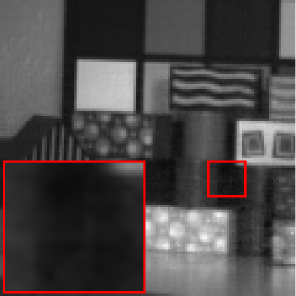}&
\includegraphics[width=0.096\textwidth]{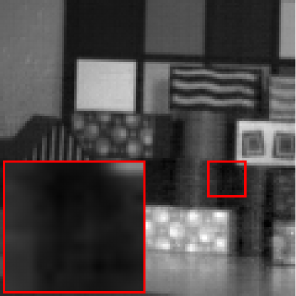}&
\includegraphics[width=0.096\textwidth]{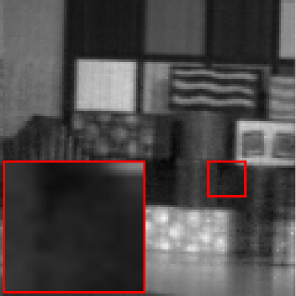}&
\includegraphics[width=0.096\textwidth]{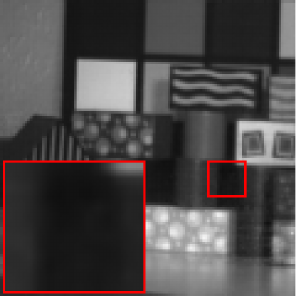}\\
\includegraphics[width=0.096\textwidth]{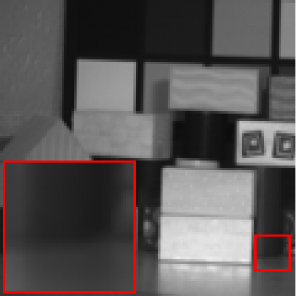}&
\includegraphics[width=0.096\textwidth]{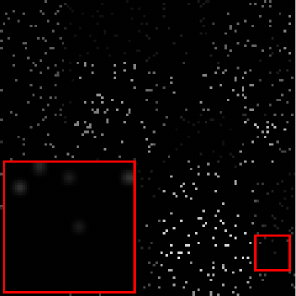}&
\includegraphics[width=0.096\textwidth]{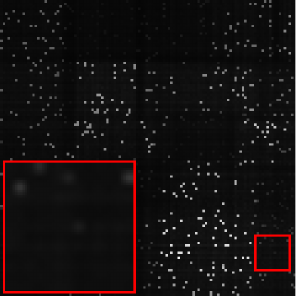}&
\includegraphics[width=0.096\textwidth]{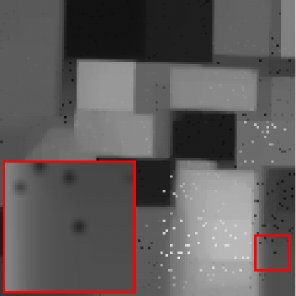}&
\includegraphics[width=0.096\textwidth]{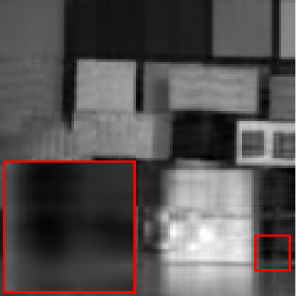}&
\includegraphics[width=0.096\textwidth]{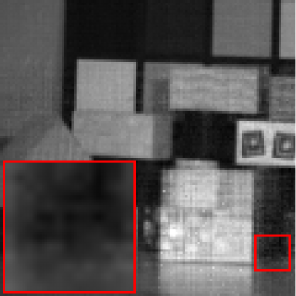}&
\includegraphics[width=0.096\textwidth]{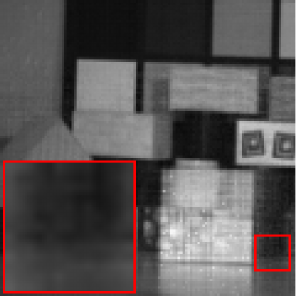}&
\includegraphics[width=0.096\textwidth]{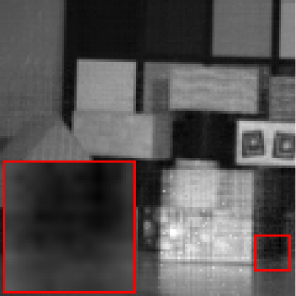}&
\includegraphics[width=0.096\textwidth]{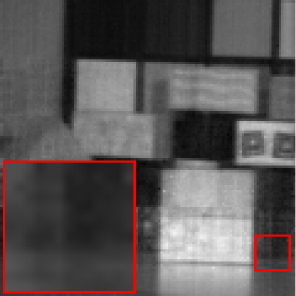}&
\includegraphics[width=0.096\textwidth]{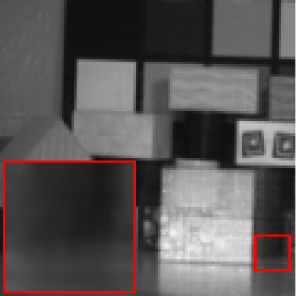}\\
Original& Observed & HaLRTC \cite{Liu2013PAMItensor}& LRTC-TVI \cite{Li2017LowRankTC} &BCPF \cite{ZhangCP} & logDet \cite{JilogDet} & TNN \cite{zhang2017exact}& PSTNN \cite{jiang2017novel2} & t-TNN \cite{HutTNN} & WSTNN
  \end{tabular}
  \caption{The completion results of the HSV with $\text{SR}=5\%$. Top row: the image located at the 15-th band and the 7-th frame. Bottom row: the image located at the 25-th band and the 30-th frame.}
  \label{HSVfig}
  \end{center}\vspace{-0.6cm}
\end{figure}

\textbf{CV completion.} We test CVs \emph{news}\footnote{\url{http://trace.eas.asu.edu/yuv/}.\label{CVweb}} and \emph{hall}\textsuperscript{\ref{CVweb}} of size $144 \times 176 \times 3\times50$. For each frames, the missing elements of each channel have the same location. Table \ref{VideoTC} lists the values of PSNR, SSIM, and FSIM of the testing CVs recovered by different LRTC methods. As observed, the proposed method has an overall better performance than the compared ones with respect to all evaluation indices. In Fig. \ref{imagevideofig}, we show one frame in these two testing CVs recovered by eight compared methods with $\text{SR}=10\%$. We observe that the results obtained by the proposed method is superior to those obtained by the compared ones.

\textbf{HSV completion.} We test HSV\footnote{\url{http://openremotesensing.net/knowledgebase/hyperspectral-video/}.} of size $120 \times 120 \times 33\times31$. Specifically, this HSV has $31$ frames and each frame has $33$ bands from $400$nm to $720$nm wavelength with a $10$nm step \cite{HSVdata}. Table \ref{HSVTC} lists the values of PSNR, SSIM, and FSIM of the testing HSV recovered by different LRTC methods. As observed, the proposed method consistently achieves the highest values in terms of all evaluation indexes, e.g., no matter what the SR is set as, the proposed method achieves around 4 dB gain in PSNR beyond the second best method. In Fig. \ref{HSVfig}, we show two images located at different frames and different bands in the HSV recovered by eight compared methods with $\text{SR}=5\%$. We observe that the proposed method is evidently superior to the compared ones, especially in recovery of texture information.

\begin{table}[!t]
\scriptsize
\setlength{\tabcolsep}{10pt}
\renewcommand\arraystretch{1.2}
\caption{Parameters setting in the proposed WSTNN-based TRPCA method on different data.}
\begin{center}
\begin{tabular}{c|c|c|c}
  \Xhline{1pt}
 Test                  &Tensor        &$\alpha$             &$\tau$  \\

\hline

\multirow{2}{*}{\tabincell{c}{synthetic data denoising}}

                      &three-way tensor         &(1,1,1)/3      & (10,10,10)        \\

                      &four-way tensor          &(1,1,1,1,1,1)/6    & (50,50,50,50,50,50)     \\

                      \hline

HSI denoising        &three-way tensor      &(0.001,1,1)/2.001      &(100,100,100)          \\

 \Xhline{1pt}
\end{tabular}
\end{center}\vspace{-0.2cm}
\label{parasetTRPCA}
\end{table}

\subsection{Tensor robust principal component analysis}

In this section, we evaluate the performance of the proposed WSTNN-based TRPCA method by synthetic data and HSI denoising. The compared TRPCA methods include SNN \cite{Goldfarb2014} and TNN \cite{Lu2016}.

\begin{figure}[!t]
\setlength{\tabcolsep}{5pt}
\begin{center}
\begin{tabular}{cccc}
\includegraphics[width=0.23\textwidth]{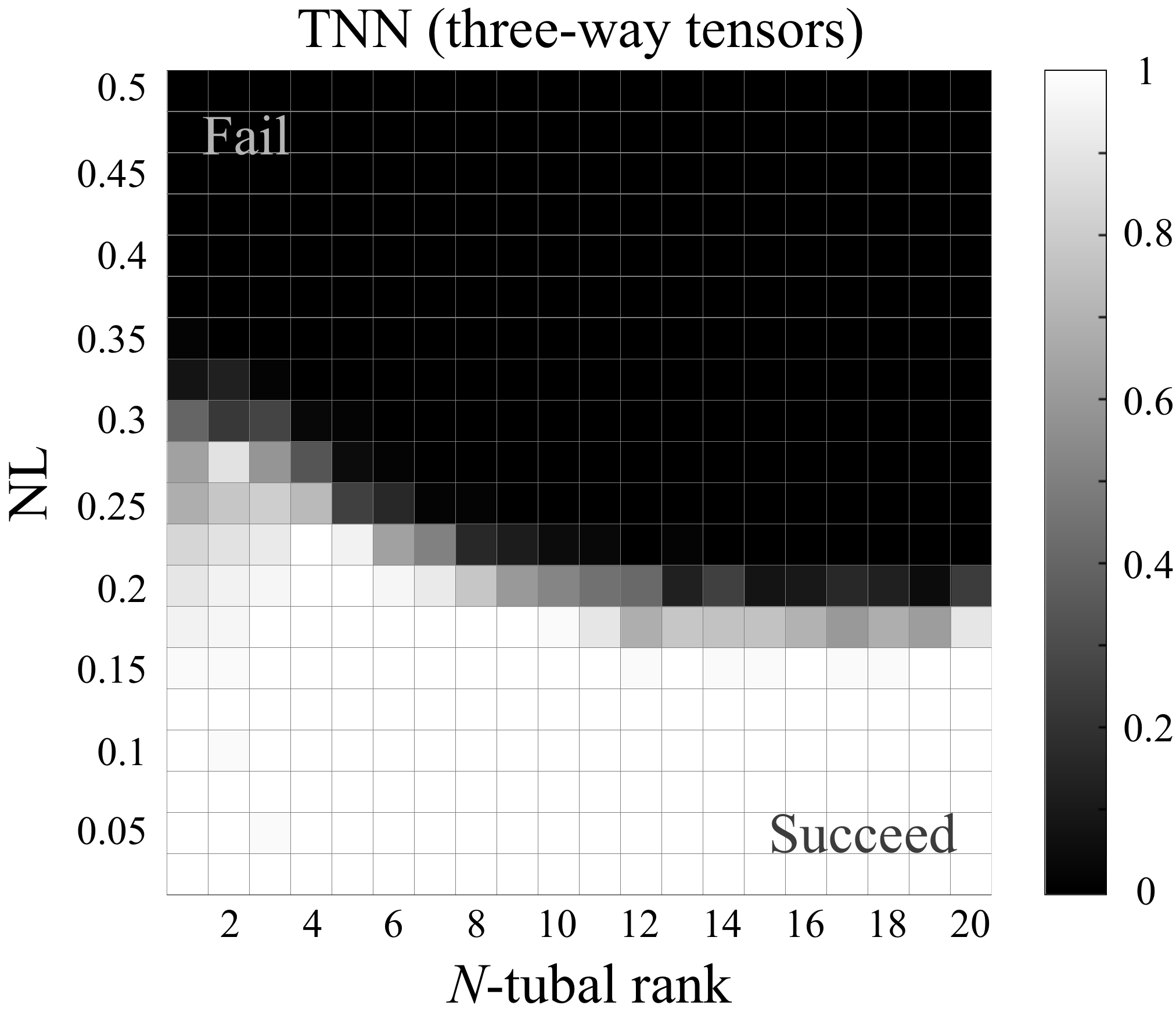}&
\includegraphics[width=0.23\textwidth]{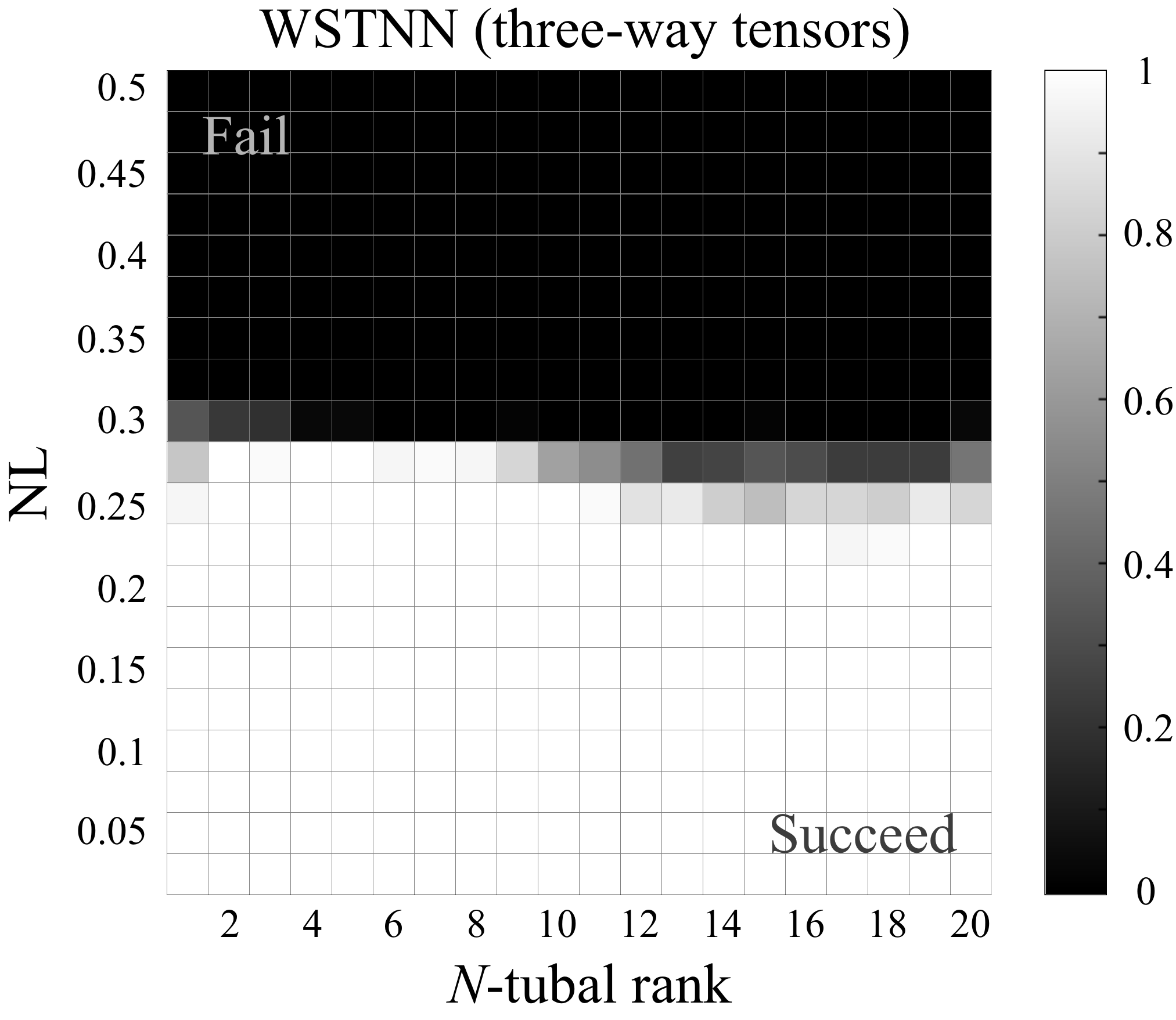}&
\includegraphics[width=0.23\textwidth]{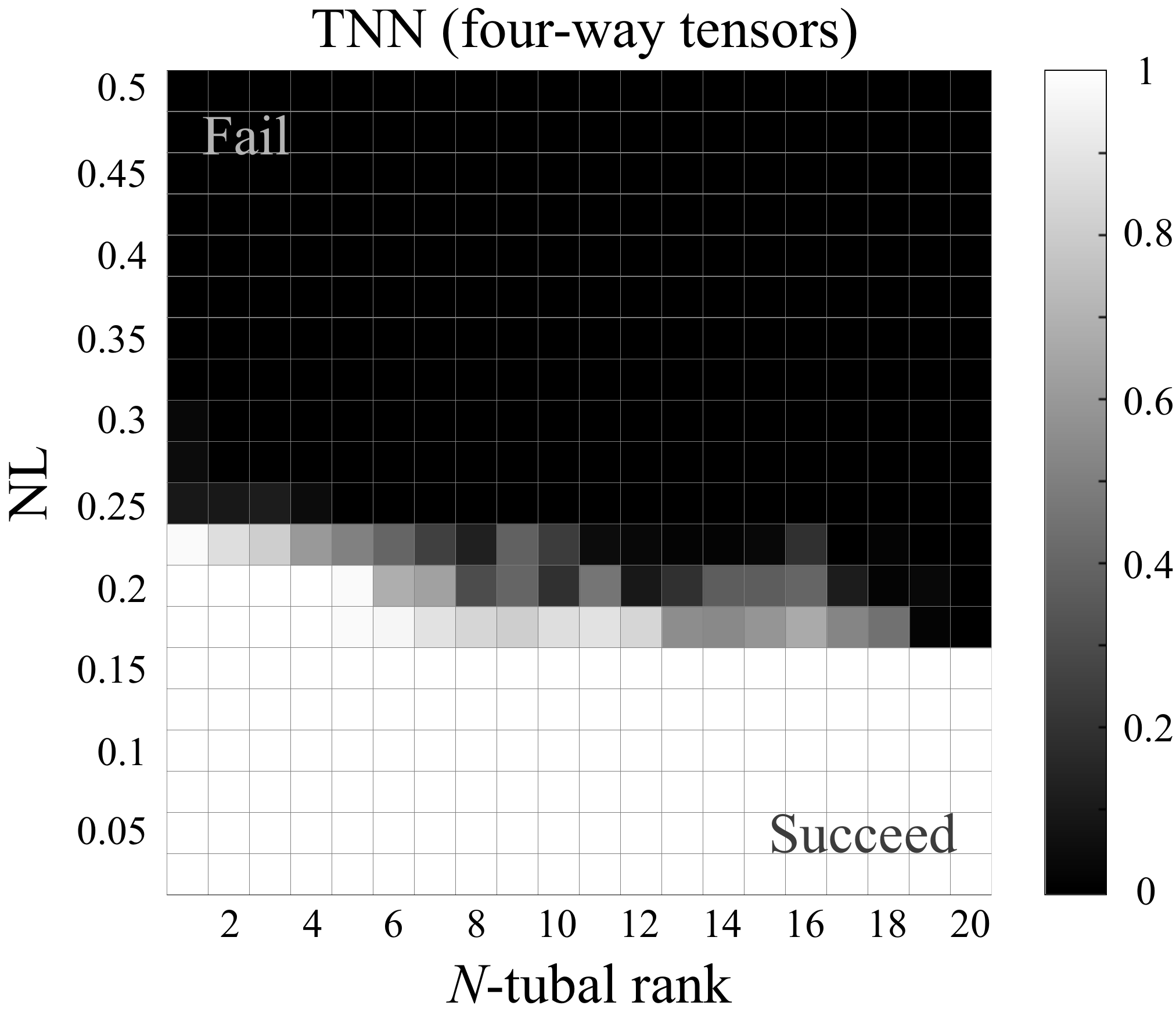}&
\includegraphics[width=0.23\textwidth]{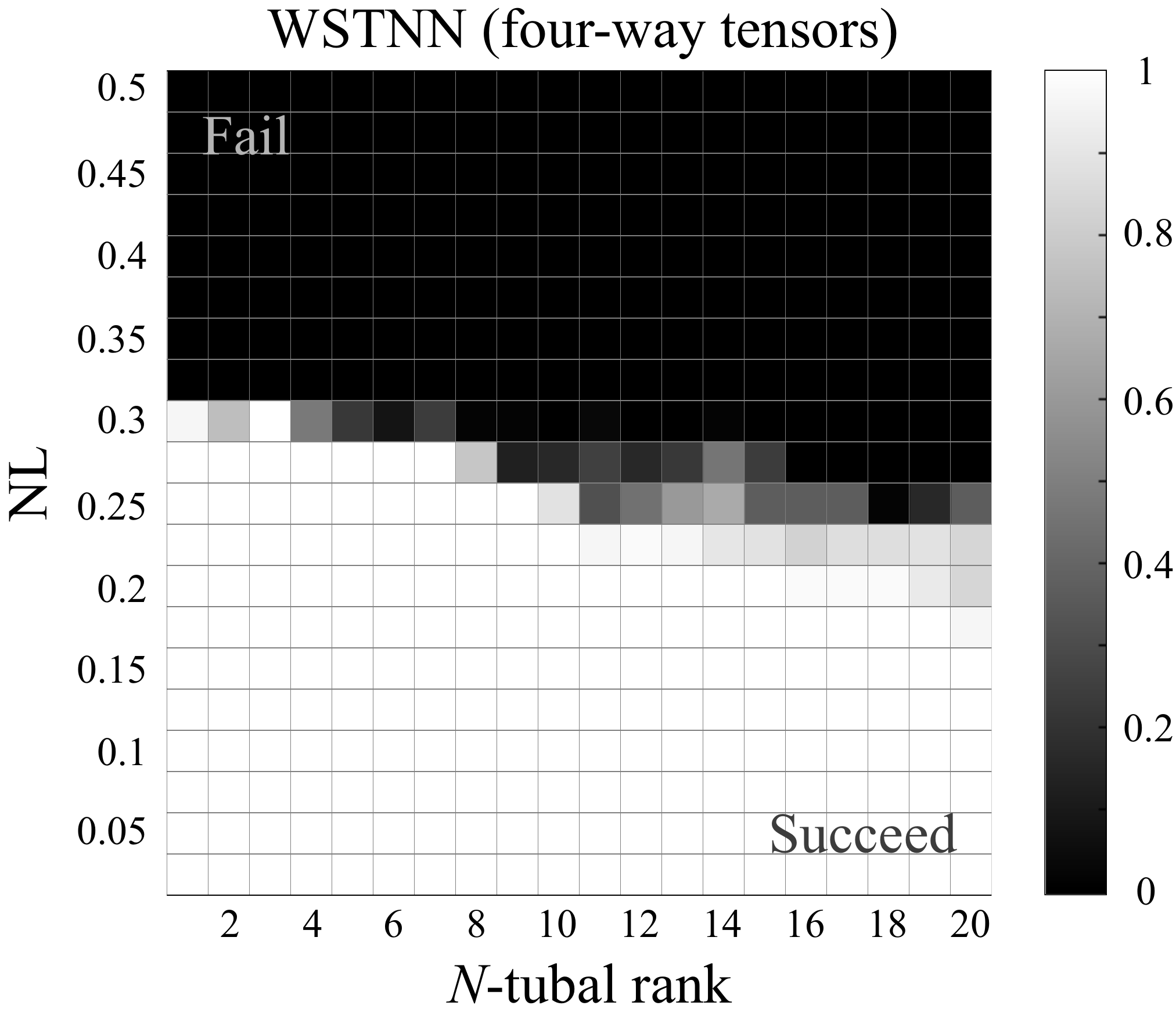}
\end{tabular}
\caption{The success rates for synthetic data with varying $N$-tubal rank and varying NLs. The left two are the results of TNN-based TRPCA method \cite{Lu2016} and the proposed WSTNN-based TRPCA method on three-way tensors. The right two are the results of TNN-based TRPCA method \cite{Lu2016} and the proposed WSTNN-based TRPCA method on four-way tensors. The gray magnitude represents the success rates.}\vspace{-0.3cm}
\label{sypca}
\end{center}\vspace{-0.3cm}
\end{figure}

\begin{table}[!t]
\scriptsize
\setlength{\tabcolsep}{8pt}
\renewcommand\arraystretch{1.2}
\caption{The PSNR, SSIM, and FSIM values output by three utilized TRPCA methods for HSIs.}
\begin{center}
\begin{tabular}{c|c|ccc|ccc|c}
  \Xhline{1pt}
\multirow{2}{*}{HSI}  &NL        &\multicolumn{3}{c|}{0.2}  &\multicolumn{3}{c|}{0.4}                                   &\multirow{2}{*}{Time(s)}\\
\cline{2-8}
                      &Method             &PSNR         & SSIM        &FSIM          &PSNR       & SSIM        &FSIM          &\\

                      \hline

\multirow{3}{*}{\tabincell{c}{\emph{Washington} \\ \emph{DC Mall} \\ $256\times256\times150$}}

                      &SNN               & 31.48          & 0.927      & 0.950       & 28.19      & 0.848      & 0.902     & \bf{79.822}  \\

                      &TNN               & 43.87          & 0.992      & 0.994       & 35.82      & 0.953      & 0.973     &  172.81\\

                      &WSTNN           & \bf{50.49}   & \bf{0.999}   & \bf{0.999}   &\bf{42.29}  &\bf{0.993}   &\bf{0.995}  & 385.39\\

                      \hline

 \multirow{3}{*}{\tabincell{c}{\emph{Pavia} \\ \emph{University} \\ $256\times256\times87$}}

                      &SNN               & 28.14         & 0.877       & 0.899       & 26.16       & 0.787      & 0.834    & \bf{56.238} \\

                      &TNN               & 38.97         & 0.983       & 0.988       & 35.42       & 0.958      & 0.975    & 120.28 \\

                      &WSTNN            &\bf{39.21}    &\bf{0.995}    &\bf{0.997}   &\bf{36.48}  &\bf{0.988}   &\bf{0.993} & 243.89\\

 \Xhline{1pt}
\end{tabular}
\end{center}\vspace{-0.0cm}
\label{HSIPCA}
\end{table}

\begin{figure}[!t]
\scriptsize\setlength{\tabcolsep}{0.9pt}
\begin{center}
\begin{tabular}{ccccc}
\includegraphics[width=0.195\textwidth]{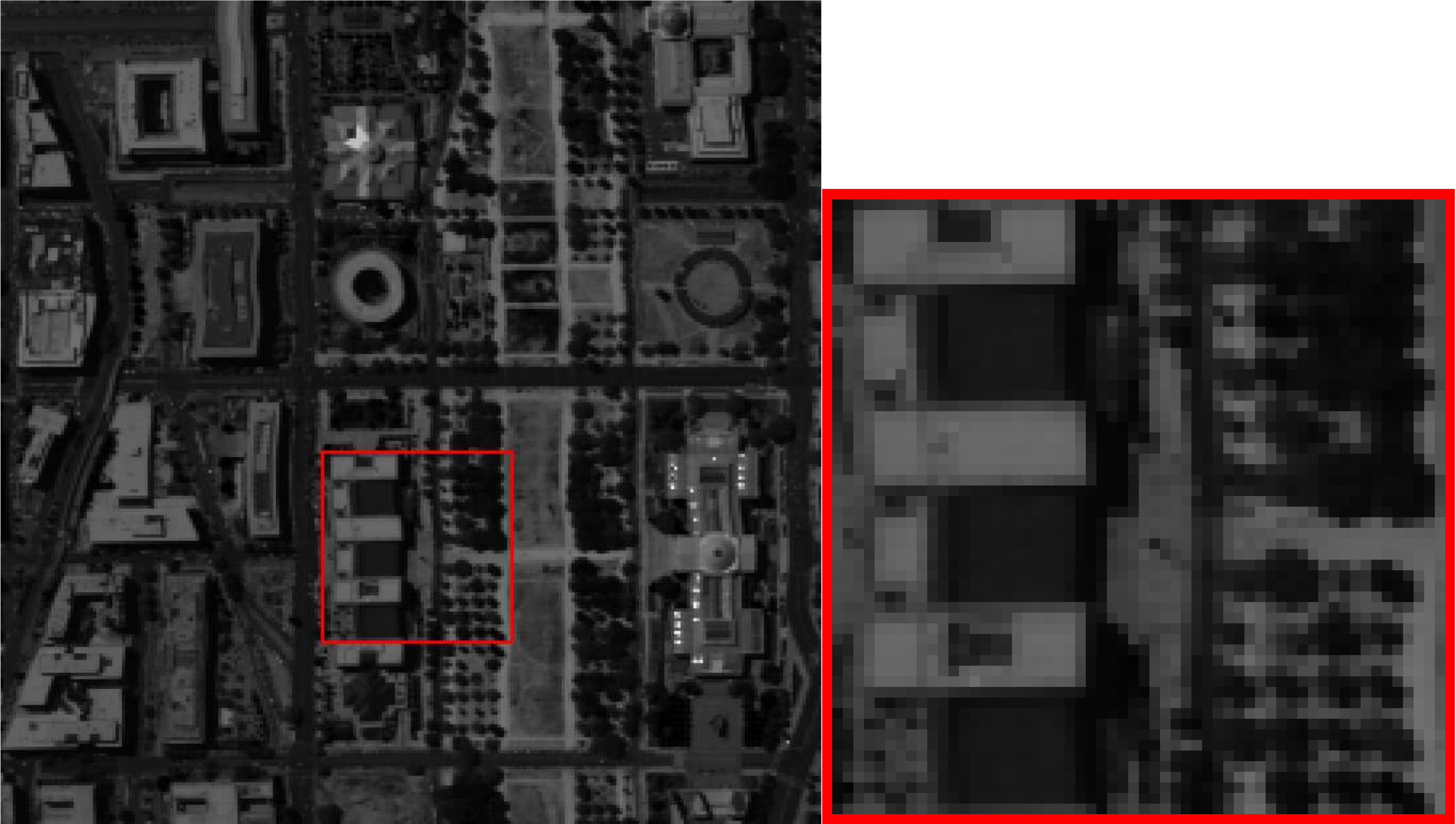}&
\includegraphics[width=0.195\textwidth]{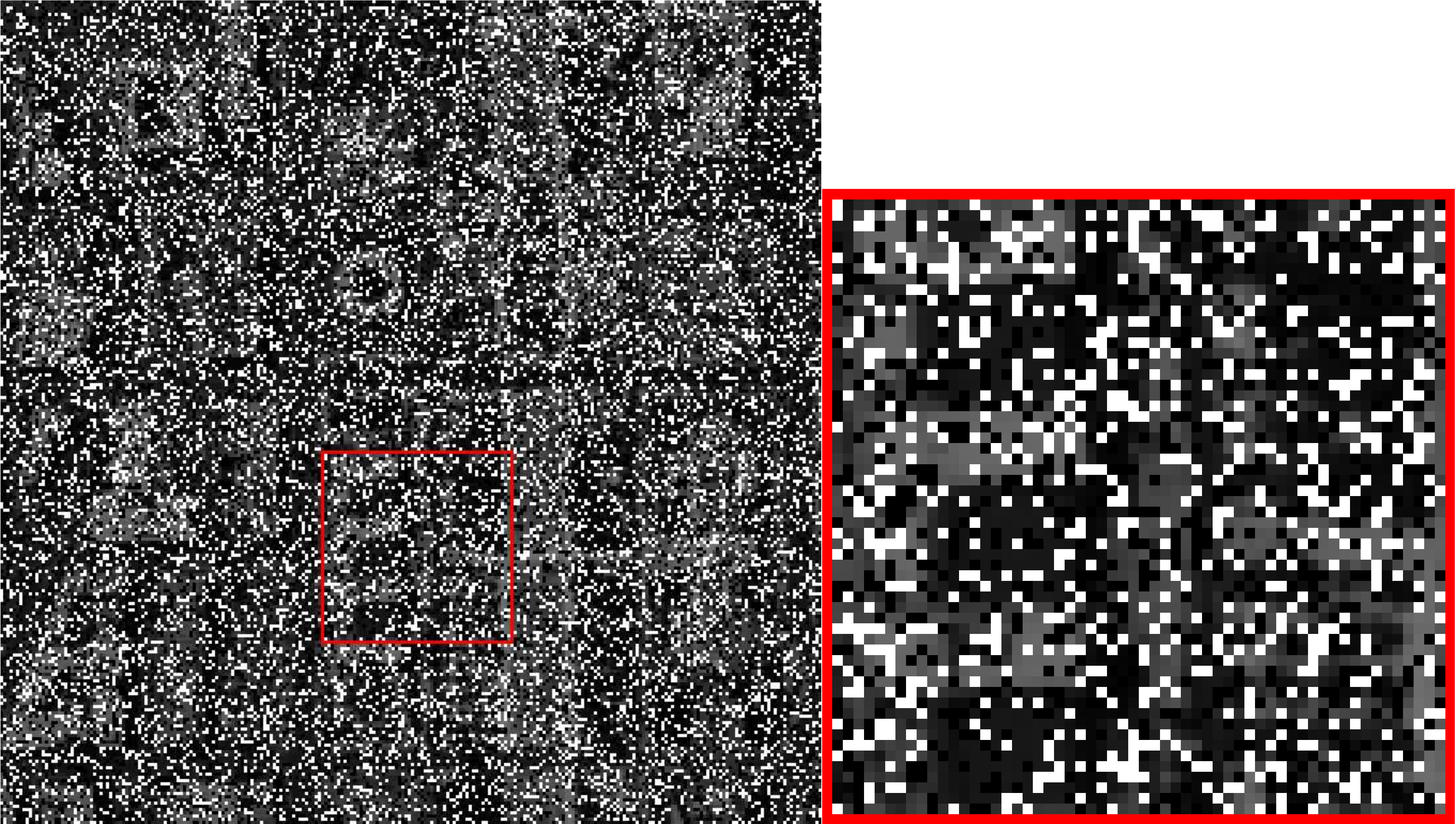}&
\includegraphics[width=0.195\textwidth]{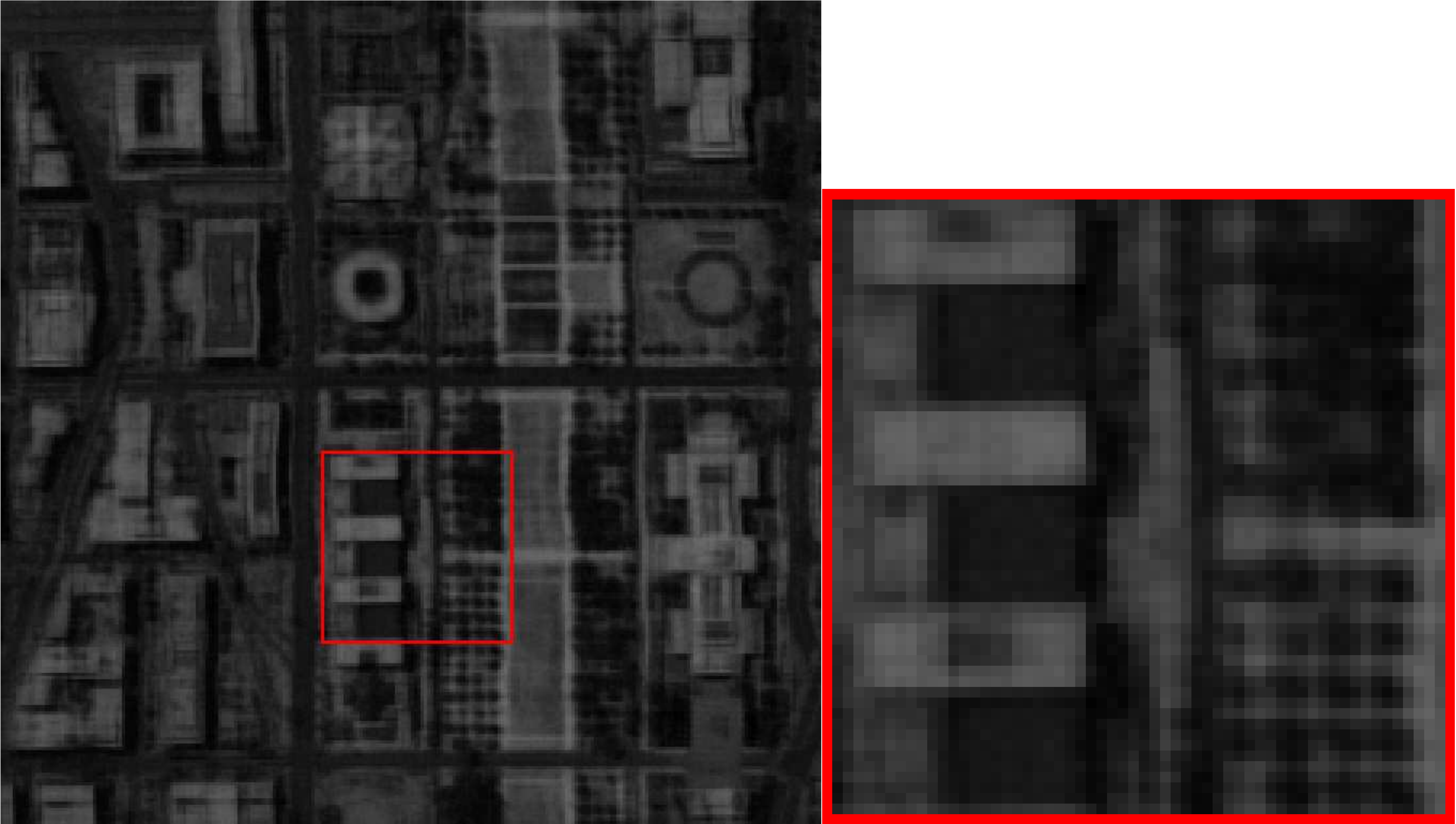}&
\includegraphics[width=0.195\textwidth]{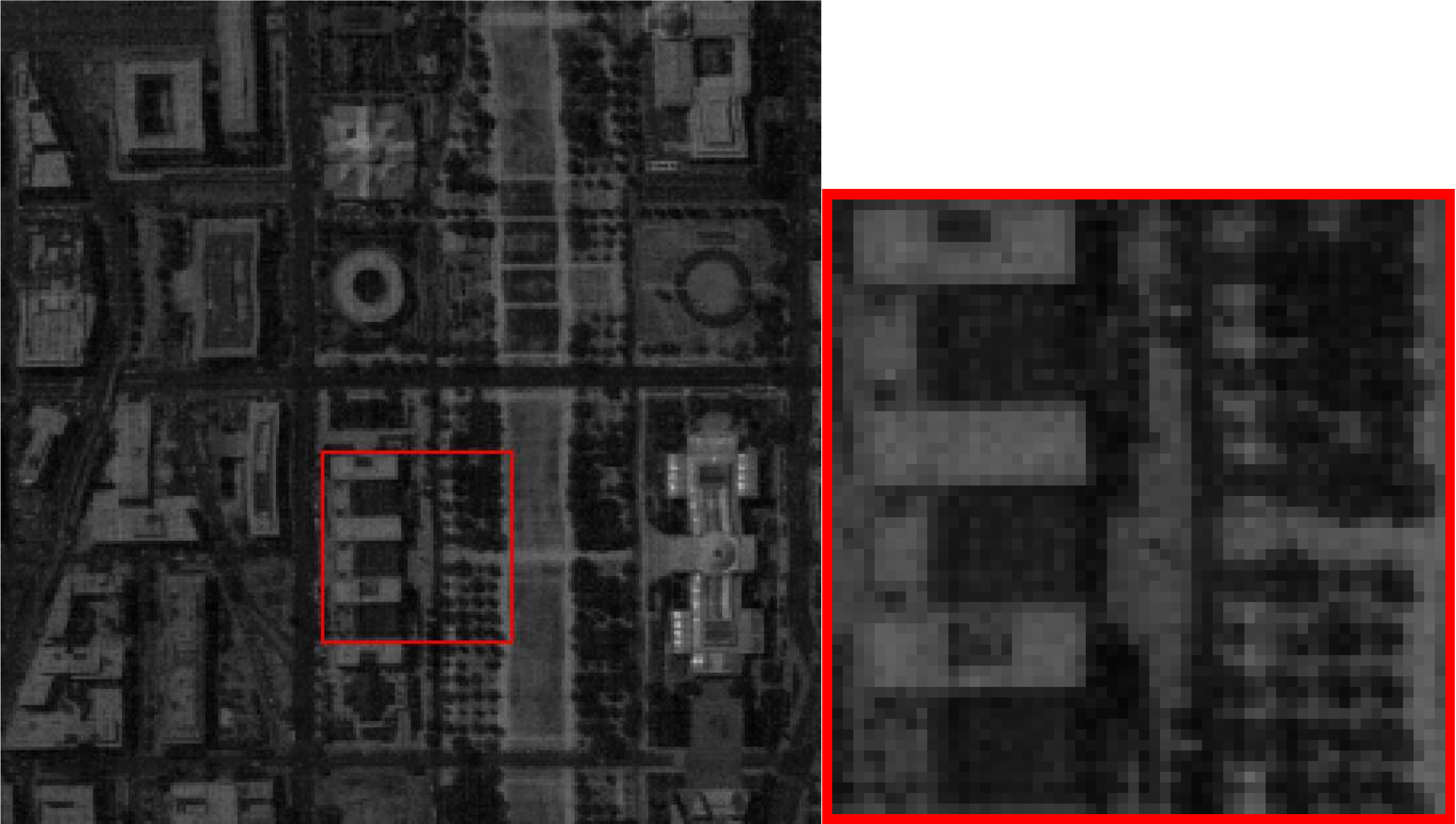}&
\includegraphics[width=0.195\textwidth]{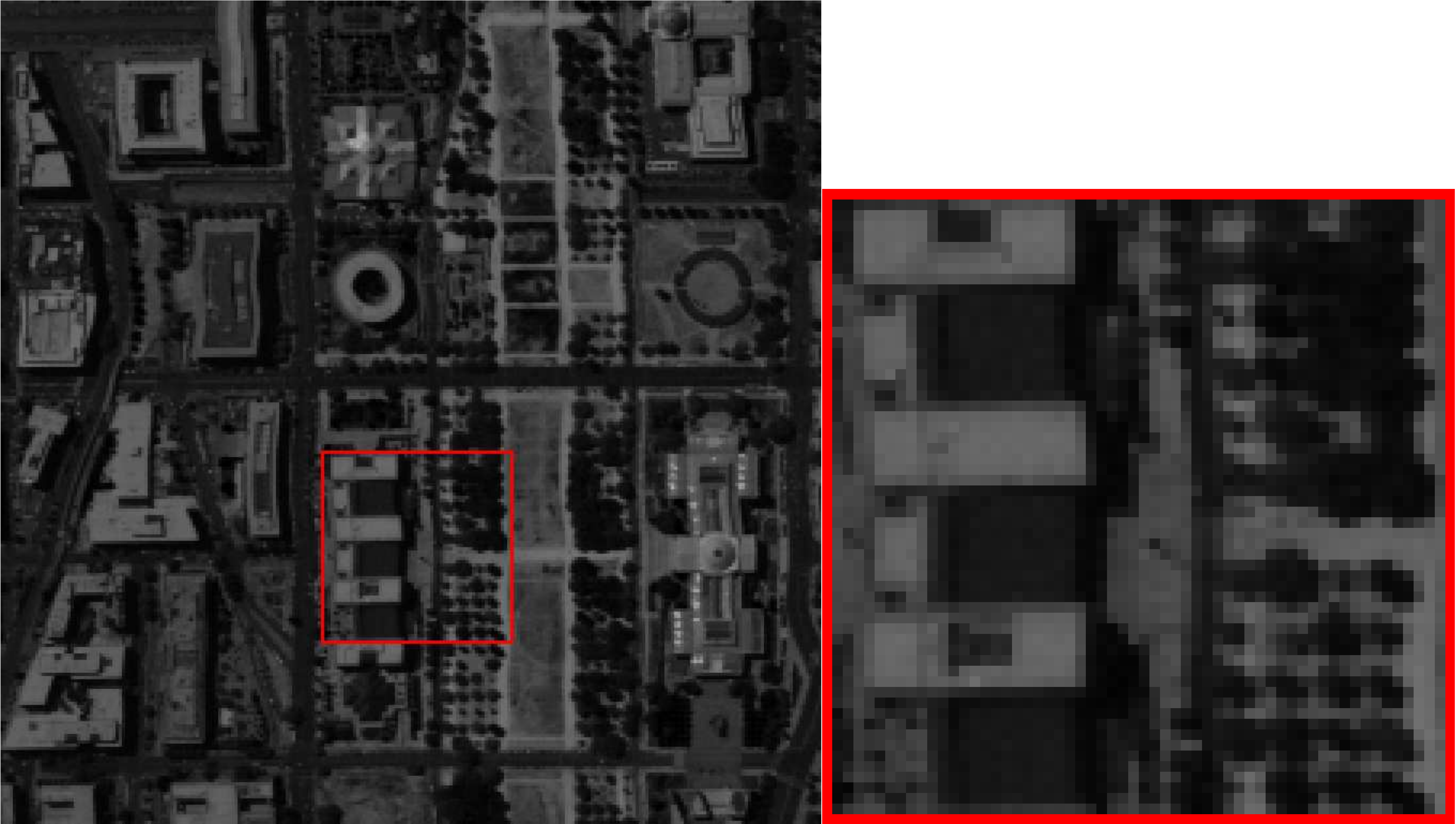}\\
\includegraphics[width=0.195\textwidth]{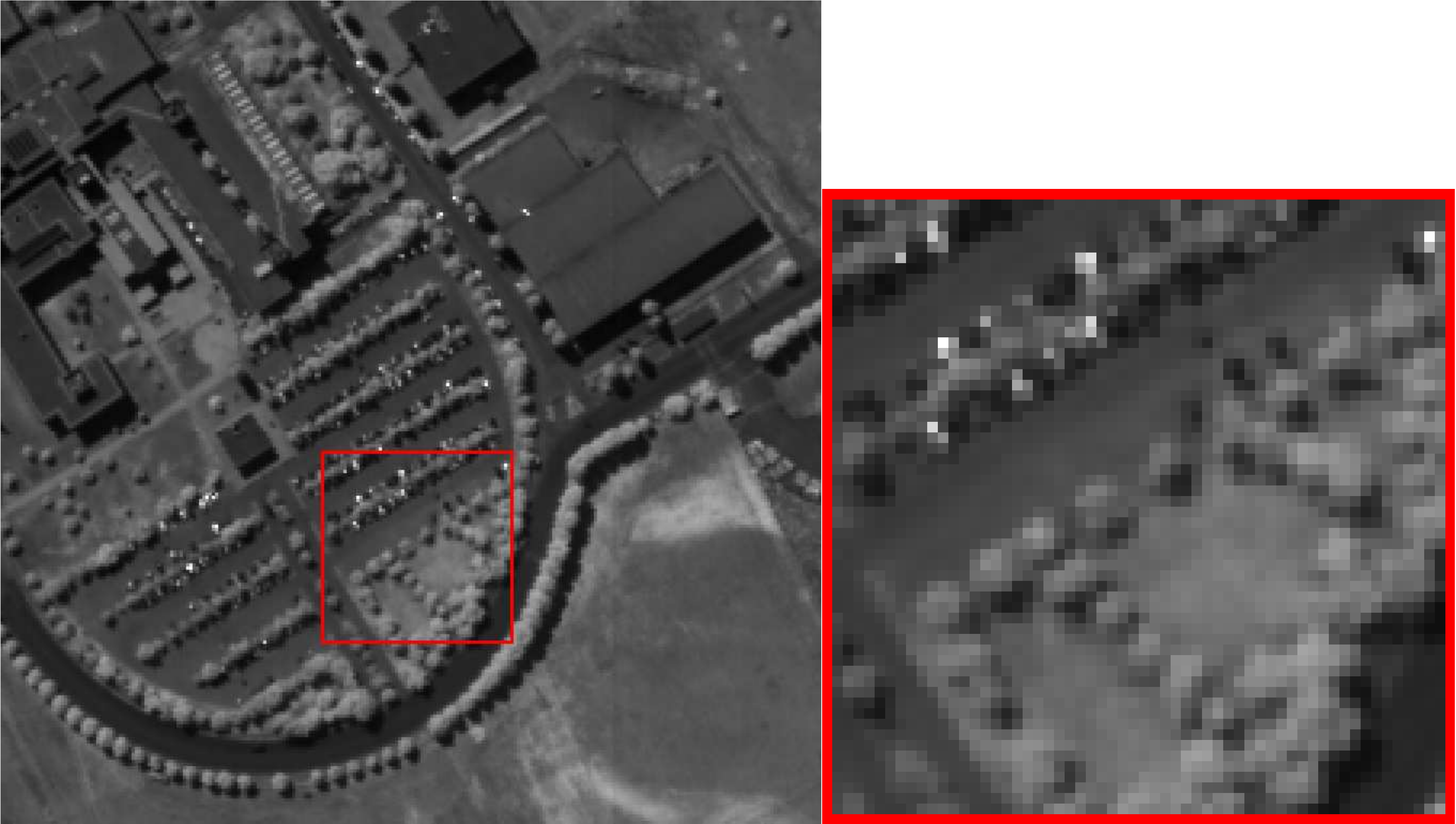}&
\includegraphics[width=0.195\textwidth]{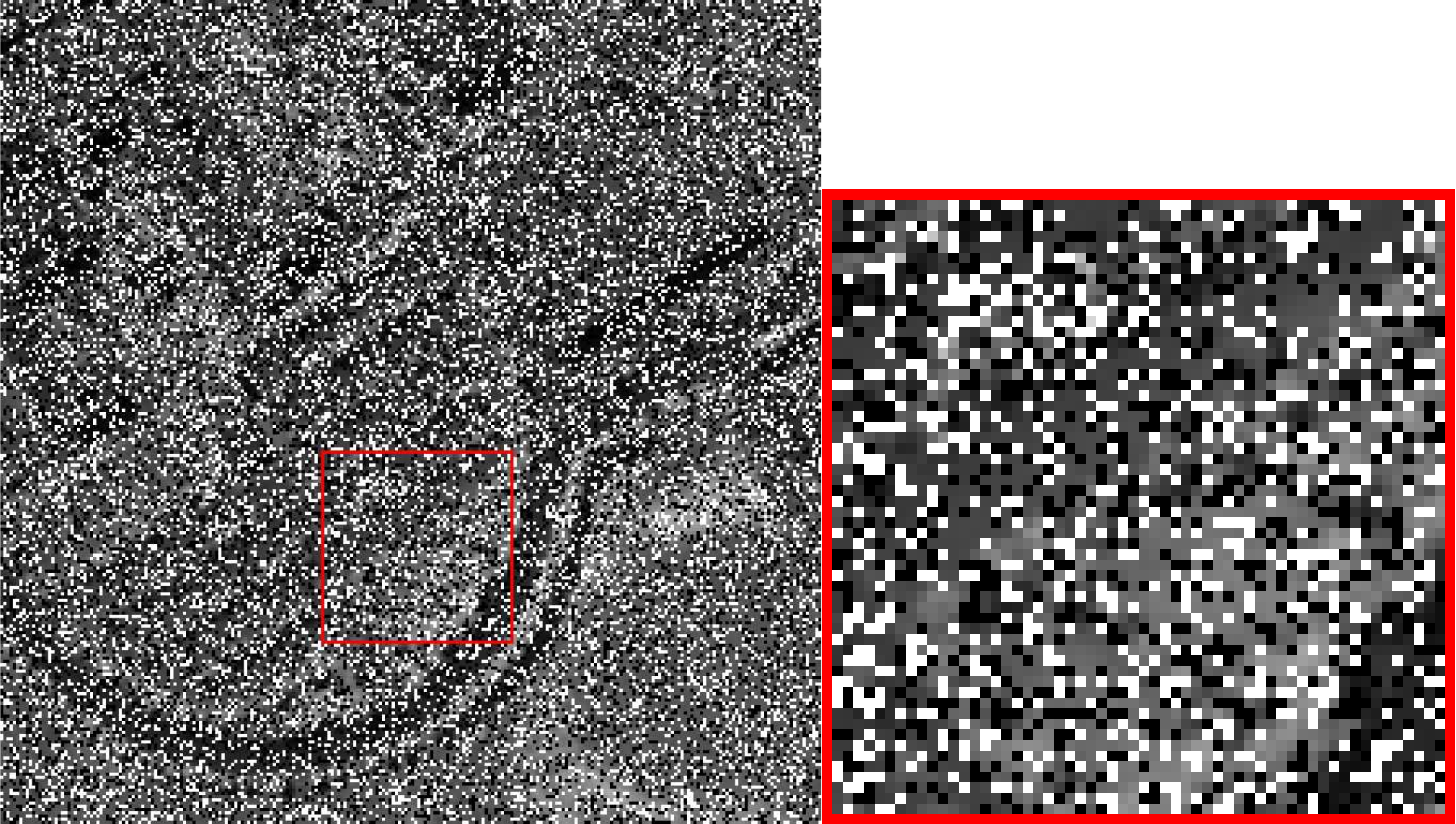}&
\includegraphics[width=0.195\textwidth]{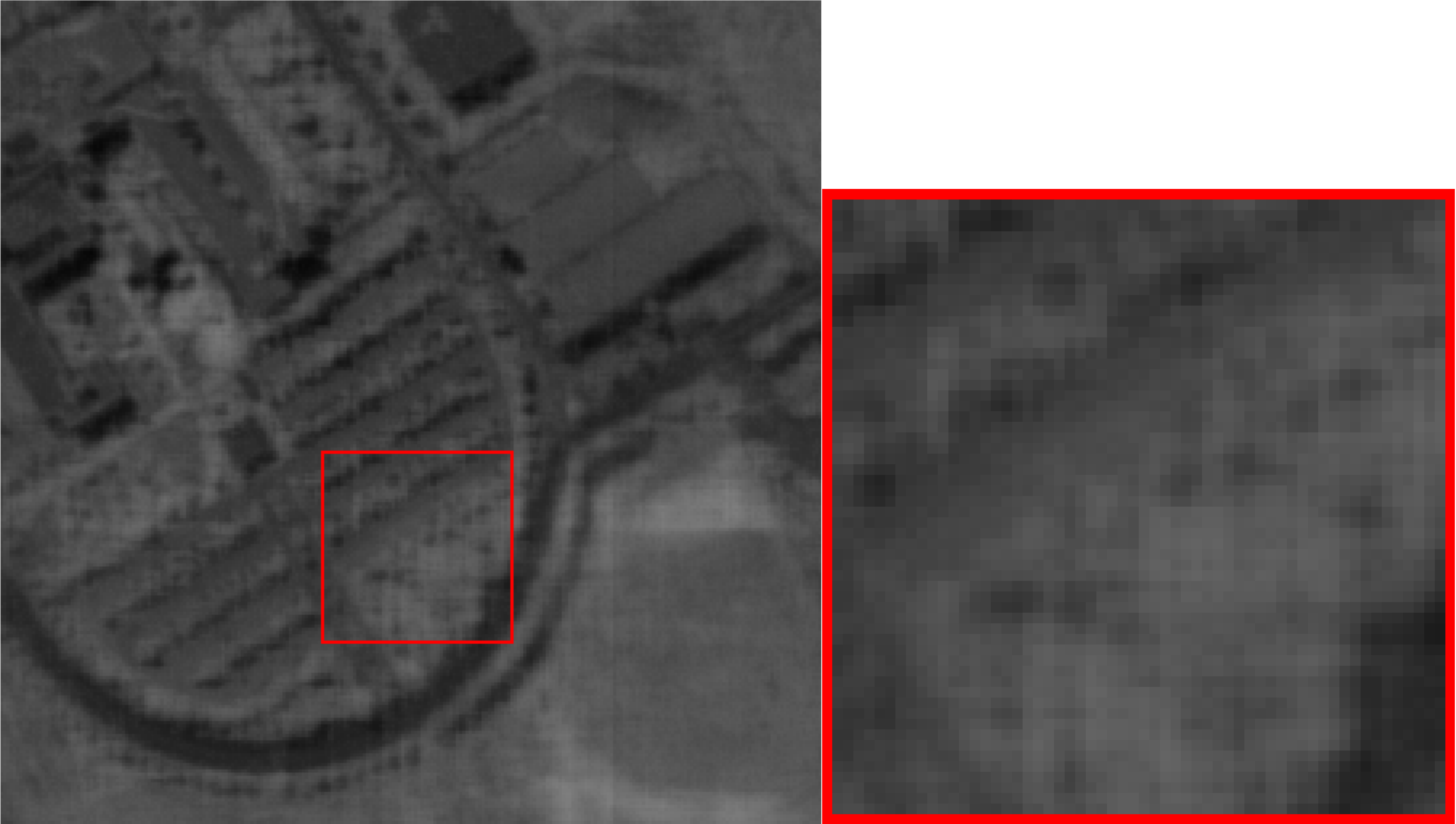}&
\includegraphics[width=0.195\textwidth]{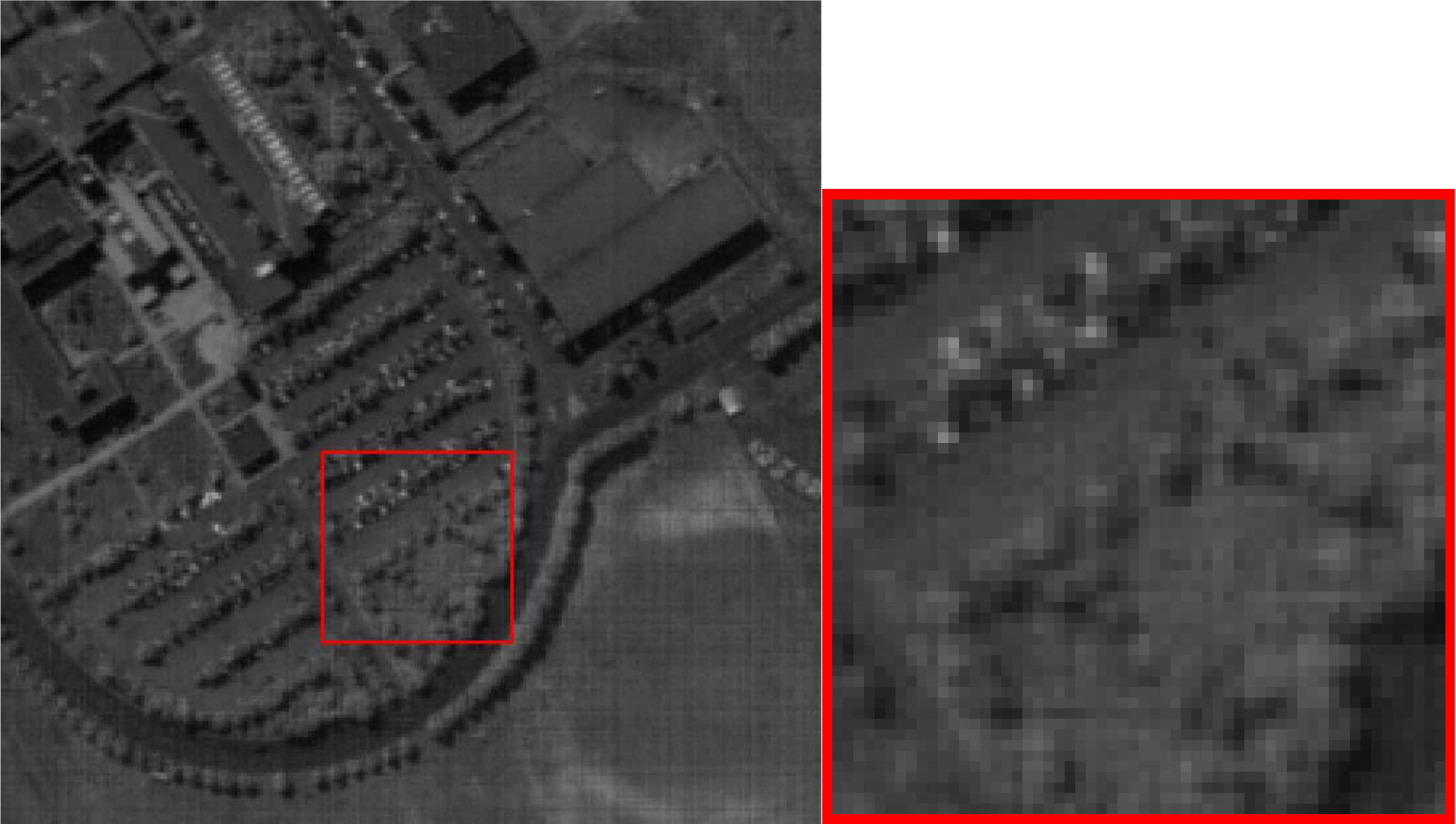}&
\includegraphics[width=0.195\textwidth]{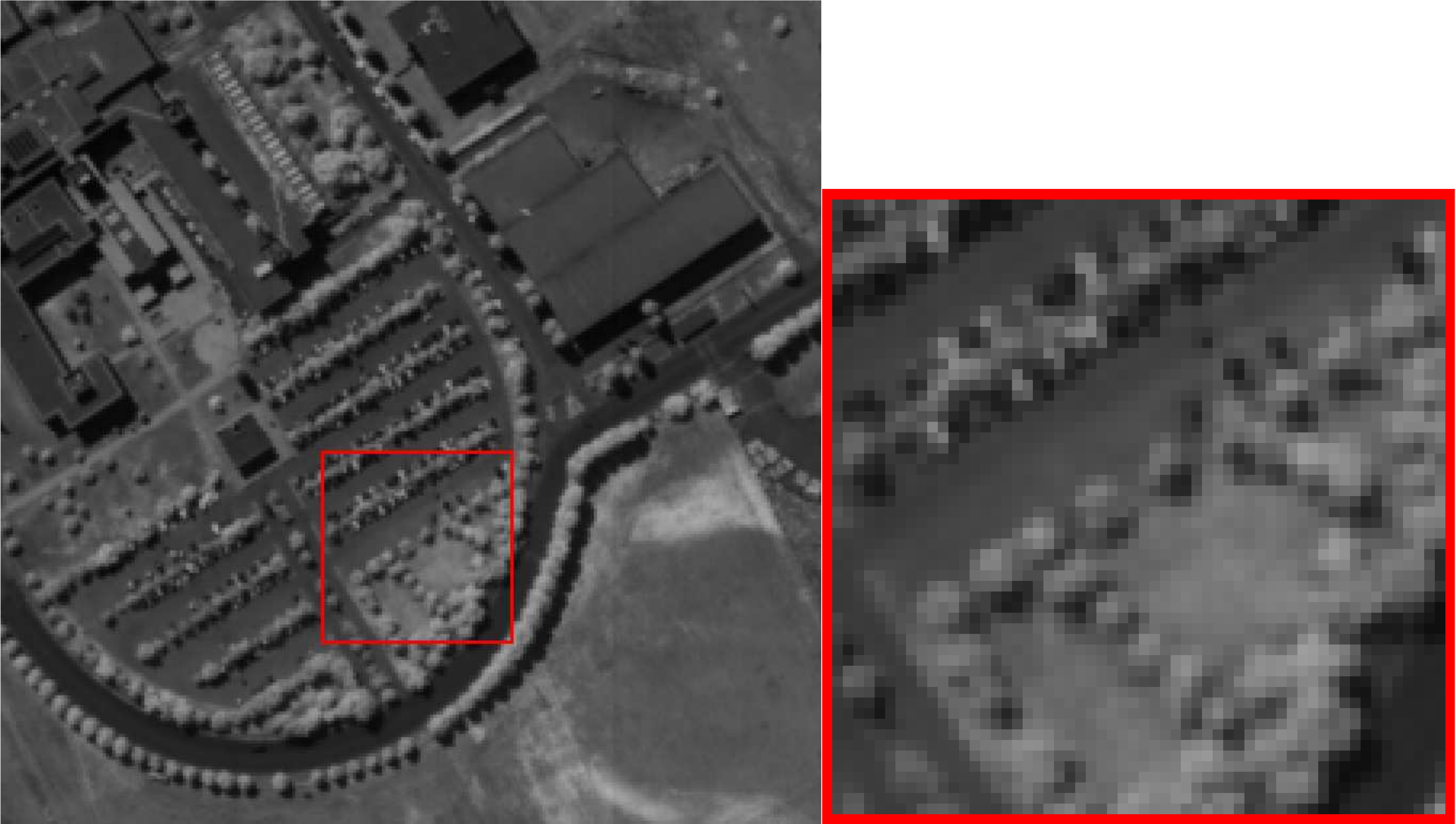}\\
Original& Observed & SNN  \cite{Goldfarb2014}&  TNN \cite{Lu2016} & WSTNN\\
  \end{tabular}
  \caption{The denoising results of the HSI \emph{Washington DC Mall} and \emph{Pavia University} with $\text{NL}=0.4$. Top row: the image located at the 150-th band in \emph{Washington DC Mall}. Bottom row: the image located at the 87-th band in \emph{Pavia University}.}
  \label{imageHSIpca}
  \end{center}\vspace{-0.6cm}
\end{figure}

\textbf{Parameter selection.} In all tests, the stopping criterion lies on the RelCha of two successive recovered tensors, i.e.,
$\text{RelCha}=\frac{\|\mathcal{L}^{(p+1)}-\mathcal{L}^{(p)}\|_{F}}{\|\mathcal{L}^{(p)}\|_{F}}< 10^{-4}$. The tuning parameter $\lambda$ is set to be
$$
\lambda=\sum_{1\leq k_1<k_2\leq N}  \frac{\alpha_{k_1k_2}}{\sqrt{\max(n_{k_1},n_{k_2})d_{k_1k_2}}}~\text{with}~d_{k_1k_2}=\prod_{s\neq k_1,k_2}n_s.
$$
Letting the threshold parameter $\tau=\alpha./\beta$, then the penalty parameter $\rho$ is set to be $\rho=1/\mathrm{mean}(\tau)$. This means that only the weight $\alpha$ and the threshold $\tau$ need to be adjusted. Table \ref{parasetTRPCA} shows these two parameters setting in the proposed WSTNN-based TRPCA method on different data, where $\alpha$ is chosen by the strategy of the weight selection in Section \ref{secmodel}, $\tau$ is set to be $\omega\times{\tt ones}(N(N-1)/2,1)$, and $\omega$ is empirically selected from a candidate set: $\{1,10,50,100,500,1000,10000\}$.

\begin{figure}[!t]
\scriptsize\setlength{\tabcolsep}{1pt}
\footnotesize
\begin{center}
\begin{tabular}{cc}
\multicolumn{2}{c}{\includegraphics[width=0.95\textwidth]{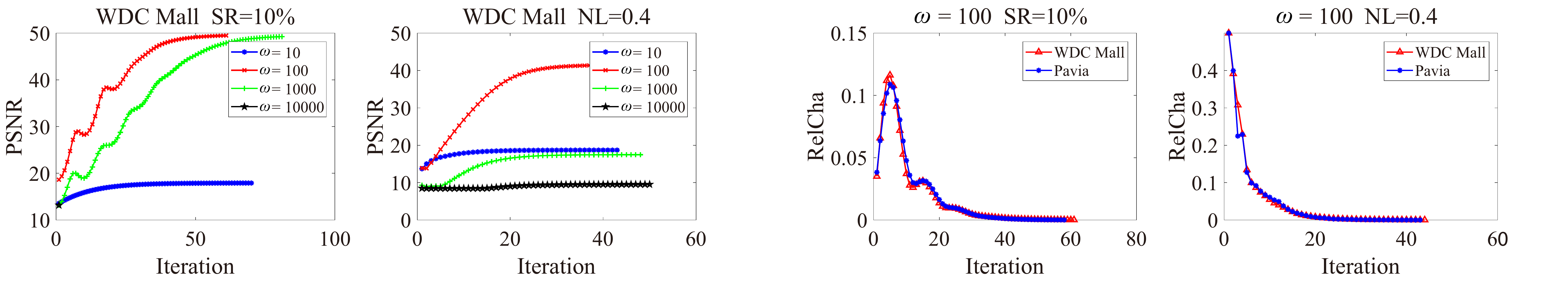}}\\
~~~~~~~~~~~~~~~~~~~~~~~~~~~~~~~~~~~~~~~~~~~(a)~~~~&~~~~~~~~~~~~~~~~~~~~~~~~~~~~~~~~~~~~~~~~~~~~(b)
\end{tabular}
\caption{(a) The PSNR values with respect to the iteration for different values of parameter $\tau$. Left column: completion tests. Right column: denoising tests. (b) The RelCha values with respect to the iteration with parameter $\tau=(100,100,100)$. Left column: completion tests. Right column: denoising tests.}
\label{para}
\end{center}\vspace{-0.6cm}
\end{figure}

\textbf{Synthetic data denoising.} We test three-way tensors of size $30 \times 30 \times 30$ and four-way tensors of size $30 \times 30 \times 30\times 30$ with different $N$-tubal rank and random salt-pepper noise level (NL). The $N$-tubal rank are set to be $r\times{\tt ones}(N(N-1)/2,1)~(r=1,2,...,20)$ and the NLs are set to be $0.025\times l~(l=1,2,\cdots,20)$. For each pair of $N$-tubal rank and NL, we conduct 50 independent tests and calculate the success rate. Fig. \ref{sypca} shows the success rates for varying $N$-tubal rank and varying NLs. The results illustrate that the proposed WSTNN-based TRPCA method is more robust and preferable than the TNN-based method \cite{Lu2016}.

\textbf{HSI denoising.} We test HSIs \emph{Washington DC Mall} and the \emph{Pavia University} data sets. The NL of random salt-pepper noise are set to be $0.2$ and $0.4$. Table \ref{HSIPCA} lists PSNR, SSIM, and FSIM values of the testing HSIs recovered by different methods. From these results, we observe that our method evidently performs better than other competing ones in terms of all the evaluation measures. In Fig. \ref{imageHSIpca}, we show one band in these two HSIs. As observed, our WSTNN-based TRPCA method achieves the best visual results among three compared methods, both in the noise removing and the details preserving.

\subsection{Parameter and convergence analysis}

In this section, we test the effects of the threshold parameter $\tau$ and the convergence of the proposed ADMM in the proposed LRTC and TRPCA problems. All tests are based on the HSI \emph{Washington DC Mall}.

\textbf{Effects of threshold parameter.} We set SR to $10\%$ in completion tests and NL to $0.4$ in denoising tests. And $\tau$ is set to be $\tau=(\omega,\omega,\omega)$.
The results is presented in Fig. \ref{para}{\color{blue} (a)}. As observed, for parameter $\tau$, too large or too small values output failure results, while the moderate values obtain the best results. This observation is consistent with the theoretical analysis. That is, for completion tests, if the parameter $\tau$ is too large (e.g., $(10000,10000,10000)$), all the singular values are replaced by 0, the algorithm iterates only one step and outputs the partial observation tensor $\mathcal{F}$. If the parameter $\tau$ is too small (e.g., $(10,10,10)$), the singular values after performing the t-SVT (in Theorem \ref{Ytheorem}) contain many corrupted information, i.e., it is not consistent with the low-rank prior of the underlying tensor. Similarly, for denoising tests, if the parameter $\tau$ is too large or too small, the low-rank term is out of action. Under the guidance of Fig. \ref{para}{\color{blue}(a)}, parameter $\tau$ is set to be $(100,100,100)$ in all experiments conducted on real-world data.

\textbf{Convergence analysis.} Owing to the use of ADMM framework and the convexity of the objective functions, the convergence of two developed algorithms is guaranteed theoretically. Empirically, such the convergence can be visually observed from Fig. \ref{para}{\color{blue}(b)}, where $\tau$ is set to be $(100,100,100)$.

\section{Conclusions} \label{sec:Con}
In this paper, we defined the mode-$k_1k_2$ tensor unfolding to reorder the elements of an $N$-way tensor into a three-way tensor, and then performed the t-SVD on each mode-$k_1k_2$ unfolding tensor to depict the correlations along different modes. Based on this, we proposed the corresponding tensor $N$-tubal rank and its convex relaxation WSTNN.
To illustrate the effectiveness of the proposed $N$-tubal rank and WSTNN, we applied WSTNN to two typical LRTR problems, LRTC and TRPCA problems, and proposed the WSTNN-based LRTC and TRPCA models. Meanwhile, two efficient ADMM-based algorithms were developed to solve the proposed models. Numerical results demonstrated the proposed method effectively exploit the correlations along all modes while preserving the intrinsic structure of the underlying tensor.

\section*{Acknowledgments}

This work is supported by the National Natural Science Foundation of China (61772003, 61876203), the Fundamental Research Funds for the Central Universities (ZYGX2016J132, 31020180QD126), the National Postdoctoral Program for Innovative Talents (BX20180252), and the Project funded by China Postdoctoral Science Foundation (2018M643611).

\vspace{1cm}


%
{\small
\bibliographystyle{ieeetran}
\bibliography{refference}
}

\end{document}